\documentclass{article}
\usepackage{PRIMEarxiv}
\usepackage{multirow}
\usepackage{graphicx} % Required for inserting images
\usepackage{algpseudocode}
\usepackage{algorithm}
\usepackage{amsmath}
\usepackage{amsfonts}
\usepackage{amssymb}
\usepackage{amsbsy}
\usepackage{bbm}
\usepackage{xcolor}
\usepackage{hyperref}
\usepackage{float}

\newcommand{\tl}{{t_{\ell}}}
\newcommand{\hsigma}{{\widehat{\sigma}}}

\newcommand{\ologrr}{O\left(\sqrt{\frac{\log\log r}{r}}\right)}
\newcommand{\orlogr}{O\left(\sqrt{r\log\log r}\right)}
\newcommand{\orrlogr}{O\left(r\sqrt{r\log\log r}\right)}
\newcommand{\rlogr}{\sqrt{r\log\log r}}
\newcommand{\rrlogr}{r\sqrt{r\log\log r}}

\usepackage[utf8]{inputenc} % allow utf-8 input
\usepackage[T1]{fontenc}    % use 8-bit T1 fonts
\usepackage{hyperref}       % hyperlinks
\usepackage{url}            % simple URL typesetting
\usepackage{booktabs}       % professional-quality tables
\usepackage{amsfonts}       % blackboard math symbols
\usepackage{nicefrac}       % compact symbols for 1/2, etc.
\usepackage{microtype}      % microtypography
\usepackage{lipsum}
\usepackage{fancyhdr}       % header
\usepackage{graphicx}       % graphics
\graphicspath{{media/}}     % organize your images and other figures under media/ folder
\newtheorem{assumption}{Assumption}
\newtheorem{lemma}{Lemma}
\newtheorem{theorem}{Theorem}
\title{Ranking and Selection with Simultaneous Input Data Collection
}
\author{
  Yuhao Wang \\
  School of Industrial and Systems Engineering\\
  Georgia Institute of Technology \\
  Atlanta\\
  \texttt{yuhaowang@gatech.edu} \\
   \And
  Enlu Zhou \\
  School of Industrial and Systems Engineering\\
  Georgia Institute of Technology \\
  Atlanta\\
  \texttt{enlu.zhou@isye.gatech.edu} \\
}

\begin{document}
\maketitle

\begin{abstract}
In this paper, we propose a general and novel formulation of ranking and selection with the existence of streaming input data. The collection of multiple streams of such data may consume different types of resources, and hence can be conducted simultaneously. To utilize the streaming input data, we aggregate simulation outputs generated under heterogeneous input distributions over time to form a performance estimator. By characterizing the asymptotic behavior of the performance estimators, we formulate two  optimization problems to optimally allocate budgets for  collecting input data and running simulations. We then develop a  multi-stage simultaneous budget allocation procedure and provide its statistical guarantees such as consistency and asymptotic normality. We conduct several numerical studies to demonstrate the competitive performance of the proposed procedure. 
\end{abstract}

% keywords can be removed

\section{Introduction}\label{Sec:intro}
\textbf{Ranking and selection} (R\&S) are fundamental techniques in scenarios where multiple system designs must be evaluated and compared, allowing for the identification of the best system design with a high confidence. The evaluation of a complex stochastic system is often through
stochastic simulation, which can be expensive or time-consuming due to the complicated system structure. Ideally one aims to minimize the total simulation effort as well as maximize the confidence of the selected best design. However, these two objectives are usually against each other as more simulation effort helps with more accurate evaluation of the system performance. Therefore, the existing R\&S procedures usually take one as the objective and the other as the constraint, and consequently can be categorized into two formulations: the fixed budget formulation, and the fixed confidence formulation. The fixed budget formulation aims to maximize the probability of correct selection (PCS), which is the probability that the procedure selects the true optimal system design, with a given simulation budget. Methods for fixed budget R\&S include but are not limited to the optimal computing budget allocation (OCBA) framework in \cite{chen2000simulation,glynn2004large}, the expected value of information (EVI) approach in \cite{chick2001new,chick2010sequential}, the knowledge gradient (KG) method in \cite{frazier2008knowledge,ryzhov2016convergence}.
The fixed confidence formulation, on the other hand, aims to minimize the simulation effort to achieve a given target PCS guarantee.  Methods for fixed confidence R\&S include but are not limited to indifference-zone (IZ) approaches (\cite{bechhofer1954single,rinott1978two,kim2001fully,frazier2014fully}, IZ-free approach \cite{fan2016indifference,wang2023bonferroni}. In this paper, we will focus on the fixed budget formulation.   
%{\color{red} The rest of this paragraph seems out of place. I added the next sentence trying to connect, but still seems awkward. Maybe review both fixed-confidence and fixed budget literature.}  

% {\color{red} (did you copy the following sentence from our previous paper? please rewrite.) In many application areas such as supply chain and healthcare, the simulation model needs to  account for not only the system's internal logic but also the random factors in the real system, e.g., random demand, transit lead time in an inventory control problem. 
% These random factors are often modeled by probabilistic distributions, which are taken as an input by the simulation model. }

The simulation is driven by the so-called ``input distribution", which captures the random factors in the real-world system, e.g., random demand, production lead time, and transit lead time in an inventory control problem. The classical R\&S procedure makes an implicit assumption that input distribution is known; however, in practice, the true input distribution are usually unknown and need to be estimated from data observed from the real system. These data are referred to as the ``input data". The availability of the input data is usually limited, and they can be either expensive to obtain or have limited data source within a period. Lack of input data
leads to inaccurate estimation of the input distribution, which further leads to model mis-specification between the input distribution and the true randonmness. Such mis-specification, often referred to as the ``input uncertainty (IU)", can severely inhibit the performance of a R\&S procedure, as the simulation cannot reduce the negative impact caused by IU. Input uncertainty quantification has been well-studied in a large body of work, including but not limited to 
\cite{barton1993uniform,draper1995assessment,cheng1997sensitivity,zouaoui2003accounting,zouaoui2004accounting,ng2006reducing,barton2014quantifying,xie2014bayesian,lin2015single,song2015quickly,xie2016multivariate,lam2017empirical,feng2019efficient,zhu2020risk,lam2021subsampling}. It is important to consider input uncertainty when designing R\&S procedures. 

Existing works studied several distinct settings for R\&S with the existence of IU. %{\color{red} there are some early papers by Biller and Corlu on this. } 
The earlier works (\cite{corlu2013subset,corlu2015subset, wu2017ranking,gao2017new,xiao2018simulation,song2019input,xu2020joint,fan2020distributionally,xiao2020optimal}) considered a fixed set of input data, where the input distributions are only estimated once and all simulations are run under fixed input distributions. %\cite{corlu2013subset,corlu2015subset,gao2017robust,xiao2018simulation,song2019input,xu2020joint,fan2020distributionally,xiao2020optimal} studied the setting with an initial given set of historical input data. 
In particular, \cite{corlu2013subset,corlu2015subset,song2019input,fan2020distributionally} took a fixed confidence formulation and developed confidence intervals that account for IU with limited historical data; whereas, \cite{gao2017robust,xiao2018simulation,xiao2020optimal} took a fixed budget formulation.  \cite{wu2017ranking}, followed by \cite{xu2020joint}, are the first to consider the issue of input data collection in R\&S with IU. They adopted a joint budget allocation framework, which first allocates part of the budget to collect input data and then allocates the remaining budget to guide simulations. All these works share in common that input data are collected all at once before simulation and no new input data are observed or used when running simulations.

In contrast to the above assumption of fixed input data, many practical application problems can observe or collect input data in a ``streaming" fashion, where batches of data, possibly of varying and random sizes, become available periodically. For instance, Alibaba, which operates one of the world’s largest e-commerce platforms, recently adopted a simulation-optimization approach to model and solve complex inventory problems under different network structures, product life cycles, and inventory policies (as detailed in \cite{deng2023alibaba}). Their large simulation model incorporates several features as input, such as supply availability, transit lead time, and consumption patterns. These features are often unknown and must be estimated from real-world data. As these complex simulation models run, new input data, such as transit lead time, become available periodically (e.g., daily, weekly). Other data, such as complex market consumption patterns (including behaviors, preferences, and trends exhibited by consumers), must be collected through surveys, which require the hiring of specialists, making the process time-consuming. It is crucial for decision-makers to carefully design surveys to collect more data for the most significant input distributions—those that have a larger impact on simulation outputs.

Consider another example in drug design. Simulation has proven to be a powerful tool for modeling the behavior and interactions of drugs with biological systems (e.g., see \cite{macalino2015role,yu2017computer}). This approach helps accelerate the drug discovery process, reduce costs, and improve the understanding of drug mechanisms. In drug design, input distributions, including the distribution of properties such as molecular weight, logP (partition coefficient), and solubility, as well as parameters like clearance and half-life, are crucial for estimating various properties and outcomes. To estimate these distributions and run simulation models efficiently, decision-makers must carefully allocate resources to conduct different experiments to collect data for various input distributions. These experiments often require specialists with specific skills and can have a very long experiment cycle. Efficiently assigning specialists from different backgrounds to different tasks can significantly improve efficiency.
%It is beneficial to incorporate these data to reduce IU and adjust the simulation model.

As in many application problems, such as those faced by Alibaba and in drug design, both simulation and input data collection are time-consuming. A simulation framework with streaming input data is desirable, where input distributions are periodically updated with newly collected data, and simulations are run under these updated distributions. However, the challenge with this framework is that the estimated input distribution changes over time with streaming data, leading to correlated and differently distributed simulation outputs. Furthermore, the interarrival time between input data may be comparable to the simulation time for complex systems, resulting in only a limited number of simulation replications between two adjacent batches of input data. To reduce simulation uncertainty (SU) due to the limited number of simulation outputs, one approach is to aggregate simulation outputs across different time stages. However, this aggregation prohibits the use of analysis approaches from existing R\&S works, where the assumption of identical and independently distributed (i.i.d.) simulation outputs is crucial but fails in the current context.

%In this work, we consider the more general setting of streaming input data. ``Streaming" means the input data are not gathered only once, but arrive periodically. For example, the daily customer demand observed each day. When new data are available, one can update the input distribution and run the simulation under the current estimate. It is always beneficial to incorporate the new data to reduce the input uncertainty. 

%\red{Are Steve Chick's papers about how to collect data to reduce input uncertainty? If yes, you may cite them here, saying something like that data collection has also been considered for input uncertainty reduction.}\blue{ One of the paper (\cite{ng2006reducing} is about collecting data to reduce input uncertainty, the other \cite{ng2004design} is not related to IU. See the blue sentence in the middle of the paragraph. Also, since we cite Chick's paper here, Do you think whether we shold keep the literature review on IU quantification on line 48-52?} 

There are only a few recent works that consider R\&S with streaming input data: \cite{wu2022data} proposes a fixed confidence approach, while \cite{wang2022fixed,kim2022optimizing,wang2023wsc} consider the fixed budget formulation. Both \cite{wu2022data} and \cite{wang2022fixed} treat the streaming input data as given, meaning they cannot control the amount of input data but can only run simulations according to the current availability of data. Works such as \cite{ng2006reducing, wu2017ranking, xu2020joint, kim2022optimizing, wang2023wsc} consider actively collecting input data to reduce input uncertainty (IU) at a cost, assuming that data collection for all input distributions shares a common budget. However, in practice, due to the large complexity of the system, data collections for different input distributions may consume different budgets, and their costs may not be comparable. For instance, in the Alibaba example, specialists are required to conduct surveys to collect data for consumption patterns, whereas data such as transit lead time simply arrive periodically. Within a given time window, the amount of data for different consumption patterns share a common budget controlled by the decision-maker through survey design, but constrained by the availability of specialists and the length of the time window. The amount of data for transit lead time can be considered as having its own budget, determined by the length of the time window. Similarly, in drug design, conducting experiments to collect data for different properties or parameters requires specialists from different backgrounds. In both examples, when a decision is required within a certain amount of time, different budgets need to be allocated simultaneously to collect input data for different input distributions and run simulations, fully utilizing different resources.

In this work, we propose a simultaneous budget allocation framework for R\&S that accommodates different types of input data. This framework generalizes the settings of aforementioned previous works, including \cite{wang2023wsc}, the early conference version of this paper, making it more practical. Additionally, compared to the conference paper \cite{wang2023wsc}, we provide a more comprehensive analysis of the performance of the proposed methods, both theoretically and numerically.
%\cite{wu2022data} took a fixed confidence formulation and \cite{wang2022fixed,kim2022optimizing}
%took a fixed budget formulation.

In addition, we consider general input distributions with minimal assumptions. This is a key distinction from both \cite{wang2022fixed} and \cite{kim2022optimizing}, which impose strong assumptions on the input distributions, requiring that the input distribution either has a finite support or belongs to a known set of finite distributions. Under such assumptions, they treat each input realization or candidate distribution as a simulation scenario and run the simulation for a design under a fixed scenario each time. This avoids non-i.i.d. simulation outputs, which is the main challenge in the streaming data setting, as mentioned above.
In contrast, we work with general input distributions whose support can be continuous or discrete. As a result, we need to run simulations under the current estimated input distribution each time it is updated. We propose a procedure that simultaneously allocates multiple budgets to collect input data and run simulations, considering the impacts of input uncertainty (IU) and simulation uncertainty (SU). We further provide statistical guarantees for the proposed procedure.

We summarize the contributions of this paper as follows.
\begin{enumerate}
    \item We propose a general  framework for a fixed budget R\&S with general input data. Compared to previous works on R\&S with streaming input data, our problem setting features two key generalizations: (1) it accommodates multiple input data streams with varying budgets and allocates these budgets simultaneously, and (2) it allows for more general input distribution assumptions, including continuous support and parameter space. Both generalizations make our framework more practical.

    \item To utilize the streaming input data, we aggregate
simulation outputs generated under heterogeneous input distributions over time to form a performance estimator. By characterizing the asymptotic behavior of the performance estimator, we formulate two optimization problems to optimally allocate budgets for collecting input data and running simulations. Based on the two optimization problems, we develop a multi-stage simultaneous budget allocation (SBA) procedure that dynamically allocates multiple budgets to run simulations and collect input data.  We also provide statistical guarantees on the performance of the proposed procedure, which requires careful convergence analysis of the algorithm's dynamic and several estimators that consist of non-i.i.d. simulation samples. 
    % In addition, in Theorem \ref{thm: optimality rate} we characterize the convergence rate of the proposed procedure. 
    \item 
    We conduct numerical experiments on two examples to demonstrate the efficiency of SBA. In both examples, SBA has the best performance. In particular,  SBA outperforms the joint budget allocation procedure (namely ``JBA") in \cite{wu2017ranking}, since SBA fully utilizes the different resources and dynamically adjusts the allocation policy for both simulation and input data collection as more information (simulation output and input data) is revealed. %\blue{ (not sure if we want to emphasize this here.) What's more, when we restrict the general input data collection to only given input data, which is the setting considered in \cite{wang2022fixed}, SBA still outperforms the procedure in \cite{wang2022fixed}, which suffers the issue of discretizing input distributions.}
    % \red{please add a paragraph on the numerical findings.}
\end{enumerate}

The rest of the paper is organized as follows. In Section \ref{sec: problem statement}, we introduce the simultaneous budget allocation framework with different types of input data. In Section \ref{sec: rate optimization}, we formulate two convex optimization problems to guide allocation of input budgets and simulation budget, with the objective of maximizing the asymptotic convergence rate of probability of acceptable estimation (PAE) and probability of correct selection (PCS), respectively. For the latter, we also derive the optimality conditions that characterizes the optimal allocation policies. In Section \ref{sec: parameter and procedure}, we develop the multi-stage simultaneous budget allocation procedure. We provide statistical guarantees on the performance of the proposed algorithm in Section \ref{sec: consistency and optimality}. We numerically test the performance of the proposed algorithm  in Section \ref{sec: numerical}, and finally conclude the paper in Section \ref{sec: conclusion}.

% There have been extensive works concerning the input uncertainty quantification for a single simulation system. We refer interested reader to \cite{corlu2020stochastic} for a comprehensive overview. For R\&S with input uncertainty, several works considered the setting when there is a fixed set of input data.
% \subsection*{Literature Review}
% The literature on ranking and selection has a long story, with the  
%  earliest work dating back to the 1950s \citep{gupta1956decision,bechhofer1954single}. 
%  % The early work of R\&S focused on optimizing the procedure for conducting actual experiments (such as land parcel selection), whereas nowadays the focus switches to optimizing the procedure for running simulations due to the development of computing techniques. 
% Most of the works on R\&S can be classified into the categories of fixed budget R\&S and fixed confidence R\&S, both of which have been widely studied. Methods for fixed budget R\&S include but are not limited to the optimal computing budget allocation (OCBA) framework in \cite{chen2000simulation,glynn2004large}, the expected value of information (EVI) approach in \cite{chick2001new,chick2010sequential}, the knowledge gradient (KG) method in \cite{frazier2008knowledge,ryzhov2016convergence}. Methods for fixed confidence R\&S include but are not limited to indifference-zone (IZ) approaches (\cite{bechhofer1954single,rinott1978two,kim2001fully,frazier2014fully} as well as IZ-free approaches \cite{fan2016indifference}.

\section{Problem Statement} \label{sec: problem statement}
\subsection{Basic Setting}
 Suppose we want to compare designs in a given finite set $\mathcal{K}$, and denote $K = |\mathcal{K}|$ the number of designs. For simplicity, we can assume $\mathcal{K} = \{1,2,\ldots,K\}$. Let $F_i^c(\cdot)$ denote the unknown input distribution of the randomness that affects the performance of design $i$. Let $X_i(F_i^c)$ and  $\mu_i(F_i^c) = \mathbb{E}[X_i(F_i^c)]$ denote the random performance and the expected performance, respectively, of design $i$. Our goal is to identify the best design with the largest expected performance, that is,
 $$
 b \in \arg\max_{i\in\mathcal{K}} \mu_i(F_i^c).
 $$
 
 Throughout the paper, we assume that all designs share the same input distribution, that is, $F_i^c = F^c$. 
 Note this does not limit any practical usage since we can simply include all design-specific input distributions into the common input distribution. In this case, every design shares the full information of all input distributions, although its performance may only depend on part of the input distributions that are relevant.  
 We make the following assumption on the commonly shared input distributions.

%{\color{red} why do you assume different input stream belong to the same parametric family? for example, customer demands and service times can follow different distribution families.}
\begin{assumption}(Parametric input distribution) \label{assump: parametric input}\\
    All designs share the same input distributions $F^c$, which contain $S$ mutually independent input distributions $F_s^c, s=1,\ldots, S$. Each input distribution $F_s^c$ belongs to some known parametric family $\{F_{\mathbf{\theta}_s}(\cdot) | \mathbf{\theta}_s \in {\Theta}_s\}$ with  density $\{f_\mathbf{\theta_s}(\cdot)|\theta_s\in\Theta_s\}$  but
    unknown true parameters $\theta^c_s, s =1,\ldots,S$. 
\end{assumption}
Let $\mathbf{\theta} = (\theta_1,\ldots,\theta_S)$ and $\Theta = \Theta_1\times\ldots\times\Theta_S$. Assumption \ref{assump: parametric input} is common in the literature of R\&S, by which the input distributions $F_{\theta}$ is then a product measure of $F_{\theta_s}, s=1,2,\ldots,S$.  That is, $F_{\theta} = \prod_{s=1}^S F_{\theta_s}$. Accordingly, we use $X_i(\mathbf{\theta})$ and $\mu_i(\mathbf{\theta})$ to denote the random and expected performance of the $i^{th}$ design under input distributions $F_{\mathbf{\theta}}$, respectively. Since the true input parameter value $\mathbf{\theta}^c$ is unknown, one needs to collect input data to get an estimator $\hat{\mathbf{\theta}}$. We make the following assumption on the input data and the associated estimator of input parameter.

\begin{assumption} (Estimator of input parameters) \label{assump: unbiased estimator}
\begin{enumerate}
    
    \item  For each $\mathbf{\theta}^c_s$, the input data $\zeta_{s,1},\zeta_{s,2},\ldots$ are independent and identically distributed (i.i.d.) with distribution $F_{\mathbf{\theta}^c_s}.$
    \item For each $\mathbf{\theta}^c_s$, there exists a function (mapping) $D_s$ such that 
    \begin{equation*}
         \mathbb{E} [D_s(\zeta_{s,1})] = \theta_s^c. 
    \end{equation*}
    \item The input data process $\{\zeta_{s,\ell}\}_{\ell=1}^\infty$ are independent for different $s\in\mathcal{S}$. 
\end{enumerate}
\end{assumption}

Assumption \ref{assump: unbiased estimator}.1 assumes the input data for different input distributions are independent. For example, in a queuing system, the customer inter-arrival time and the service time are independent. For a fixed input distribution $s$, $\zeta_{s,\ell}$ can be a vector with correlated components. Assumption \ref{assump: unbiased estimator}.2 can be satisfied when the moments of a distribution serve as sufficient statistics for that distribution, enabling the use of moment estimation methods (e.g., see \cite{bowman2004estimation}). This is valid for most of the common distributions such as normal distribution, exponential distribution, Poisson distribution, and Gamma distribution. For example, if \(\zeta\) follows a normal distribution with mean \(\mu\) and variance \(\sigma^2\), we can parametrize the normal distribution as \(\theta^c = (\mu, \sigma^2 + \mu^2)\), and let \(D(\zeta) = (\zeta, \zeta^2)\), which satisfies \(\mathbb{E}[D(\zeta)] = \theta^c\). Then, with an estimated input parameter \(\hat{\theta}\), the input distribution is updated, and simulations are run for each design \(i\) to generate \(r\) samples, \(X^{1}_i(\hat{\mathbf{\theta}}), \ldots, X^{r}_i(\hat{\mathbf{\theta}})\), for estimating its expected performance. 
% {\color{red} Can delete this if you want. Please see the response to Comment 1 of Reviewer 2.  For more general cases, assume an input distribution \( F_s^c \) belongs to an exponential family of distributions, with density function \( f_s^c \) expressed as:
% \[
% f_s^c(\zeta_s) = f(\zeta_s \mid \eta^c_s) = h(\zeta_s) \exp(\eta_s^c T(\zeta_s) - A(\eta_s^c)).
% \]
% It is known that \( T(\zeta_s) \) is the sufficient statistic for \( f_s^c \), and that \( \mathbb{E}_{\zeta_s \sim f_s^c} [T(\zeta_s)] = \nabla A(\eta_s^c) \) (e.g., see \cite{lehmann2006theory}). After \( \nabla A(\eta_s^c) \) is known, we can solve for \( \eta_s^c \) 
% % (depending on the specific form of \( A(\eta_s^c) \)) 
% and thereby recover the entire distribution. In this context, we define \( \theta_s = \nabla A(\eta_s^c) \) as the input parameter and \( D_s(\cdot) = T(\cdot) \). }

In addition, an advantage resulted from Assumption \ref{assump: unbiased estimator}.2 is that we can use a sample-average estimator to estimate the unknown input parameter. We will show in Section \ref{sec: simulation budget} that the simple form of sample-average estimator is crucial in characterizing the asymptotic convergence of the design's performance estimator under the setting of sequentially collected input data.

\subsection{A Multi-Stage Framework for Simultaneous Resource Allocation with Different Types of Input Data} \label{sec: unified framework}

Recall that there are \(S\) input distributions \(F_1^c, \ldots, F_S^c\). With a slight abuse of notation, we use \(\mathcal{S}\) to denote the set of all input distributions: \(\mathcal{S} = \{1, 2, \ldots, S\}\). As motivated by the examples from Alibaba and drug design, input data collection for different input distributions may share different types of budgets and can be collected simultaneously. To accommodate this flexibility, let \(\{\mathcal{S}_j\}_{j=1}^D\) be a partition of \(\mathcal{S}\). That is, \(\mathcal{S} = \bigcup_{j=1}^D \mathcal{S}_j\) and \(\mathcal{S}_j \cap \mathcal{S}_{j'} = \emptyset\) for \(j \neq j'\). Within a partition \(\mathcal{S}_j\), the data collection for distributions in \(\mathcal{S}_j\) shares the same budget.

Suppose collecting one data point from an input distribution \(s \in \mathcal{S}\) incurs the cost \(c_s\), and generating one simulation output sample for design \(i\) incurs the cost \(d_i\). Notably, the values \(c_s\) should have similar magnitudes only for distributions that share the same budget (i.e., for \(s \in \mathcal{S}_j\) for some \(j\)). The simulation cost can be measured by the computing time required to generate one simulation output. The data collection cost \(c_s\) can be measured by the average time required to collect one data point (e.g., the total time spent surveying divided by the number of surveys when collecting data for consumption patterns) or the price paid for hiring specialists to collect input data. Regardless of the cost measurement, both collecting input data and running simulations are time-consuming. When a decision is required within a certain period, it is crucial to simultaneously conduct both input data collection and simulations to fully utilize the available time. If simulations are not run until all input data are collected, computing resources are wasted during the data collection phase, and the remaining time before the decision deadline may be insufficient for evaluating designs through simulations. Moreover, for input data such as daily customer demand in an inventory control example, which arrive in a streaming fashion, it is beneficial to sequentially adjust the simulation model with the newly arrived data to improve the estimation of the input distribution. Motivated by this, we consider a multi-stage framework where the simulation model is updated with new input data from stage to stage. Within each stage, we simultaneously collect input data for different input distributions and run simulations.

To be specific, let \(T\) be the total number of stages. At the beginning of each stage \(t \in \{1,2,\ldots,T\}\), input data collected during the previous stage arrives in batches and is used to update the estimation of input distributions. Within each stage, the stage-wise budget \(U_j\) is allocated to collect input data for input distributions in \(\mathcal{S}_j\) and \(M\) is allocated to run simulations. In practice, the total number of stages \(T\) represents the ``time due" before the final selection needs to be made. The stage-wise simulation budget \(M\) is determined by the length of one stage, and the input budget \(U_{j}, j=1,\ldots,D\) is determined jointly by the stage length and the corresponding resource limit. For example, when collecting input data on potential customer demands for different products, the total number of surveys that can be conducted within a stage (e.g., one day) is limited by the number of specialists and their available working hours. 
For simplicity, we assume a constant stage-wise budget. However, our proposed algorithm in Section \ref{sec: parameter and procedure} can be directly applied with varying stage-wise budgets.

We then have the following stage-wise budget constraint over running simulation and input data collection at stage $t$: 
\begin{align}
    &    \sum_{i\in\mathcal{K}} d_i m_{i,t}  \le  M \label{eq: sim budget}\\
    & \sum_{s\in \mathcal{S}_j} c_s n_{s,t} \le U_j, ~~ j=1,\ldots,D, \label{eq: input budget}
\end{align}
where $ m_{i,t},  n_{s,t}$ are stage-wise allocation policies for simulation and input data collection, respectively, at stage $t$. At the beginning of each stage $t$, one first updates the input distribution and expected performance using the input data and simulation outputs collected from the previous stage. Then, one computes a stage-wise allocation policy $\{m_{i,t}\}_{i\in\mathcal{K}}$ and $\{n_{s,t}\}_{s\in \mathcal{S}}$ to allocate the stage-wise budget  $M$ and $\{U_j\}_{j=1}^D$  to run the simulation for designs and collect input data, respectively.  

Our problem framework generalizes several previous works by encompassing them as special cases, which are detailed as follows. \vspace{-3mm}
\begin{itemize}
\item If \(|\mathcal{S}_j|=1\) for some \(j\), then to reduce the input uncertainty (uncertainty in the input distribution) of the single input distribution \(s \in \mathcal{S}_j\) to a minimum, one should exhaust the budget \(U_j\) to collect as much input data for \(s\) as possible. This results in \(n_{s,t} = \frac{U_j}{c_s}\) (ignoring the issue of it not being an integer). In this case, taking \(c_s = 1\), we can view \(U_j\) as the batch size of batched input data that comes periodically. When \(|\mathcal{S}_j|=1\) for all \(j=1,\ldots,D\), all input data can be regarded as given input data that arrive periodically. This reduces to the same setting as \cite{wang2022fixed} and \cite{wu2022data}, assuming one cannot control the amount of input data but can only receive the batched streaming input data. Our formulation accommodates this specific setting and adds the flexibility to allow active collection of input data that may consume different budgets.
Additionally, there is a major difference in budget allocation compared with \cite{wang2022fixed}. In \cite{wang2022fixed}, they assume finite support of the input distribution, and they conduct simulations on a fixed pair of design and input realization each time.  When the number of points in the support is large, the number of design-input pair to simulate can be very large as it equals the number of designs multiplied by the number of points in the support. In contrast, we allocate the simulation budget only to different designs. Moreover, the assumption of finite support in \cite{wang2022fixed} limits its practical value, while we consider the more general continuous input distribution. In \cite{wu2022data}, they adopted a fixed confidence formulation, whereas we adopt a fixed budget formulation.

    \item If \(D=1\), then the problem reduces to the conference version \cite{wang2023wsc}, where two budgets (one for input data collection and one for simulation) are simultaneously allocated to collect input data and run simulations. The formulation in this paper is a generalization of its conference version to accommodate multiple budgets for input data collection. Moreover, as discussed above, this general formulation allows for the existence of given streaming input data, such as transit lead time, which further broadens its application domains. 
\end{itemize}

\section{Rate Optimization} \label{sec: rate optimization}

In this section, we discuss how to optimize the simulation budget allocation and input budget allocation. Ideally, we want to maximize the PCS by jointly optimizing the simulation budget allocation and input budget allocation with the budget constraints given in \eqref{eq: sim budget} and \eqref{eq: input budget}. However, the complexity of joint optimization over simulation and input data collection imposes great technical challenges. To overcome these challenges, 
we formulate two optimization problems to determine the allocation policy for data collection budgets and simulation budget,  organized as follows. In Section \ref{sec: input budget}, we formulate an input budget allocation problem with the objective of maximizing the asymptotic convergence rate of the probability of acceptable estimation (PAE). PAE is defined similarly to PCS but focuses solely on the estimation error in the input parameters. 
% The rationale for focusing exclusively on input budget allocation is that input data does not depend on simulation outputs. This separation simplifies the problem, making it more tractable to solve.
Subsequently, in Section \ref{sec: simulation budget}, we formulate a simulation budget allocation problem aimed at maximizing the asymptotic convergence rate of PCS, given the input budget allocation determined by the previous problem. 
% By isolating simulation budget allocation, the problem becomes more manageable. 
It is important to note that PCS is influenced by the input budget allocation. 
This two-step formulation simplifies problem of jointly optimizing both input and simulation budgets. At the same time, the dependence of simulation budget allocation on input budget allocation captures the relationship between simulation outputs and the input distribution.

% This approach is taken because, for input budget allocation, the input data are collected from the real system and do not depend on the simulation outputs. Moreover, the cost of collecting input data is usually far more expensive than the cost of running simulations. Consequently, we focus solely on input uncertainty and assume no simulation error when allocating the input budget.
% For simulation budget allocation, however, we consider maximizing the PCS affected by both simulation and input uncertainty instead of merely simulation uncertainty because the simulation output depends on the input estimation.

\subsection{Input Budget Allocation} \label{sec: input budget}
We begin with input budget allocation.  One reason for solely considering the input budget allocation here is that the estimation of input distribution does not depend on the simulation output. In addition, this avoids the technical difficulty caused by joint optimizing both input budget and simulation budget allocation. 
Let $N_{s,t} = \sum_{\tau=1}^t n_{s,\tau}$ (i.e., $c_s N_{s,t}$ is the total budget assigned to input distribution $s$ up to stage $t$). The input parameter estimator at the beginning of stage $t+1$ is $\widehat{\mathbf{\theta}}_{t}$ with $\widehat{\mathbf{\theta}}_{s,t} = \frac{1}{N_{s,t}} \sum_{r=1}^{N_{s,t}} D_s(\zeta_{s,r})$. Let \(\mu_i(\widehat{\theta})\) denote the expected performance of design \(i\) under input parameter \(\widehat{\theta} = (\widehat{\theta}_s)_{s \in \mathcal{S}}\).  An estimator \(\widehat{\theta}\) is said to be acceptable if
\[
\mu_b(\widehat{\theta}) > \mu_i(\widehat{\theta}), \forall i \neq b,
\]
where \(b := \arg\max_{i \in \mathcal{K}} \mu_i(\theta^c)\) and \(\theta^c := (\theta_s^c)_{s \in \mathcal{S}}\) is the true best design. That is, the true optimal design is still optimal under an acceptable estimated input parameter $\widehat{\theta}$.  Then, we define the following quantity of ``Probability of Acceptable Estimation" (PAE),
$$\operatorname{PAE} = \mathbb{P}\left(\mu_b(\widehat{\theta}) > \mu_i(\widehat{\theta}), \forall i \neq b \right).$$
The simulation output helps identify the true unknown optimal design only when the estimator $\widehat{\theta}$ is acceptable. Otherwise, if the estimation is not acceptable, there is no way to find the true best design by only running more simulations. Consequently, we determine the optimal budget allocation rules for active input data collections through optimizing PAE. The exact characterization of PAE with finite samples is difficult. Instead, we study the asymptotic convergence rate of PAE. For this purpose, we first derive the asymptotic normality for \(\mu_i(\widehat{\theta}) - \mu_j(\widehat{\theta}), \forall i, j \in \mathcal{K}\).
We make the following mild assumptions for regularity.

\begin{assumption} \label{assump: CLT active}
For all $i \in \mathcal{K}$ and $s \in \mathcal{S}$,
\quad
\begin{enumerate}
\item
$\Sigma_{D,s} := \operatorname{Cov}(D_{s}(\zeta_{s,1}))$ exists.

\item
$\mu_i(\cdot)$ and $\nabla \mu_i(\cdot)$ are continuously differentiable and bounded in the compact parameter space $\Theta$.
% \item[(iv)] For each $s\in \mathcal{S}_g$, $\{n_{s,t}\}$ are uniformly bounded and there exists $\bar{n}_s>0$ such that $\lim_{t\rightarrow \infty} \frac{N_{s,t}}{t} = \bar{n}_s$ almost surely. 
\item The optimal design ${b}$ is unique.
% \item[(iv)]
% $\{n_s(t)\}$ and $\{m_i(t)\}$ are uniformly bounded. Furthermore, there exist positive constants $\bar{n}_s$ and $\bar{m}_i$ such that $N_s(t)/t \rightarrow \bar{n}_s$ and $M_i(t) / t \rightarrow \bar{m}_i$ as $t\rightarrow \infty$.
\end{enumerate}
\end{assumption}
\begin{lemma} \label{lem: CLT input}
Under Assumption \ref{assump: CLT active} and
supposing $\{n_{s,t}\}_{s\in\mathcal{S},t\ge1}$ are uniformly bounded and for each $s\in \mathcal{S}$, there exists $\Bar{n}_s >0$ such that 
$ \lim_{t\rightarrow \infty} \frac{N_{s,t}}{t} = \Bar{n}_s$. Then,
\begin{equation}
    \sqrt{t} \left[\mu_i(\widehat{\theta}_t) - \mu_k(\widehat{\theta}_t) - \left( \mu_i(\theta^{c}) - \mu_k(\theta^{c})\right)  \right] \Rightarrow \mathcal{N}\left(\mathbf{0}, \Bar{\sigma}_{ik}^2 \right),
\end{equation}
where 
$$\Bar{\sigma}_{ik}^2 = \sum_{s\in\mathcal{S}} \frac{1}{\Bar{n}_s } \partial_{\theta_s} \delta_{ik}^\top \Sigma_{D,s} \partial_{\theta_s} \delta_{ik}, ~~ \delta_{ik} = \mu_i(\theta ^{c}) - \mu_k(\theta^{c}).$$
\end{lemma}
 The proof can be found in Appendix \ref{appsec: Lemma CLT input}. Notice the PAE at stage $t$ with estimator $\widehat{\theta} _t$ satisfies $\forall i \neq {b}$,
\begin{equation*}
   1 - \mathbb{P}\left(\mu_{{b}}(\widehat{\theta} ) \le \mu_{i}(\widehat{\theta} ) \right) \ge  \operatorname{PAE} \ge 1 - \sum_{i\neq b}\mathbb{P}\left(\mu_{{b}}(\widehat{\theta} ) \le \mu_{i}(\widehat{\theta} ) \right).
\end{equation*}
Using the asymptotic normality in Lemma \ref{lem: CLT input}, with a similar derivation as in \cite{wang2023wsc}, we have approximately 
$$ \lim_{t\rightarrow\infty} \frac{1}{t} \log \left( 1 -\operatorname{PAE} \right) = -\min_{i\neq b} \frac{\delta_{bi}^2}{2\Bar{\sigma}^2_{bi}}.$$

We can then determine the allocation policy of input budgets by maximizing the convergence rate of PAE:
\begin{equation} \label{eq: PAE optimize}
    \max_{\bar{n}\ge 0} \min_{i\neq b} \frac{\delta_{bi}^2}{\Bar{\sigma}^2_{bi}}~ s.t. \left\{\sum_{s\in\mathcal{S}_j} c_s \Bar{n}_s = U_j,~j=1,\ldots,D\right\}.
\end{equation}
This is equivalent to solve:
\begin{equation} \label{eq: input allocation}
    \begin{aligned}
    \max_{\Bar{n}\ge0, z} ~~&z\\
    s.t.~~ & \frac{\delta_{bi}^2}{\Bar{\sigma}^2_{bi}} \ge z, ~~i\neq b\\
    &\sum_{s\in\mathcal{S}_j} c_s \Bar{n}_s = U_j,~~ j=1,\ldots,D.
\end{aligned}
\end{equation}
We make the following assumption on the regularization of the problem. 

\begin{assumption} (Impact of input uncertainty) \label{assump:input effect}\\
        For each $s \in \mathcal{S}$, there exists $i\neq b$, $\partial_{\theta_s}\delta_{bi}(\theta) \neq 0$  for $\theta = \theta^c$ and  almost every $\theta \in \Theta$, where $\delta_{ik}(\theta) = \mu_i(\theta) - \mu_k(\theta), i,k\in\mathcal{K}$.
  \end{assumption}
Assumption \ref{assump:input effect} guarantees that, each input distribution $s$ has an impact on distinguishing at least $1$ sub-optimal design from the optimal design. Here the requirement of $\delta_{bi} \neq 0$ for ``almost every" $\theta$ ensures that when we plug in the estimator $\widehat{\theta}$, with probability 1, $\partial_{\theta_s} \delta_{b i}(\widehat{\theta}) \neq 0$. This implies, the input distribution $s$ has an impact on estimating $\delta_{bi}$,  and consequently, more input data for input distribution $s$ will be collected. This assumption is usually satisfied because the input distribution has different impacts on the output performance of different designs. With Assumption \ref{assump:input effect}, we have the following Lemma \ref{lem: convex input rate optimization}, whose proof can be found in Appendix \ref{appsec: lem convex input rate optimization}.
\begin{lemma}
     \label{lem: convex input rate optimization}
    Suppose Assumptions \ref{assump: parametric input}-\ref{assump:input effect} hold. Then the PAE rate optimization problem \eqref{eq: input allocation} is a concave optimization problem with positive optimal solutions (but not necessarily unique).
\end{lemma}
Lemma \ref{lem: convex input rate optimization} indicates we can solve Problem \eqref{eq: input allocation} numerically using convex optimization solvers. The existence of multiple optimal solutions come from the fact that Problem \eqref{eq: PAE optimize} (the equivalent form of \eqref{eq: input allocation}) is not strictly concave. Nonetheless, if $\partial_{\theta_s} \delta_{bi} \neq 0, \forall s \in\mathcal{S}, i\neq b$, then \eqref{eq: PAE optimize} becomes a strictly concave problem with a unique optimal solution. This condition holds if each input distribution has an impact on all designs.

% Theorem \ref{thm: convex input rate optimization} guarantees that Problem \ref{eq: input allocation} is a concave problem, and hence can be solved by standard optimization tools such as gradient descent. Furthermore, we show that \eqref{eq: input allocation} admits a unique optimal solution and this optimal solution is strictly positive. The uniqueness is crucial for the performance analysis in Section \ref{sec: consistency and optimality} and the proof of uniqueness is not trivial. The uniqueness of the optimal solution is usually guaranteed if the problem is strict concave, which does not hold for \eqref{eq: input allocation} (or \eqref{eq: PAE optimize}) if $\partial_{\theta_s} \delta_{bi} = 0$ for some $s,i\neq b$. This occurs if input distribution $s$ is some design-specific input distribution for some sub-optimal design $k$, in which case $\partial_{\theta_s} \delta_{bi} = 0, i\neq b\neq k $. We prove the uniqueness of \eqref{eq: PAE optimize} by showing the strict concavity of \eqref{eq: PAE optimize} when the domain is constrained to the set of all its optimal solutions, with extra analysis on the objective function.
\subsection{Simulation Budget Allocation}
\label{sec: simulation budget}
In this section, we consider the simulation budget allocation.
Let  $M_{i,t} = \sum_{\tau=1}^t m_{i,\tau}$, (i.e., $d_i M_{i,t}$ is the total budget assigned to design $i$ up to stage $t$). Recall at stage $t+1$, the simulation is run under $\widehat{\theta}_{t}$ to get $X_i^1(\widehat{\theta}_{t}),X_i^2(\widehat{\theta}_{t}),\ldots$, which are i.i.d. samples  of the random performance of the $i^{th}$ design conditioned on input parameter $\widehat{\theta}_t$. The performance estimator $\widehat{\mu}_{i,t}$ is defined as 
\begin{equation} \label{eq:mae}
    \hat{\mu}_{i,t} := \frac{1}{M_{i,t}} \sum_{\ell=1}^t \sum_{r=1}^{m_{i,\ell}} X^r_{i}(\hat{\theta}_\ell).
\end{equation}

Let $\hat{b}_t = \arg\max_i \hat{\mu}_{i,t}$ denote the estimated best design at the end of  stage $t$, that is, the design with the largest estimated expected performance. 
Consequently, we can define the  probability of correct selection (PCS), which measures the quality of the selection, at stage $t$ as
$$ \text{PCS} = \mathbb{P}(\hat{b}_t = b) = \mathbb{P}(\hat{\mu}_{b,t} \ge \hat{\mu}_{i,t} ,\ \forall i\neq b). $$
Similar as for input budget allocation, we will analyze the asymptotic property of PCS by characterizing the asymptotic normality of the pairwise performance difference estimator $\widehat{\delta}_{ik,t} := \widehat{\mu}_{i,t} - \widehat{\mu}_{k,t}$. 
Unlike classic R\&S, here a main challenge of studying the asymptotic normality comes from the correlated simulation samples $\{X_i^r(\widehat{\theta}_\ell)\}$ because $\{\widehat{\theta}_\ell\}$ are correlated for different $\ell$. Nonetheless, thanks to the sample-average expression of $\widehat{\theta}_\ell$, we are able to characterize such correlation between $\widehat{\theta}_\ell$ and $\widehat{\theta}_{\ell'}, \ell \neq \ell'$, which is crucial in proving the asymptotic normality of $\widehat{\delta}_{ik,t}$.

We make the following assumption of regularity conditions on the simulation outputs.

\begin{assumption} \label{assump:CLT}
For all $i \in \mathcal{K}$ and $s \in \{1,2,\ldots, S\}$,
\quad
\begin{enumerate}

\item
$\sigma_i(\theta)$ exists and is continuous in $\theta$ for all $\theta \in \Theta$, where $\sigma^2_{i}(\theta)$ is the variance of $X_i(\theta)$ conditioned on $\theta$.
\item For any given $\theta$, the simulator can generate i.i.d. samples $X^\ell_i(\theta), \ell=1,2,\ldots$ for design $i$, where $X^\ell_i(\theta)$ has mean $\mu_i(\theta)$ and variance $\sigma^2_i(\theta) \le \sigma^2$. 
\end{enumerate}
\end{assumption}

Let $\sigma_i = \sigma_i(\theta^c)$ for simplicity. The next theorem establishes the asymptotic normality of the performance estimator $\widehat{\mu}_{i,t}$, which is a variant of Theorem 3 in \cite{wu2022data}. The proof can be found in Appendix \ref{appsec: thm normality}.
\begin{theorem} \label{thm: normality}
Suppose $\{n_{s,t}\}$ and $\{m_{i,t}\}$ are uniformly bounded. Furthermore, there exist positive constants $\bar{n}_s$ and $\bar{m}_i$ such that $N_{s,t}/t \rightarrow \bar{n}_s$ and $M_{i,t} / t \rightarrow \bar{m}_i$ as $t\rightarrow \infty$ almost surely. Then, 
\begin{equation*}
\sqrt{t} \left[\widehat{\delta}_{ik,t} - \delta_{ik}\right] \Rightarrow \mathcal{N}(0,\Tilde{\sigma}^2_{ik}), \quad \text{as } t\rightarrow \infty,
\end{equation*}
where $ \Rightarrow$ means convergence in distribution, 
$$ \Tilde{\sigma}^2_{ik} = 2 \sum_{s \in \mathcal{S}}\bar{n}_s^{-1} \partial_{\theta_s} \delta_{ik}^\top \Sigma_{D,s}\partial_{\theta_s} \delta_{ik}   +   \bar{m}_i^{-1} \sigma_i^2 +   \bar{m}_k^{-1}\sigma^2_k,$$
and $\partial_{\theta_s}$ denotes the partial derivative taken with respect to $\theta_s$.
\end{theorem}

 Then, with the normal approximation by Lemma \ref{thm: normality} and a similar argument as for PAE, we obtain
\begin{equation} \label{eq: exponential decay rate}
    -\lim_{t\rightarrow\infty} \frac{1}{t}\log{(1-\operatorname{PCS})} = \min_{i\neq b} \frac{1}{2} \frac{\delta^2_{bi} }{\Tilde{\sigma}_{bi}^2},
\end{equation}
We refer to \eqref{eq: exponential decay rate} the asymptotic exponential convergence rate of PCS.
 Then, we determine the stage-wise simulation budget allocation policy by maximizing this convergence rate. 
 
  \begin{equation}\label{eq: simulation allocation}
  \begin{aligned}
      \max_{\Bar{m}_i \ge 0, z}\quad & z \\% =\mathbb{P} (\widehat{\mu}_b(\widehat{\theta}) \ge \widehat{\mu}_i(\widehat{\theta}), i\neq b)\\
       s.t. \quad&    \frac{\delta^2_{bi} }{\Tilde{\sigma}_{bi}^2} \ge z  \qquad \forall i \neq b\\ 
        & \sum_{i\in\mathcal{K}} d_i \Bar{m}_i = M  \end{aligned},      
  \end{equation} 
where $\Tilde{\sigma}_{bi}$ are defined in Lemma \ref{thm: normality}. Theorem \ref{thm: optimality conditions} below characterizes the optimal solution of \eqref{eq: simulation allocation}, whose proof is in Appendix \ref{appsec: thm optimality conditions}.

\begin{theorem} \textbf{(Optimality Conditions.)} \label{thm: optimality conditions} Under Assumption \ref{assump: parametric input}-\ref{assump:CLT}, given $\Bar{n}>0$, a solution $\Bar{m}$ is optimal to \eqref{eq: simulation allocation} if and only if it satisfies the following conditions.
\begin{flalign}
    & (\textbf{Rate Balance}) \ \frac{\delta_{bi}^2}{  2 \sum_{s \in \mathcal{S}} \Bar{n}_s^{-1} g(i,s)   +   \bar{m}_i^{-1} \sigma_i^2 +   \bar{m}_b^{-1}\sigma^2_b} = \frac{\delta_{bk}^2}{  2 \sum_{s \in \mathcal{S}} \Bar{n}_s^{-1}g(k,s)   +   \bar{m}_k^{-1} \sigma_k^2 +   \bar{m}_b^{-1}\sigma^2_b} && \label{thmeq: rate balance}\\
 &  \hskip 5in \forall i\neq k\neq b && \notag
    \\
    & (\textbf{Global balance}) \ \bar{m}_b^2 = \frac{\sigma^2_{b}}{d_b} \sum_{i\neq b} \frac{d_i\bar{m}_i^2}{\sigma^2_i} && \label{thmeq: global balance}
    % &(\textbf{Input Data Balance}) %{\color{red} Spell it out? or maybe call it Input Data Balance}
    % \ \bar{n}_s^2 = \frac{ 2}{c_s } \sum_{i\neq b} \frac{d_i\bar{m}_i^2}{\sigma_i^2(\mathbf{\theta^c})} g(i,s)\quad \forall s \in \mathcal{S}_a && \label{thmeq: Input Data Balance}
    % &(\textbf{Input Data Balance-Disjoint})\ \frac{1}{c_s\bar{n}_s^2} \sum_{i\neq b} \frac{d_i\bar{m}_i^2}{\sigma_i^2(\mathbf{\theta^c})} g(i,s) =    \frac{1}{c_{s'}\bar{n}_{s'}^2} \sum_{i\neq b} \frac{d_i\bar{m}_i^2}{\sigma_i^2(\mathbf{\theta^c})} g(i,s') \quad \forall s\neq s' \in S_j, \quad j=1,\ldots,D && \label{thmeq: Input Data Balance disjoint}
\end{flalign}
where  $g(i,s) = \partial_{\theta_s}\delta_{bi}^\top \Sigma_{D,s}\partial_{\theta_s}\delta_{bi}$.
\end{theorem}

Notably, in classical Ranking and Selection (R\&S) without input uncertainty, the ``Rate Balance" and ``Global Balance" optimality conditions were derived similarly to previous works \cite{glynn2004large, chen2022balancing}. Here, with the existence of input uncertainty, a crucial distinction is that the rate function in the ``Rate Balance" condition now includes an extra term \(2 \sum_{s \in \mathcal{S}} \Bar{n}_s^{-1} g(i,s)\), which characterizes the impact of input uncertainty caused by random input data from different input distributions. Consequently, the optimal simulation budget allocation rule depends on this impact of input uncertainty. 
% In addition, we also generalize the assumption on the simulation costs, which are usually assumed to be the same for different designs, allowing different costs (e.g., different simulating time) for simulating different designs.

\subsection{Remark on Input Budget Allocation and Simulation Budget Allocation}
In Section \ref{sec: input budget}, when allocating the input budget, we maximize the convergence rate of PAE, which only contains the input uncertainty. Then, in Section \ref{sec: simulation budget}, to allocate the simulation budget, we maximize the convergence rate of PCS, which includes both input uncertainty and simulation uncertainty (uncertainty caused by random simulation outputs).
This approach is taken because, for input budget allocation, the input data are collected from the real system and do not depend on the simulation outputs. Moreover, the cost of collecting input data is usually far more expensive than the cost of running simulations. Consequently, we focus solely on input uncertainty and assume no simulation error (that is, the value $\mu_i(\theta)$ is assumed to be known given $\theta$) when allocating the input budget.
For simulation budget allocation, however, we consider maximizing the PCS affected by both simulation and input uncertainty instead of merely simulation uncertainty because the simulation output depends on the input estimation.

\section{Multi-Stage Simultaneous Budget Allocation Procedure} \label{sec: parameter and procedure}
In the previous section, to determine the allocation policy for input budgets and simulation budgets, we formulated two rate optimization problems, \eqref{eq: input allocation} and \eqref{eq: simulation allocation}. These problems aim to maximize the convergence rate of PAE over the input budget allocation policy and to maximize the convergence rate of PCS over the simulation budget allocation policy, respectively. However, it remains unsolved how to design an implementable algorithm from these optimization problems due to the following two reasons.

First, both \eqref{eq: input allocation} and \eqref{eq: simulation allocation} contain several unknown parameters that need to be estimated for practical implementation. We will replace these unknown parameters with their estimators constructed using previously collected input data and simulation outputs. Additionally, these estimators will be updated from stage to stage with new input data and simulation outputs to reduce estimation error. The construction of the estimators will be discussed in Section \ref{sec: parameter estimation}.

Second, after replacing the unknown parameters with their estimators, it remains unclear whether we should solve both \eqref{eq: input allocation} and \eqref{eq: simulation allocation} to optimality to obtain a stage-wise budget allocation policy. Notably, because we will periodically update the estimators of unknown parameters in \eqref{eq: input allocation} and \eqref{eq: simulation allocation}, we need to resolve the two optimization problems at the beginning of each stage with the updated estimators, which imposes a computational challenge.

For input budget allocation, since the optimization cost is usually negligible compared to the cost of input data collection, we will solve \eqref{eq: input allocation} to optimality. However, for simulation budget allocation, we cannot afford to solve \eqref{eq: simulation allocation} to optimality because both optimization and simulation consume computing resources, and the cost of optimization is often not negligible compared to the simulation cost. 
% Solving \eqref{eq: simulation allocation} to optimality can prevent running extra simulations as it takes extra computing time, which can impair the algorithm's performance. 
Moreover, \eqref{eq: input allocation} is usually much easier to solve than \eqref{eq: simulation allocation} since the number of input distributions is typically much smaller than the number of designs.
To address the computational issue of optimizing \eqref{eq: simulation allocation}, in Section \ref{sec: simulation algorithm}, we design an algorithm that sequentially allocates the simulation budget based on the evaluation of the optimality conditions given by Theorem \ref{thm: optimality conditions}.

\subsection{Parameter Estimation} \label{sec: parameter estimation}
To design an algorithm, several unknown parameters need to be estimated. They include
\begin{enumerate}
    \item The true input parameter $\mathbf{\theta^c}$ and its covariance matrix $\Sigma_{D,s}$ for $s=1,2,\ldots,S$.
    \item  The true expected performance $\mu_i(\mathbf{\theta^c})$ and variance $\sigma_i^2(\mathbf{\theta^c})$.
    \item  The gradient $\nabla \mu_i(\mathbf{\theta^c})  = (\partial_{\theta_1} \mu_i(\mathbf{\theta^c}),\partial_{\theta_2} \mu_i(\mathbf{\theta^c}),\ldots,\partial_{\theta_S} \mu_i(\mathbf{\theta^c}))^\top.$
\end{enumerate}

To estimate $\mathbf{\theta^c}$ and $\Sigma_{D,s}$, by Assumption \ref{assump: parametric input} and \ref{assump: unbiased estimator}, we can use the sample average and sample variance, respectively. That is, let $\widehat{{\theta}}_{s,t} = \frac{1}{N_{s,t}}\sum_{\ell=1}^{N_{s,t}} D_s(\zeta_{s,\ell})$ and $\widehat{\Sigma}_{D,s,t} = \frac{1}{N_{s,t}-1} \sum_{\ell=1}^{N_{s,t}} (D_s(\zeta_{s,\ell}) - \widehat{\theta}_{s,t})(D_s(\zeta_{s,\ell}) - \widehat{\theta}_{s,t})^\top$. To estimate $\mu_i(\mathbf{\theta^c})$, we use performance estimator defined in \eqref{eq:mae}. Similarly, we estimate $\sigma_i^2(\mathbf{\theta^c})$ by 
\begin{equation*}
    \widehat{\sigma}^2_{i,t} = \frac{1}{M_{i,t}-1} \sum_{\ell =1}^t \sum_{r=1}^{m_{i,\ell}} (X_i^r(\widehat{\mathbf{\theta}}_\ell) - \widehat{\mu}_{i,t})^2.
\end{equation*}

Finally, to estimate $\nabla \mu_i(\mathbf{\theta^c})$, let $\xi$ denote the randomness in generating one simulation output $X_i(\theta)$ and let $q_\theta$ denote the density function of $\xi$. Denote by $ X_i(\theta,\xi)$ the random simulation output generated under $\xi \sim q_\theta$. Then we have,
$\mu_i(\theta) = \mathbb{E}_{\xi \sim q_\theta}[X_i(\theta)] = \mathbb{E}_{\xi \sim q_\theta}[X_i(\theta,\xi)], \forall \theta\in \Theta$. Suppose we have access to $q_\theta$ and the gradient $\nabla_\theta q_\theta(\xi)$.
Suppose $\forall \mathbf{\theta} \in \Theta$, $q_\mathbf{\theta}$ has the same input support $\Omega$. That is, $\forall \xi \in \Omega$, $q_\mathbf{\theta}(\xi) > 0$ for all $\mathbf{\theta} \in \Theta$. %Since $\xi$ represents the randomness in the simulation, when generating $X_i^r(\widehat{\mathbf{\theta}}_\ell)$, one can first generate $\xi_{i,\ell}^r \sim f_{\widehat{\mathbf{\theta}}_\ell}$, then runs the simulation under $\xi_{i,\ell}^r $ and obtains the simulation output $X_i(\xi_{i,\ell}^r) = X_i^r(\widehat{\mathbf{\theta}}_\ell)$. 
Since
\begin{equation*}
    \nabla_{\mathbf{\theta}} \mu_i(\mathbf{\theta^c}) = \nabla_{\mathbf{\theta}} \mathbb{E}_{\xi\sim q_{\mathbf{\theta^c}}} [X_i(\xi)] =\nabla_{\mathbf{\theta}}\int_{\xi\in\Omega} q_{\mathbf{\theta^c}}(\xi) X_i(\theta,\xi) \mathrm{d}\xi,
\end{equation*} 
assuming the interchangeability of the integration and the gradient, we have
\begin{equation*}
    \nabla_{\mathbf{\theta}} \mu_i(\mathbf{\theta^c}) =\int_{\xi\in\Omega} \nabla_{\mathbf{\theta}}q_{\mathbf{\theta^c}}(\xi) X_i(\theta,\xi) \mathrm{d}\xi =  \mathbb{E}_{\xi\sim q_{\mathbf{\theta}}} \left[  \frac{\nabla_{\mathbf{\theta}}q_\mathbf{\theta^c}(\xi)}{q_\mathbf{\theta}(\xi)} X_i(\theta,\xi)\right], \qquad \forall \mathbf{\theta} \in \Theta
\end{equation*} 
An analogy can be drawn to design the following estimator by replacing $\mathbf{\theta^c}$ and $\theta$ with $\widehat{\mathbf{\theta}}_\ell$ for $\ell = 1,\ldots, t$.
\begin{equation*}
    \widehat{\nabla \mu}_{i,t} = \frac{1}{M_{i,t}}\sum_{\ell=1}^t\sum_{r=1}^{m_{i,\ell}}  \frac{\nabla q_{\widehat{\mathbf{\theta}}_\ell}(\xi_{i,\ell}^r)}{q_{\widehat{\mathbf{\theta}}_\ell}(\xi_{i,\ell}^r)}X_i^r(\widehat{\mathbf{\theta}_\ell}).
\end{equation*}

Denote by \(\widehat{g}_{t}(i,s)\) the estimate of \(g(i,s)\) as defined in Theorem \ref{thm: optimality conditions}, with the unknown parameters replaced by their estimates. It is worth noting that to estimate the unknown parameters, we use all the simulation outputs and do not run any extra simulations for the purpose of efficient sampling. In addition, while we provide specific estimators, any other estimators can be directly applied to Algorithm \ref{alg: unified}. 

\subsection{Sequential Allocation of Simulation Budget} \label{sec: simulation algorithm}
In this section, we design a sequential algorithm to allocate the simulation budget at each stage without solving \eqref{eq: simulation allocation} to optimality, inspired by the so-called ``Balancing" approach in \cite{chen2022balancing}. ``Balancing" means reducing the gap between the two sides of the optimality equations. 

To see how this works, recall that \(\Bar{n}_s\) and \(\Bar{m}_i\) represent the average data batch size as defined in Theorem \ref{thm: normality}, and \(N_{s,t}\) and \(M_{i,t}\) are the total amount of data up to stage \(t\). At stage \(t+1\), we substitute \(\Bar{n}_s\) and \(\Bar{m}_i\) with \(N_{s,t}/t\) and \(M_{i,t}/t\) in \eqref{thmeq: rate balance} and \eqref{thmeq: global balance} and multiply both sides by \(t\). In addition, we replace all unknown parameters with their current estimators as discussed in Section \ref{sec: parameter estimation}. Let \(\widehat{b}_t := \max_i \widehat{\mu}_{i,t}\) be the current estimated best design.

To determine which design to simulate, we first look at the global balance equation. We will simulate design \(\widehat{b}_t\) if
\begin{equation}\label{eq: global balance evaluate}
M_{\widehat{b}_t,t}^2 <  \widehat{\sigma}^2_{\widehat{b}_t,t} \sum_{i\neq \widehat{b}_t} \frac{M_{i,t}^2}{\widehat{\sigma}^2_{i,t}}, \quad \text{(global balance)}.   
\end{equation}
 By doing this, we increase the left-hand side of \eqref{eq: global balance evaluate}. Otherwise, we pick a sub-optimal design to simulate to increase the right-hand side of \eqref{eq: global balance evaluate}. In either case, we intend to reduce the gap between the two sides of \eqref{eq: global balance evaluate}. Moreover, if we decide to simulate a sub-optimal design, we further choose the design \(i\) that minimizes
\[
\frac{(\widehat{\mu}_{\widehat{b}_t,t} - \widehat{\mu}_{i,t})^2 }{  2 \sum_{s \in \mathcal{S}} N_{s,t}^{-1} \widehat{g}_t(i,s) + M_{i,t}^{-1} \widehat{\sigma}_{i,t}^2 + M_{\widehat{b}_t,t}^{-1} \widehat{\sigma}^2_{\widehat{b}_t,t}} \quad \text{(rate balance)}.
\]
 % \subsection{A Unified Procedure for Multi-stage Simultaneous Budget Allocation} \label{sec: unified procedure}
 
Equipped with Section \ref{sec: simulation algorithm}, we present the detailed algorithm of \textbf{Simultaneous Budget Allocation (SBA) for data collection and simulation} in Algorithm \ref{alg: unified}. 
% Notably, for simultaneous input budgets allocation, we do not directly use the computed optimal solution $\{\widehat{n}_s\}_{s\in\mathcal{S}}$ of \eqref{eq: input allocation} as the stage-wise budget allocation rule but consider the total amount of data collected so far. This is to deal with the issue of $\widehat{n}_s$ not being an integer as well as avoid wasting unused budgets from the previous stage caused by the integral policy.

\begin{algorithm}[h]
   \caption{Simultaneous Budget Allocation for Data Collection and Simulation}
   \label{alg: unified}
\begin{algorithmic}
   \State \textbf{Input:} Set of designs $\mathcal{K}$, set of input distributions $S = \cup_{j=1}^D \mathcal{S}_j$, input data collection cost $c_s, s\in \mathcal{S}$, simulation cost $d_i, i\in \mathcal{K}$, stage-wise budget $M,U_j, j =1\ldots,D$, initial budget $n_0$ and $m_0$, maximal stage $T$.
%   \REPEAT
   \State \textbf{Initialize} Collect $n_0$ input data for each input distribution and run $m_0$ simulation replications for each design. Estimate $\widehat{\mathbf{\theta}}_0$, $\widehat{\Sigma}_{D,s,0}$, $\widehat{\mu}_{i,0}$, $\widehat{\sigma}_{i,0}$ and $\widehat{\nabla \mu}_{i,0}$ as in Section \ref{sec: parameter estimation}.  $\widehat{b} = \arg\max_i \widehat{\mu}_{i,0}$, ${N}_{s,0} = n_0, \forall s \in \mathcal{S}_a$, ${M}_{i,0} =m_0, \forall i$. Set $t=0$.
   \For{$t =1:T$ }
   \State $M_{i,t} = M_{i,t-1}$, $N_{s,t} = N_{s,t-1}$, $\forall  i \in \mathcal{K}$, $ s \in \mathcal{S}$. 
   \State Update $\widehat{\theta}_s$ and $\widehat{\Sigma}_{D,s,t}$, $\widehat{\mu}_{i,t}$, $\widehat{\sigma}_{i,t}$ and $\widehat{\nabla \mu}_{i,t}$ for all $s,i$ with input data and simulation output from stage $t-1$. Set $\widehat{b} = \arg\max_i \widehat{\mu}_{i,t}$.
   % \State Do the following two \textbf{WHILE} loops  for \textbf{Data Collection} and \textbf{Simulation} simultaneously.
   \State Compute $\{\hat{n}_s\}_{s\in\mathcal{S}}$ that optimizes \eqref{eq: input allocation} with unknown parameters replaced with their estimators.
   \For{$j=1:D$ } 
   \While{$\sum_{s \in \mathcal{S}_j} c_s (N_{s,t}-n_0) <t\times U_j $}
   \State $s^* = \arg\max_{s\in\mathcal{S}_j} \{t \times \hat{n}_s - N_{s,t}\}$. 
   \State $N_{s^*,t} = N_{s^*,t}+1$. \hfill \Comment{\textbf{Input Budget Allocation}}
   \EndWhile
   \EndFor
   \While{ $\sum_{i\in\mathcal{K}} d_i (M_{i,t}-m_0) < t \times M \ $}
  \If{$M^2_{\widehat{b},t} -  \frac{\widehat{\sigma}^2_{\widehat{b},t}}{d_b} \sum_{i\neq \widehat{b} } \frac{d_iM^2_{i,t}}{\widehat{\sigma}_{i,t}^2} <0$} 
  \State $M_{\widehat{b},t} = M_{\widehat{b},t} +1$. \hfill \Comment{\textbf{Global balance}} % \hfill \Comment{\textbf{Simulate the optimal design}}
  \Else 
  \State $i^* = \arg\min_{i\neq \widehat{b}} \frac{(\widehat{\mu}_{\widehat{b},t}-\widehat{\mu}_{i,t})^2 }{ 2 \sum_{s \in \mathcal{S}}\frac{ \widehat{g}(i,s)}{N_{s,t}}  +  \frac{\widehat{\sigma}_{i,t}^2}{M_{i,t}}  +  \frac{\widehat{\sigma}^2_{\widehat{b},t}}{M_{\widehat{b},t}}}$. $M_{i^*,t} = M_{i^*,t} +1$.  \hfill \Comment{\textbf{Rate Balance}}
  \EndIf
   \EndWhile 
   \State  Collect $n_{s,t}:=N_{s,t} - N_{s,t-1}$ input data for input distribution $s$ for any $s\in\mathcal{S}$ and generate $m_{i,t} := M_{i,t}-M_{i,t-1}$ simulation outputs for design $i$ for any $i\in\mathcal{K}$. \label{line: alg1 endwhile}
   \EndFor
   \State {\bfseries Output:} $\widehat{b} = \arg\max_i \widehat{\mu}_{i,T}$.
%   \UNTIL{$noChange$ is $true$}

\end{algorithmic}
\end{algorithm}

\section{Consistency and Asymptotic Optimality} \label{sec: consistency and optimality}
In this section, we provide the statistical guarantee, namely the consistency and asymptotic normality, for our proposed algorithm. The consistency result guarantees that the algorithm can select the optimal design almost surely as the number of stages goes to infinity (with fixed stage-wise budgets). The asymptotic optimality further characterizes the convergence of the allocation policy given by the algorithm. Specifically, we prove the allocation policy given by Algorithm~\ref{alg: unified} converges to the optimal allocation rule defined by \eqref{eq: input allocation} and \eqref{eq: simulation allocation}, with a slight approximation which will be detailed in the next subsection. 
% Furthermore, we also characterize the convergence rate of the allocation policy.
\subsection{Consistency}
To prove the consistency result, we make the following assumptions.
\begin{assumption} \label{assump: consistency sigma 
 nable}\ 
\begin{enumerate}
\item  (Interchangeability of gradient and integral) For all $i\in\mathcal{K}$,
\begin{equation*}
    \nabla_{\mathbf{\theta}} \mu_i(\mathbf{\theta^c}) =\int_{\xi\in\Omega} \nabla_{\mathbf{\theta}}q_{\mathbf{\theta^c}}(\xi) X_i(\theta,\xi) \mathrm{d}\xi =  \mathbb{E}_{\xi\sim q_{\mathbf{\theta}}} \left[  \frac{\nabla_{\mathbf{\theta}}q_\mathbf{\theta^c}(\xi)}{q_\mathbf{\theta}(\xi)} X_i(\theta,\xi)\right], \qquad \forall \mathbf{\theta} \in \Theta,
\end{equation*}   
where $X_i(\theta,\xi) = X_i(\theta)$ is defined in Section \ref{sec: parameter estimation}.
\item There exists $\bar{x},\bar{Q}>0$, such that $|X_i(\mathbf{\theta})| \le \bar{x}$ almost surely and $\mathbb{E}_{\xi \sim q_\theta}\left[\left(\frac{\nabla_\theta q_\theta(\xi)}{q_\theta(\xi)}\right)^2 \right] \le \bar{Q} <\infty$  for all $\mathbf{\theta} \in \Theta$ and $1\le i\le K$. \label{assump item: consistency bounded x}
    % \item For each $i \neq i' \in \mathcal{K}$, there exists $s\in \mathcal{S}_a$, $\partial_{\theta_s} \delta_{ii'} \neq 0$ almost everywhere.
    % \item For each $s \in \mathcal{S}_a$ and $i\in \mathcal{K}$, there exists $i'\in\mathcal{K}$, $\partial_{\theta_s} \delta_{ii'}(\theta) \neq 0$ almost everywhere.\label{assump item: consistency partial 2}
    \item $b(\theta):= \arg\max_{i\in\mathcal{K}} \mu_i(\theta)$ is unique for  $\theta = \theta^c$ and almost every $\theta \in \Theta$.
    \label{assump item: unique b}
\end{enumerate}
\end{assumption}
Assumption \ref{assump: consistency sigma nable}.1 allows us to switch the order of gradient and integral, and hence we can adopt the gradient estimator introduced in Section 
\ref{sec: parameter estimation}.  Assumption \ref{assump: consistency sigma nable}.\ref{assump item: consistency bounded x} is a regularity condition that allows us to apply Strong Law of Large Number for martingale difference sequences. The boundedness assumption on the random simulation output can be relaxed to boundedness on the $4^{th}$ moment. Assumption \ref{assump: consistency sigma nable}.3 ensures when we plug in the input estimator $\widehat{\theta}$ at any stage, with probability $1$ there is a unique optimal design $b(\widehat{\theta})$.
We first present the following result on the consistency of the parameter estimates obtained from Algorithm~\ref{alg: unified}. The proof can be found in Appendix \ref{appsec: lem estimate consistency}.
\begin{lemma} \label{lem: estimate consistency}
   Under Assumption  \ref{assump: parametric input}-\ref{assump: consistency sigma nable} {\color{red}}, suppose $\hat{\theta}_t \rightarrow \bar{\theta}$ almost surely, where $\bar{\theta}$ can be random.  Then, if  $M_{i,t} \rightarrow \infty$  as $t \rightarrow \infty$ almost surely for all $1\le i \in K$, the following results hold: (a) $\widehat{\mu}_{i,t} \rightarrow\mu_i({\bar{\theta} })$, (b) $\widehat{\sigma}^2_{i,t} \rightarrow \sigma^2_i ({\bar{\theta} })$ and (c) $\widehat{\nabla \mu}_{i,t} \rightarrow \nabla_\theta \mu_i(\bar{\theta} )$
    %, (c) $\widehat{\nabla \mu}_{i,t} \rightarrow\nabla\mu_i(\mathbf{\theta^c})$ 
    as $t \rightarrow \infty$ almost surely. Moreover, if $N_{s,t} \rightarrow \infty$ almost surely, we have $\bar{\theta} = \theta^c$ almost surely.
\end{lemma}

%{\color{red} Need some assumptions to use SLLN for MDS. One possible choice: $X_i(\theta)$ is bounded for proof of (a) and (b). $\nabla f_{\theta}(\zeta) $ is bounded, $\mathbb{E}_{\zeta\sim f_{\mathbf{\theta}}} \left[ \frac{1}{f^2_{\mathbf{\theta}}(\zeta)}\right]$ is bounded by a positive number for proof of (c).}
With Lemma \ref{lem: estimate consistency},  we can prove the consistency of Algorithm \ref{alg: unified} by proving $N_{s,t} \rightarrow \infty$ and $M_{i,t} \rightarrow \infty$  as $t \rightarrow \infty$, which is stated in the following Theorem \ref{thm: consistency}. The proof can be found in Appendix \ref{appsec: thm consistency}.

\begin{theorem} \textbf{(Consistency)} \label{thm: consistency}
    Under Assumption \ref{assump: parametric input} - \ref{assump: consistency sigma nable}, Algorithm \ref{alg: unified} selects the optimal design almost surely as $T \rightarrow \infty$. 
\end{theorem}

\subsection{Asymptotic Optimality}
We next prove the asymptotic optimality, which refers to the convergence of the input budget allocation policy \(\bar{n}_{s,t} = \frac{N_{s,t}}{t}\) given by Algorithm \ref{alg: unified} to \({n}^*\) defined by \eqref{eq: input allocation} and the convergence of the simulation budget allocation policy \(\bar{m}_{i,t} = \frac{M_{i,t}}{t}\) to \({m}^*\) defined by \eqref{eq: simulation allocation} (or equivalently by \eqref{thmeq: rate balance} and \eqref{thmeq: global balance}) with given \(\bar{n} = {n}^*\), as \(t\) goes to infinity. We need the following extra assumptions on the expected performance.
\begin{assumption} \label{assump: asymptotic optimality}
\begin{enumerate}
    \item (Lipschitz continuity): For each design $i\in\mathcal{K}$, there exists $L_i > 0$, such that $\left|\mu_i(\theta_1) - \mu_i(\theta_2)\right| \le L_i \|\theta_1-\theta_2\|, \forall \theta_1,\theta_2 \in \Theta$. \label{assump: asymptotic optimality lips}
    \item Problem \eqref{eq: input allocation} has a unique optimal solution $n^* = (n^*_s)_{s\in\mathcal{S}}$. \label{assump: asymptotic optimality unique}
    % \item $c_s < U_j$ for $s \in \mathcal{S}_j,~j=1,\ldots,D$.
\end{enumerate}
        % \item For each $s\in \mathcal{S}_g$, there exists $\bar{n}_s >0$ such that $\lim_{t\rightarrow\infty}\frac{N_{s,t}}{t} =\bar{n}_s$ almost surely.
\end{assumption}
Assumption \ref{assump: asymptotic optimality}.\ref{assump: asymptotic optimality lips} is a mild assumption on the continuity of expected performance. Assumption \ref{assump: asymptotic optimality}.\ref{assump: asymptotic optimality unique} ensures if the input budget allocation given by Algorithm \ref{alg: unified} is asymptotically optimal, then it converges to the unique optimal solution of \eqref{eq: PAE optimize}. As mentioned in Section \ref{sec: input budget}, a sufficient condition for Assumption  \ref{assump: asymptotic optimality}.\ref{assump: asymptotic optimality unique} to hold is $\partial_{\theta_s} \delta_{bi} \neq 0, \forall s \in\mathcal{S}, i\neq b$, in which case \eqref{eq: PAE optimize} is a strictly concave problem. 

In addition, we make the following approximation of the rate function. 
$$ G(i) := \frac{(\mu_{b}(\mathbf{\theta^c})-\mu_{i}(\mathbf{\theta^c}))^2 }{ 2 \sum_{s \in \mathcal{S}}\frac{ g(i,s)}{\bar{n}_{s}}  +  \frac{\sigma_{i}^2}{\bar{m}_{i}}  +  \frac{\sigma^2_{b}}{\bar{m}_{b}}} \approx \frac{(\mu_{b}(\mathbf{\theta^c})-\mu_{i}(\mathbf{\theta^c}))^2 }{ 2 \sum_{s\in \mathcal{S}}\frac{ g(i,s)}{\bar{n}_{s}}  +  \frac{\sigma_{i}^2}{\bar{m}_{i}} } .$$
With such approximation, the rate balance Condition \eqref{thmeq: rate balance} becomes
\begin{equation} \label{thmeq: rate balance modified}
    \frac{(\mu_{b}(\mathbf{\theta^c})-\mu_{i}(\mathbf{\theta^c}))^2 }{ 2 \sum_{s\in \mathcal{S}}\frac{ g(i,s)}{\bar{n}_{s}}  +  \frac{\sigma_{i}^2}{\bar{m}_{i}} } = \frac{(\mu_{b}(\mathbf{\theta^c})-\mu_{i'}(\mathbf{\theta^c}))^2 }{ 2 \sum_{s\in \mathcal{S}}\frac{ g(i',s)}{\bar{n}_{s}}  +  \frac{\sigma_{i'}^2}{\bar{m}_{i'}} } \quad i\neq i'\neq b.
\end{equation}
The approximation is inspired by the following observation: from \eqref{thmeq: global balance}, we know that the optimal solution \(m^*\) should satisfy
\[
(m^*_b)^2 = \frac{\sigma_b^2}{d_b} \sum_{i \neq b} \frac{d_i (m_i^*)^2}{\sigma_i^2}.
\]
\noindent When the number of designs is large (i.e., \(K\gg 1\), which is often the case in R\&S) and the simulation noise and simulation cost for different designs are of similar magnitude, we have approximately
\[
(m^*_b)^2 \approx \sum_{k\in\mathcal{K}} (m_k^*)^2 \gg (m_i^*)^2, \forall i \neq b.
\]
Hence, \(\frac{\sigma_b^2}{m_b^*} \ll \frac{\sigma_i^2}{m_i^*}\) and we can ignore the term \(\frac{\sigma_b^2}{m_b^*}\) in the rate balance condition. On a related note, the assumption \(m^*_b \gg m^*_i\) for \(i \neq b\) was also made for the well-known OCBA algorithm in \cite{chen2000simulation} to obtain an optimal solution with explicit form. In our setting with input data, however, we still do not have an explicit solution due to the input uncertainty term \(2 \sum_{s \in \mathcal{S}} \frac{g(i',s)}{\bar{n}_{s}}\).
% To summarize, we assume the number of designs $K$ is sufficiently large such that both $ n_s^*, s\in \mathcal{S}_a$ and $m_b^*$ are much larger than the allocation ratio of those sub-optimal designs $m_i^*, i\neq b$.

Now we are ready to prove the asymptotic optimality of Algorithm \ref{alg: unified} with modified rate balance condition, as summarized in the following theorem. The proof can be found in Appendix \ref{appsec: thm optimality}.

\begin{theorem} Suppose Assumption \ref{assump: parametric input}-\ref{assump: asymptotic optimality} hold. Then, \text{almost surely, }\label{thm: optimality} 
$$\lim_{t\rightarrow\infty} \Bar{n}_{t} \rightarrow n^*,$$
where $n^*$ is the unique optimal solution of \eqref{eq: input allocation} and $\bar{m}_t$ satisfies
    \begin{flalign}
    &1.  (\textbf{global balance}) \  \lim_{t\rightarrow\infty} \left\{\bar{m}_{b,t}^2 - \frac{\sigma^2_{b}}{d_b} \sum_{i\neq b} \frac{d_i\bar{m}_{i,t}^2}{\sigma^2_{i}}\right\} = 0    && \label{eq: global balance converge} \\
    &2. (\textbf{rate balance)} \lim_{t\rightarrow\infty} \left\{ \frac{(\mu_{b}-\mu_{i})^2 }{ 2 \sum_{s\in \mathcal{S}_g}\frac{ g(i,s)}{n^*_s}  +  \frac{\sigma_{i}^2}{\bar{m}_{i,t}} } - \frac{(\mu_{b}-\mu_{i'})^2 }{ 2 \sum_{s\in \mathcal{S}_g}\frac{ g(i',s)}{n^*_s}  +  \frac{\sigma_{i'}^2}{\bar{m}_{i',t}} }\right\} = 0   \quad i\neq i'\neq b.&& \label{eq: rate balance converge}
    \end{flalign}
\end{theorem}

\section{Numerical Study} \label{sec: numerical}
In this section, we first test the performance of Algorithm~\ref{alg: unified} on a synthetic quadratic problem, whose simple and nice structure allows us to gain a good understanding of the algorithm's empirical performance. Then we apply the algorithm to a more realistic example of inventory control with multi-channel demand, demonstrating the practicality of the algorithm. %{\color{red} will read details of these numerics later.}

\subsection{Quadratic problem} \label{sec: numerical quadratic}
Consider the problem 
$$\max_x \mathbb{E}[f(x)] = -\mathbb{E}\left[\left(x-\sum_{s=1}^S \zeta_s\right)^2 \right] + \varepsilon,$$
where $\zeta_s$ follows an exponential distribution with  mean $\theta_s$ which is unknown to the decision maker, and $\varepsilon$ is a Gaussian noise with mean $0$ and variance $1$ whose distribution is known to the decision maker. Collecting data for distribution $s\le S/2$ consumes a stage-wise budget $U = 10$ with cost $c_s = 1$. For $s> S/2$, the realizations of $\zeta_s$ arrive periodically with batch size $100$ (i.e., for $s >S/2$, it consumes an individual budget equal to the batch size and the unit cost is $1$).   In addition, the simulation consumes a stage-wise budget $M=100$ with unit simulation cost $d_i = 1$, where $i \in\mathcal{K} =\{ 0,1,\ldots, 20\}$ refers to the $i$th candidate design with $x_i = x^* + i$, and $x^*$ is the optimal solution which is equal to $\sum_{s=1}^S \theta^c_s$.  We first set $S = 6$ and $\theta^c = (1,2,3,3,2,1)$.

We compare the performance of our proposed Algorithm \ref{alg: unified}, SBA, with two other algorithms. Since the problem studied in this paper is new, there is no existing work that can be directly applied to this problem. So, we adapt existing algorithms and come up with the following two procedures: (i) Equal allocation procedure (Equal), which allocates an equal budget to run the simulation for each design or collect data to estimate $\theta^c_s, s\le S/2$. That is, the algorithm sets $c_s N_{s,t} = c_{s'}N_{s',t}$ for $s,s' \le S/2$ and $ d_i M_{i,t} = d_{i'} M_{i',t}$ for all $i,i' \in \mathcal{K}$; (ii) An extension of the joint budget allocation procedure in \cite{wu2017ranking}, denoted as JBA. JBA assumes both input data collection and simulation consume the same budget, and it first decides the total budget for input data collection and then allocates the remaining budget to run simulations. Notably, in SBA we have the input data collection and simulation consume different budgets and are conducted simultaneously. When data collection is conducted ahead of simulations, the simulation resources during the data collection period are wasted and the total simulation budget is no more than $M\times T$ since the decision needs to be made before stage $T$. 
To apply JBA, we first re-scale $U$ and $c_s$ to $U=100 $ and $c_s = 10, s\le S/2$, which leads to $M=U$. Then, we set the joint total budget equal to $M \times T$ (or $U\times T$ as $U=M$). In addition, since JBA does not consider streaming given input data, the allocation policy is calculated by ignoring the streaming input data. 

\begin{figure}[h]
    \centering
    \includegraphics[width=0.7\linewidth]{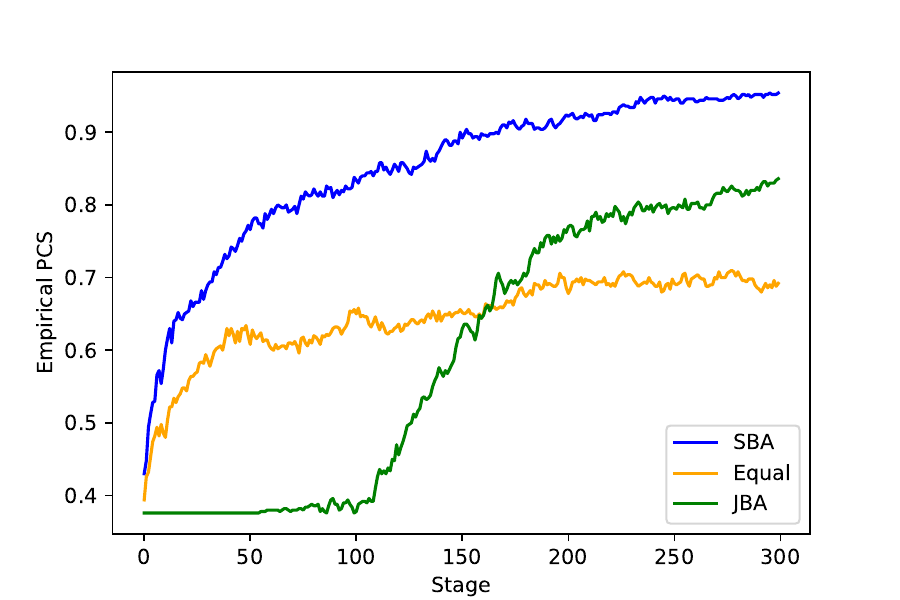}
    \caption{Quadratic example with two types of input data}
    \label{fig: quadratic general}
\end{figure}

For other experiment details, we set the total number of stage $T=400$, batch size of given input data $n_{s,t} = 20$ for $s \ge S/2$, initial simulation budget $m_{i,0} = 10$, initial batch size of input data $n_{s,0} = 50$ for $s \in \mathcal{S}$. We compare the empirical PCS (number of correct selections divided by the number of replications $N_{rep}$) 
%as well as its 95\% confidence interval ($z_{0.95}*\frac{S_{N_{rep}}}{\sqrt{N_{rep}}}$, where $z_{\alpha}$ is the $\alpha$ quantile of standard normal distribution and $ S_{N_{rep}}$ is the sample variance of correct selections with $N_{rep}$ replications) 
at each stage obtained by each algorithm. For all the following results, the empirical PCS is calculated by running $N_{rep} = 500$ micro-replications. The result is shown in Figure \ref{fig: quadratic general}.
% And we numerically show the convergence of the optimality conditions.
% \begin{figure}[h]
%     \centering
%     \includegraphics[width=0.9\linewidth]{figure_journal/Optimality_test.png}
%     \caption{Optimality condition convergence test}
%     \label{fig: optimality test}
% \end{figure}
Figure \ref{fig: quadratic general} indicates our proposed algorithm reaches the highest empirical PCS at almost all stages  %\red{is 0.73 computed as \# macro-reps with correct selections divided by 500? what do you mean by "always reaches..."? do you mean the variability across macro-replications is very small? can you given precise definitions/expressions of how you compute empirical PCS and the confidence interval and the corresponding numerical results?} \blue{I want to say SBA has the best empirical PCS at almost every stages. Now modified.}
among all, indicating the efficiency of SBA. Specifically, after $400$ stages, SBA reaches a final PCS around 0.95, JBA reaches a final PCS around 0.81 and Equal reaches a final PCS around 0.7.  The slow convergence of Equal shows the importance of optimizing the budget allocation. Furthermore, compared to JBA, our proposed algorithm SBA obtains much higher PCS during the intermediate stages than JBA does.  This is because JBA does not run the simulation at early stages, and one has very little information on the performance of designs. In case of early interruption of the procedure, SBA will still provide a selection with good quality.%\red{how do you explain that the confidence interval width of JBA increases as stage? \blue{I realized the width of CI is calculated as $C*\sigma/\sqrt{n}$, where C is constant depends on the confidence level, $\sigma^2$ is the sample variance of correct selection and $n$ is the number of micro-replications. Given $n$ and the empirical PCS, $\sigma^2$ is deterministic since the random variable $\mathbf{1}_{\text{select the optimal design}}$ only takes 2 values $\{0,1\}$. In fact, when given the empirical PCS and $n$, we know how many runs select the correct design and how many runs select the wrong designs. Also, the more empirical PCS is closer to $0.5$, the larger the sample variance. This means the CI is \textbf{not} useful to compare the stability of the algorithm. We only need to compare empirical PCS. A narrower CI here actually can only be used to show that the result is more convincing (not by accident)). I also need to delete the ``robust" argument made in the inventory example, which is not true.}} \red{I see. Please give the expressions or explain how you compute emprical PCS, confidence interval, and the implication of CI width in the paper.}

We next consider a special scenario where we only have streaming given input data. In such settings, we can compare our proposed algorithm with the algorithm proposed in \cite{wang2022fixed}, Data-Driven OCBA (denoted by DD-OCBA), which assumes the input distribution has a finite support. To apply DD-OCBA,  we discretize the support of each input distribution. DD-OCBA allocates the budget to different design-input pairs (i.e., all combinations of designs and input support points), which makes the total number of simulation alternatives equal to the product of the number of designs and the number of discretized support points, the latter of which unfortunately grows exponentially in $S$. Hence, here the initialization of DD-OCBA requires a larger amount of initial simulation replications than other algorithms. To make computation tractable for DD-OCBA in this problem, we set $S=2, \theta^c = (2,1)$. %\red{The simulation cost $d_i =1, i\in \mathcal{K}$. (why is this sentence here? it's out of context, and it has already been mentioned before.) } 
For DD-OCBA, we set the initial number of simulation replications for each design-input  pair to be $2$, which is a minimum number to obtain a variance estimate. For other algorithms, we set this initial number of simulation replications to be $m_{i,0} = 10$.
For a fair comparison, we set the same total budget for all algorithms, where the total budget includes the initial simulation replications and the subsequent simulations. %\red{I rewrote the sentence above. Is that what you meant? If yes, I  this sentence should be moved much earlier since it applied to all algorithms. } 
For this numerical experiment, we set $M = 30$, $T=300$, $n_{s,t} = 10, n_{s,0} = 20$ for $s \in \mathcal{S}$, $\mathcal{K}= \{0,1,\ldots,12\}$. Since the performance of DD-OCBA depends on how the supports of input distributions are discretized, we test on different discretizations by setting the number of discretized points $N_{dis}$ for each input distribution to range from $\{6,8,10\}$ and the precision (distance between two support points) to be $0.5$. 
\begin{figure}[h]
    \centering
    \includegraphics[width = 0.7\linewidth]{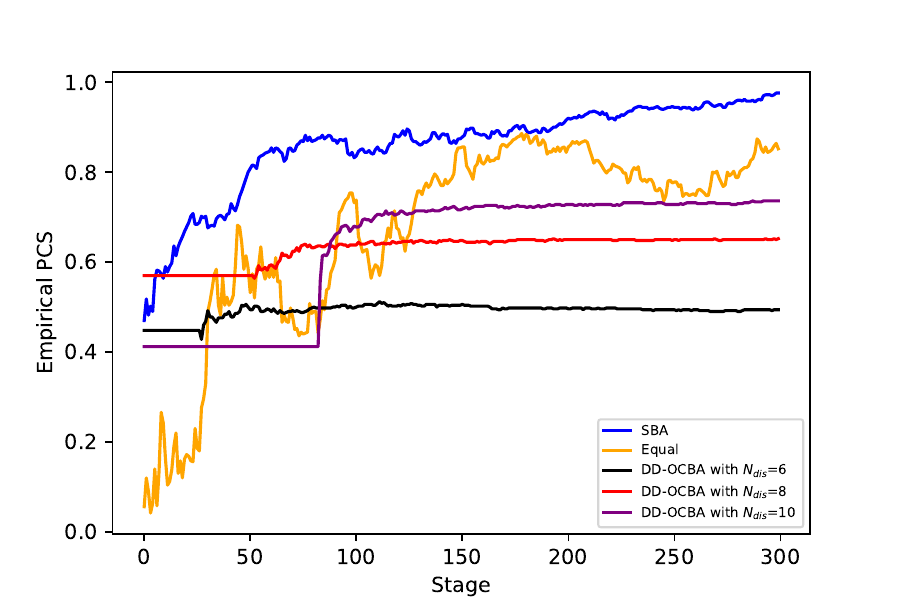}
    \caption{Quadratic example with given data}
    \label{fig: quadratic given}
\end{figure}

Figure \ref{fig: quadratic given} indicates that even in the special setting with only given input data, SBA has the most competitive performance (highest final empirical PCS at around $0.87$). By comparison, Equal reaches a final PCS of around $0.5$. For DD-OCBA, when the discretization number ($N_{dis}$) ranges in $\{6,8,10\}$,
% the resulting approximated problem is actually misspecified as the best design is not equal to the true best design. As a result, the empirical PCS of the DD-OCBA will not converge to $1$ but decrease as more simulations are run. For those larger discretization numbers ($N_{dis} = 8,10$), the approximated problem has the same optimal design as the original problem and the empirical PCS increases as more simulations are run,
DD-OCBA reaches a final empirical PCS around $0.5$, $0.6$ and $0.7$, respectively, all of which are lower than the final empirical PCS reached by SBA. Moreover, due to the large number of combinations of designs and input support points, it takes a large amount of simulation budget to simply run a small number of initial simulation replications for DD-OCBA (it takes around 35, 60, 90 stages of simulation budgets for discretization number 6, 8, 10, respectively), whereas SBA does not suffer from this issue.

%Besides the simple quadratic example in this section, we next consider a more general problem.
\subsection{Inventory control with multi-channel demand}
Consider the following inventory control problem with multi-channel demand. Suppose that we are running a capacitated production system and want to minimize the expected total cost over a finite number of time periods. The decision variable is the order-up-to level, i.e., the quantity that we should fill up to once the inventory falls below that level. %Please note this is an offline planning problem, where we select the best inventory policy after finishing the simulation.   
Meanwhile, the production amount in each time period is capped. At the beginning of each period, we are given the amount produced in the previous period. Then, the demand is revealed over the span of the period, and we fulfill the total demand (both backlog and current demand) using the current inventory, after which unfulfilled demand becomes the new backlog. The decision on the production amount is carried out at the end of the period.  The demand can come from different channels, including demand from different online retailing platforms and those from different physical stores. For some channels (e.g., old channels that have operated for a while), the demand data arrives periodically, whereas for others (e.g., new channels), one needs to actively collect the demand data. All demands share the same inventory.
 
The variables are listed as follows: $i$ is the order-up-to level, $I_u$ is the inventory level at the end of the $u$th period, $S$ is the number of channels, $\xi_{v,s}$ is the demand from channel $s$ at the $v$th period, and $R_v$ is the production amount at the $v$th period. Let $I_0 = i$ and $R_0 = 0$. Starting from $v=1$, the system dynamics evolve according to the following equations,
\begin{align*}
& I_{v} = I_{v-1} + R_{v-1} - \sum_{s=1}^S \xi_{v,s},\\
& R_v = \min\{R^*, (i - I_{v})^{+} \},
\end{align*}
where $a^+ := \max\{0, a\}$, and $R^*$ is the maximum production amount. Assume that the demand quantities are independent random variables, where each $\xi_{v,s}$ follows a Poisson distribution with mean $\theta^c_{s}$. Let $c_H$ be the holding cost per unit and $c_B$ be the backlog cost per unit. Then, the cost at the $v$th period is 
\begin{equation*}
c_v := c_H (R_{v-1} + I^+_v) + c_B I_s^-,
\end{equation*}
where $a^- := -\min\{a, 0\}$. The expected total cost over $V$ number of periods is 
\begin{equation*}
\mu_i(\theta^c) = \mathbb{E}\left(\sum_{v=1}^V c_v \right).
\end{equation*}
The goal is to select the optimal order-up-to level $i$ to minimize the expected total cost.

\begin{figure}[ht]
    \centering
    \includegraphics[width = 0.7\linewidth]{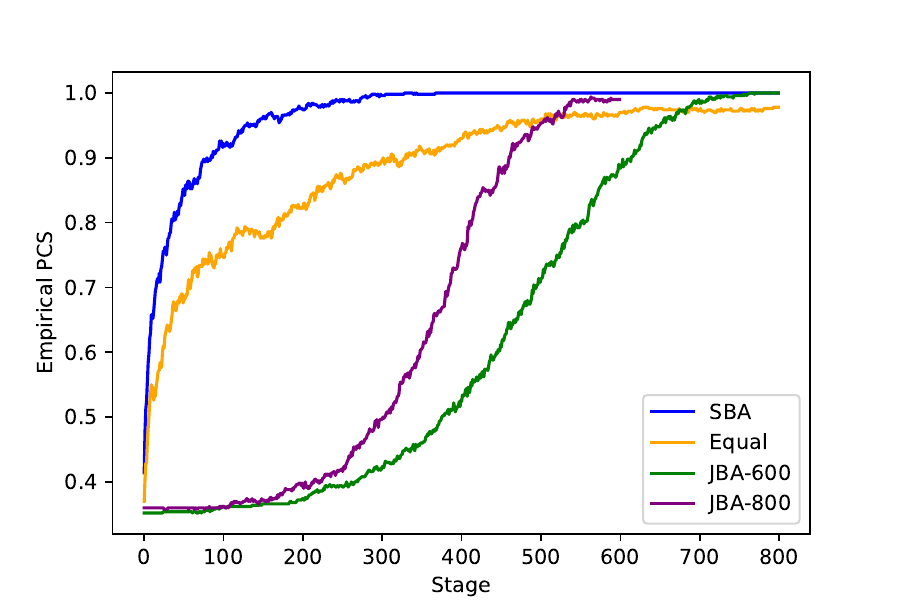}
    \caption{Inventory example with $2$ demand channels.}
    \label{fig: inventory 2 channel}
\end{figure}

\begin{figure}[h]
    \centering
    \includegraphics[width = 0.7\textwidth]{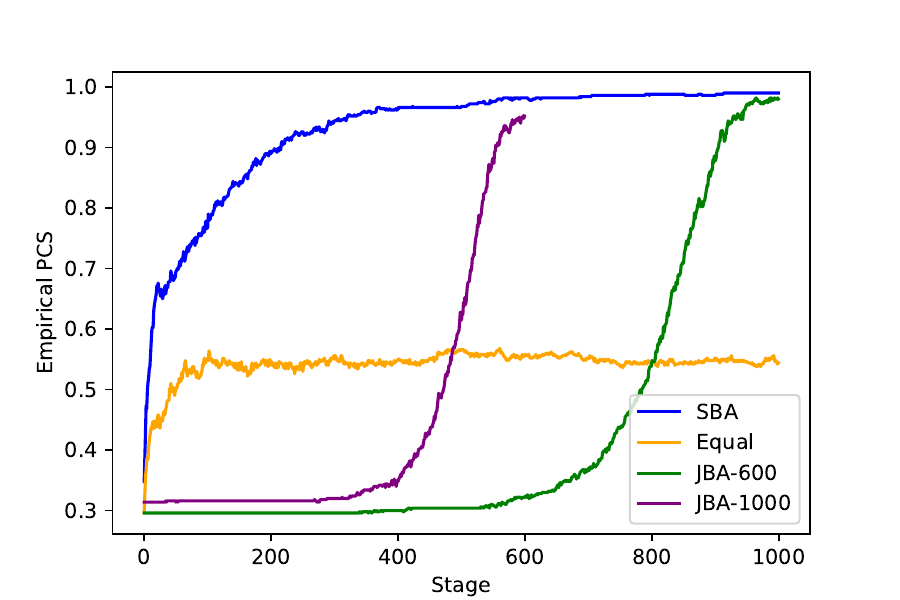}
    \caption{ Inventory example with $4$ demand channels.}
    \label{fig: inventory 4 channel}
\end{figure}

In this numerical experiment, we set $V=6$, $c_H=0.5,c_B=1$. %\red{add the notations for these two costs}
We test for two scenarios. In the first scenario,  there are two demand channels in total, and we set $S=2$ and $\theta^c = (5,2)$. The candidate set is $\mathcal{K} = \{0,1,2,\ldots,0\}$ with $i$th candidate to be order-up-to level $i+1$. In the second scenario, there are four demand channels, and we set $S=4$ and $\theta^c = (4,4,3,2)$. The candidate set is also  $\mathcal{K} =\{ 0,1,2,\ldots, 9\}$ but with $i$th candidate to be order-up-to level $ 10+ 2*i$. In both scenarios, the input data for $s \le S/2$ share a budget $U = 30$ with unit cost $c_s = 5$ and are actively collected, and the input data for $s>S/2$ are streaming given data with batch size $50$.  We also set $M = 30,d_i=1, i\in\mathcal{K}$, $n_{s,t} = 50, s\ge S/2$, $n_{1,0} = n_{2,0} = 10$,  $m_{i,0} = 10, i\in \mathcal{K}$. The total number of stages $T$ is set to $800$ for the first scenario and $1000$ for the second scenario.%\red{it seems you sometimes use $\mathcal{K}$ and sometimes $\mathcal{K}$} 
We plot the empirical PCS with respect to the stage %\red{stage or period? stick to one term. make sure the labels in figures are consistent with the texts.} 
in Figure \ref{fig: inventory 2 channel} and \ref{fig: inventory 4 channel}. 

The result in Figure \ref{fig: inventory 2 channel} and Figure \ref{fig: inventory 4 channel} are similar to the result in Figure \ref{fig: quadratic general}, where SBA obtains the highest empirical PCS in almost all stages (reaches final empirical PCS $1.00$ within $400$ stages for $S=2$ and final empirical PCS $0.98$ within $1000$ stages for $S=4$), showing its efficiency in different applications. By comparison, Equal reaches final PCS around $0.92,0.55$  for $S=2,4$, respectively. For JBA, we plot two PCS curves corresponding to a total number of stages \( T \) set to \( 600 \) (JBA-600) and \( 800 \) (JBA-800) in Figure \ref{fig: inventory 2 channel}, and \( 1000 \) (JBA-1000) in Figure \ref{fig: inventory 4 channel}. In Figure \ref{fig: inventory 4 channel}, although JBA-600 seems to have the potential to outperform SBA if \( T \) were extended, in fact when we increase \( T \) from \( 600 \) to \( 1000 \), SBA reaches a final PCS (approximately \( 0.98 \)) still higher than JBA-1000 (approximately \( 0.97 \)). The underlying reason lies in the way allocation policies are determined. JBA requires the total number of stages \( T \) to be specified in advance, and its allocation policy at any stage depends directly on \( T \). In contrast, SBA's allocation policy is independent of \( T \) and relies solely on past samples.
As a result, when \( T \) increases from \( 600 \) to \( 1000 \), JBA-1000 effectively ``starts'' its simulation later than JBA-600, as it allocates additional time for collecting input data. On the other hand, SBA maintains the same allocation policy for stages prior to \( 600 \).
The reasons why SBA outperforms JBA in Figure \ref{fig: inventory 2 channel} and Figure\ref{fig: inventory 4 channel} (as well as Figure \ref{fig: quadratic general}) can be understood as follows. First, as just mentioned, JBA runs simulation only after input data collection, which does not fully utilize the total available computing resources within $T$ stages. Second, it
ignores the impact of given input data when calculating the allocation policy for either simulations or input data collection. Third, it computes its allocation policy for input data collection only once based on some very rough estimations of several parameters based on a few simulations under roughly estimated input distributions. The allocation policy computed by SBA, however, is adjusted from stage to stage and as a result converges to the optimal allocation policy.
% What's more, the performance of SBA is also much more ``robust"  in the sense that the width of the confidence interval for SBA is much smaller than the confidence interval for other algorithms, especially for JBA. This is because the allocation policy by SBA is adjusted from stage to stage and it converges to the optimal allocation policy quickly, while JBA only computes the  Hence, SBA is more reliable for such robust performance.
% In addition, the difference between SBA and Equal in this inventory problem is not as stark as in the quadratic problem, since the number of designs $K=10$ in this problem is smaller than  $K=21$ in the quadratic problem. 

\section{Conclusion} \label{sec: conclusion}
This paper addresses the problem of ranking and selection with streaming input data, where the input data collection potentially consumes different types of resources and can be conducted simultaneously with the simulation. The formulation involves characterizing the asymptotic behavior of performance estimators that deals with non-i.i.d. data. Two optimization problems are then introduced to optimally allocate multiple budgets for input data collection and simulation budget, with the primary goal of maximizing the asymptotic exponential convergence rates of the probability of acceptable estimation and probability of correct selection, respectively. We then develop a multi-stage simultaneous budget allocation procedure that is supported by statistical guarantees, including consistency and asymptotic optimality. Finally, the paper concludes with numerical studies that demonstrate the highly competitive performance of the proposed procedure.

%\section*{Acknowledgement}
%The authors gratefully acknowledge the support by the Air Force Office of Scientific Research under Grant FA9550-22-1-0244, the National Science Foundation under Grant NSF-DMS2053489, and the National Science Foundation AI Institute for Advances in Optimization under Grant NSF-2112533.
% \red{NSF, Air force, AI4opt}

% \bmsection*{Author contributions}

% This is an author contribution text. This is an author contribution text. This is an author contribution text. This is an author contribution text. This is an author contribution text.

\section*{Acknowledgments}
The authors gratefully acknowledge the support by the Air Force Office of Scientific Research under Grant FA9550-22-1-0244, the National Science Foundation under Grant NSF-ECCS2419562 and the NSF AI Institute for Advances in Optimization (AI4OPT) under Grant
NSF-2112533.

\bibliographystyle{plain}
\bibliography{ref}

\newpage
\appendix
\section{Technical Proof}
\newcommand{\barm}{\bar{m}}
\newcommand{\bn}{\bar{n}}
\newcommand{\hatb}{{\widehat{b}}}
\newcommand{\srlogr}{\sqrt{r\log\log r}}
\newcommand{\slogrr}{\sqrt{\frac{\log\log r}{r}}}
\newcommand{\tj}{{t^j_{\ell_j}}}
\subsection{Proof of Lemma \ref{lem: CLT input}} \label{appsec: Lemma CLT input}
\textbf{Proof:}
By Assumption \ref{assump: CLT active}.2, we have 
\begin{align*}
    &\mu_i(\widehat{\theta}_t) - \mu_k(\widehat{\theta}_t) - \left( \mu_i(\theta^c) - \mu_k(\theta^c)\right) \\
    = &\underbrace{\left(\nabla \mu_i(\theta^c)-\nabla\mu_k(\theta^c)\right)^\top\left(\widehat{\theta}_t - \theta^c\right)}_{(I)} + \underbrace{O\left(\|\widehat{\theta}_t - \theta^c \|_2^2\right)}_{(II)}
\end{align*}
For the first term $(I)$, note 
$$\widehat{\theta}_{s,t} = \frac{1}{N_{s,t}}\sum_{\ell=1}^{N_{s,t}} D_s(\zeta_{s,\ell}).$$
Since $\zeta_{s,\ell}, \ell\ge 1$ are i.i.d. data and $\lim_{t\rightarrow\infty} \frac{N_{s,t}}{t} = \Bar{n}_s$ almost surely, we have as $t\rightarrow \infty$,
$$\sqrt{t}(\widehat{\theta}_{s,t} - \theta_s^c) = \sqrt{\frac{t}{N_{s,t}}} \sqrt{N_{s,t}}(\widehat{\theta}_{s,t} - \theta_s^c) \Rightarrow  \Bar{n}_s^{-\frac{1}{2}} \mathcal{N}(0,\Sigma_{D,s}).$$
Furthermore, since the input data process for different distributions are independent for different $s$ by Assumption \ref{assump: unbiased estimator}.3, we obtain
$$\sqrt{t} (I) \Rightarrow \mathcal{N}\left(0, \sum_{s\in\mathcal{S}} \Bar{n}_s^{-1} \partial_{\theta_{s}} \delta_{ik}^\top \Sigma_{D,s} \partial_{\theta_s} \delta_{ik}\right).$$
For the second term $(II)$, we can apply the law of iterated logarithm to obtain almost surely,
\begin{align*}
    & \sqrt{t} \|\widehat{\theta}_{s,t} - \theta_s^c\|_2^2 =   \sqrt{t} \cdot O\left( \frac{\log\log t}{t}\right) = O\left( \frac{\log\log t}{\sqrt{t}}\right).
\end{align*}
This implies $\sqrt{t} \|\widehat{\theta}_{t} - \theta_s^c\|_2^2 \rightarrow 0$ almost surely. This completes the proof. \hfill $\blacksquare$.
\subsection{Proof of Lemma \ref{lem: convex input rate optimization}}
\label{appsec: lem convex input rate optimization}
\textbf{Proof:}
We first prove the convexity, which can be implied by the concavity of function 
$h_i(\bar{n}):= \frac{\delta^2_{bi}}{\sum_{s\in\mathcal{S}} \bar{n}^{-1}_s\partial_{\theta_s} \delta_{bi}\Sigma_{D,s}\partial_{\theta_s} \delta_{bi}}$. We prove the a more general form in the following lemma.
\begin{lemma} \label{lem: concave function}
    Let $ f(x) =1/( \sum_{i=1}^n \frac{a_i}{x_i} + d)$,  $a_i > 0$ for $i=1,2,\dots,n$ and $d\ge0$. Then $f(x)$ is concave for all $x>0$. 
   \end{lemma}
\textbf{Proof:} 
    
    We prove the concavity of the multivariate function by proving the concavity along all lines. For any $y \in \mathbb{R}^n$, let $g(t) = f(x+ty)$ where $t \in \mathbb{R}$ such that $x+ty > 0$. We have 
 \begin{align*}
     g^{\prime\prime}(t) 
     =& \frac{2}{(\sum_{i=1}^n\frac{a_i}{x_i+ty_i} +d)^3  } \left\{ \left[ \sum_{i=1}^n \frac{a_iy_i}{(x_i+ty_i)^2}  \right]^2 - \sum_{i=1}^n \frac{a_i y_i^2}{(x_i+ty_i)^3} \left(\sum_{i=1}^n \frac{a_i}{x_i+ty_i}+d\right) \right\}  \\
     \le &\frac{2}{(\sum_{i=1}^n\frac{a_i}{x_i+ty_i} +d)^3  } \left\{ \left[ \sum_{i=1}^n \frac{a_iy_i}{(x_i+ty_i)^2}  \right]^2 - \sum_{i=1}^n \frac{a_i y_i^2}{(x_i+ty_i)^3} \sum_{i=1}^n \frac{a_i}{x_i+ty_i} \right\}\\
     \le &0
 \end{align*}
 where the last inequality uses the Cauchy inequality. Hence, $f$ is concave in $x > 0$. 
\hfill $\blacksquare$

Hence, we prove the concavity of Problem \eqref{eq: input allocation}. Notably, if $\Bar{n}_s = 0$ for some $s\in\mathcal{S}$, then we have there exists $i\neq b$ with $\partial_{\theta_s} \delta_{bi} \neq 0$, such that 
$ h_i(\bar{n}) = 0$. This implies $\bar{n}$ cannot be an optimal solution. Hence, the optimal solution for $\bar{n}$ must be strictly positive.  \hfill $\blacksquare$

% Finally, we prove the uniqueness. First notice \eqref{eq: input allocation} is equivalent to the following problem.
% \begin{align*}
%     \max_{\bar{n}} \min_{i\neq b} &~~h_i(\bar{n})\\
%     s.t. &~~ \sum_{s\in\mathcal{S}_j} c_s \bar{n}_s = U_j, ~~ j=1,\ldots,D
% \end{align*}

% Let $\Omega$ be the set of all optimal solutions to \eqref{eq: input allocation}. Then we know 
% \eqref{eq: input allocation} is further equivalent to 
% \begin{align*}
%     \max_{\bar{n}} \min_{i\neq b} &~~h_i(\bar{n})\\
%     s.t.&~~ \bar{n} \in \Omega
% \end{align*}
% Since $\Omega $ is also a convex set, we can prove the uniqueness by proving the function 
% $\min_i h_i(\Bar{n})$ is strictly concave over $\Omega$. Consider an arbitrary $\tilde{n} \in \Omega$. Let $I(\tilde{n}) = \{i \neq b, i \in  \arg\min_{k\neq b} h_k(\tilde{n}) \}$. 

% We first claim $\forall s\in\mathcal{S}$, there exists $i\in I(\tilde{n})$,such that $ \partial_{\theta_s} \delta_{bi} (\theta^c) \neq 0$. Otherwise, we have $h_i(\bar{n}), i \in I(\tilde{n})$ is a constant in term of $\bar{n}_s$. Since $\tilde{n}$ is an optimal solution, $\tilde{n}_s > 0$. Further since $h_i(\bar{n})$ is continuous in $\bar{n}$ in a neighborhood of $\tilde{n}$, then we can slightly decrease the value of $\tilde{n}_s$ and increase the value of $\tilde{n}_{s'}, s'\neq s$ to obtain $\tilde{n}'$ such that $\sum_{s'\in\mathcal{S}} c_{s'} \tilde{n}'_{s'} = U$ 

\subsection{Proof of Theorem \ref{thm: normality}}
\label{appsec: thm normality}

\textbf{Proof:}
Since 
\begin{align*}
     \widehat{\delta}_{ij,t} - \delta_{ij}(\theta^c)  = & \frac{1}{M_{i,t}}\sum_{\ell=1}^t \sum_{r=1}^{m_{i,\ell}}\left(X_i^r(\widehat{\theta}_\ell) - \mu_i(\theta^c) \right) - \frac{1}{M_{j,t}}\sum_{\ell=1}^t \sum_{r=1}^{m_{j,\ell}}\left(X_j^r(\widehat{\theta}_\ell) - \mu_j(\theta^c) \right)\\
    =&\frac{1}{M_{i,t}}\sum_{\ell=1}^t \sum_{r=1}^{m_{i,\ell}}\left(X_i^r(\widehat{\theta}_\ell) -\mu_i(\widehat{\theta}_\ell) \right) +\frac{1}{M_{i,t}} \sum_{\ell=1}^t m_{i,\ell}\left(\mu_i(\widehat{\theta}_\ell) -\mu_i(\theta^c)\right)\\
    &-\left[\frac{1}{M_{j,t}}\sum_{\ell=1}^t \sum_{r=1}^{m_{j,\ell}}\left(X_j^r(\widehat{\theta}_\ell) -\mu_j(\widehat{\theta}_\ell) \right) +\sum_{\ell=1}^t \frac{1}{M_{j,t}}m_{j,\ell}\left(\mu_j(\widehat{\theta}_\ell) -\mu_j(\theta^c)\right) \right] \\
    = &\underbrace{\frac{1}{M_{i,t}}\sum_{\ell=1}^t \sum_{r=1}^{m_{i,\ell}}\left(X_i^r(\widehat{\theta}_\ell) -\mu_i(\widehat{\theta}_\ell) \right) }_{Z_1} + \underbrace{\frac{1}{M_{j,t}}\sum_{\ell=1}^t \sum_{r=1}^{m_{j,\ell}}\left(X_j^r(\widehat{\theta}_\ell) -\mu_j(\widehat{\theta}_\ell) \right)}_{Z_2}\\
    &+\underbrace{\frac{1}{M_{i,t}}\sum_{\ell=1}^t m_{i,\ell}\left(\mu_i(\widehat{\theta}_\ell) -\mu_i(\theta^c)\right) - \frac{1}{M_{j,t}} \sum_{\ell=1}^t 
 m_{j,\ell}\left(\mu_j(\widehat{\theta}_\ell) -\mu_j(\theta^c)\right) }_{Z_3}.
\end{align*}
 Denote by $\mathcal{F}_t = \sigma(\widehat{\theta}_1,\ldots,\widehat{\theta}_t)$, the sigma-algebra generated by past input estimators. Note conditioned on $\mathcal{F}_t$, for $\ell \le t$, $X_i^r(\widehat{\theta}_\ell)$ only depends on $\widehat{\theta}_\ell$ since the simulation output does not affect future input data.
 We have the following lemma from the proof of Theorem 3 in \cite{wu2022data}.

 \begin{lemma} Under the same assumptions as Theorem \ref{thm: normality}
     $$\sqrt{t}(Z_1+Z_2)|\mathcal{F}_t \Rightarrow \mathcal{N}\left(0,\bar{m}_i^{-1}\sigma^2_i(\theta^c) + \bar{m}_j^{-1}\sigma^2_j(\theta^c)\right ).$$
 \end{lemma}
 
For $Z_3$, it  can be further expressed as
\begin{align*}
    Z_3 = &\frac{1}{M_{i,t}}\sum_{\ell=1}^t m_{i,\ell}\left(\nabla \mu_i(\theta^c)^\top (\widehat{\theta}_\ell - \theta^c ) + \mathcal{O}(\|\widehat{\theta}_\ell - \theta^c\|_2^2)\right) - \\
    &\frac{1}{M_{j,t}} \sum_{\ell=1}^t 
 m_{j,\ell}\left(\nabla \mu_j(\theta^c)^\top (\widehat{\theta}_\ell - \theta^c )+\mathcal{O}(\|\widehat{\theta}_\ell - \theta^c\|_2^2)\right) \\
 =&\underbrace{\sum_{\ell=1}^t \left[\frac{m_{i,\ell}}{M_{i,t}}\nabla\mu_i(\theta^c) - \frac{m_{j,\ell}}{M_{j,t}}\nabla\mu_j(\theta^c)\right]^\top (\widehat{\theta}_\ell - \theta^c)}_{Z_{3,1}}  \\
 &+\underbrace{\sum_{\ell=1}^t \left[ \frac{m_{i,\ell}}{M_{i,t}} + \frac{m_{j,\ell}}{M_{j,t}}\right] \mathcal{O}(\|\widehat{\theta}_\ell - \theta^c\|_2^2)}_{Z_{3,2}}.
\end{align*}
Denote by $D_{s,r}^\tau$ the $\tau^{th}$ input data at stage $r$ for input distribution $s$. i.e., $D_{s,r}^\tau = D_s(\zeta_{s,N_{s,r-1}+\tau})$.
Since for each $s$, $\widehat{\theta}_{s,\ell} = \frac{1}{N_{s,\ell}} \sum_{r=1}^\ell \sum_{\tau=1}^{n_{s,r}} D_{s,r}^\tau$. We have the component in the product of $Z_{3,1}$ with respect to input distribution $s$ (denote by $Z_{3,1,s} = \sum_{\ell=1}^t \left[\frac{m_{i,\ell}}{M_{i,t}}\nabla_{\theta_s}\mu_i(\theta^c) - \frac{m_{j,\ell}}{M_{j,t}}\nabla_{\theta_s} \mu_j(\theta^c)\right]^\top (\widehat{\theta}_{s,\ell} - \theta^c_s) $ and $Z_{3,1} = \sum_{s\in\mathcal{S}} Z_{3,1,s}$) can be expressed as 
\begin{align*}
    Z_{3,1,s} = & \sum_{\ell=1}^t \left[\frac{m_{i,\ell}}{M_{i,t}}\nabla\mu_i(\theta^c) - \frac{m_{j,\ell}}{M_{j,t}}\nabla\mu_j(\theta^c)\right]^\top ( \frac{1}{N_{s,\ell}} \sum_{r=1}^\ell \sum_{\tau=1}^{n_{s,r}} D_{s,r}^\tau) \\
    =&\sum_{\ell=1}^t \left(\sum_{r=\ell}^t \frac{1}{N_{s,r}}  \left[\frac{m_{i,r}}{M_{i,t}}\nabla\mu_i(\theta^c) - \frac{m_{j,r}}{M_{j,t}}\nabla\mu_j(\theta^c)\right]\right)^\top\left( \sum_{\tau=1}^{n_{s,\ell}}  D_{s,r}^\tau\right).
\end{align*}
Here $D_{s,r}^\tau \forall r,\tau$ are i.i.d. data. The intuition for the second equality here is to ``regroup" the data such that we can rewrite $Z_{3,1,s}$ as a summation of weighted i.i.d. data.
The following lemma holds, which is a result from Proof of Theorem 3 in \cite{wu2022data}.
\begin{lemma}
Under the same assumptions as Theorem \ref{thm: normality},
    $$ \sqrt{t} Z_{3,1,s} \Rightarrow \mathcal{N}\left(0, \frac{2}{\bar{n}_s}\nabla_{\theta_s} \delta_{ij}(\theta^c)^\top \Sigma_{D,s} \nabla_{\theta_s} \delta_{ij}(\theta^c)\right).$$
\end{lemma}

Then, note data for different input distributions are independent. 
We then have
$$ \sqrt{t} Z_{3,1} \Rightarrow \mathcal{N}\left(0, \sum_{s\in\mathcal{S}} \frac{2}{\bar{n}_s}\nabla_{\theta_s} \delta_{ij}(\theta^c)^\top \Sigma_{D,s} \nabla_{\theta_s} \delta_{ij}(\theta^c)\right).$$
For $Z_{3,2}$, it is actually an error term with higher order compared with $Z_{3,1}$, it can be easily shown that, (following proof for Theorem 3 in \cite{wu2022data}),$\sqrt{t}Z_{3,2} \rightarrow 0$ almost surely.

Finally, note the characteristic function of $\sqrt{t}(Z_1+Z_2+Z_3)$ is
$$f(x) = \mathbb{E}[\mathrm{e}^{ix\sqrt{t}(Z_1+Z_2+Z_3)}] = \mathbb{E}[ \mathrm{e}^{ixZ_3} \mathbb{E}[ \mathrm{e}^{ix\sqrt{t}(Z_1+Z_2)} |\mathcal{F}_t]].$$
Since $|\mathrm{e}^{ix\sqrt{t}Z_3}|\le 1$, we have
if both $\mathbb{E}[ \mathrm{e}^{ix\sqrt{t}(Z_1+Z_2)} |\mathcal{F}_t] $ and $\mathbb{E}[ \mathrm{e}^{ix\sqrt{t}Z_3} ]$
converge to a constant as $t\rightarrow \infty$ almost surely, then we know $$\lim_{t\rightarrow\infty} f(x)  = \lim_{t\rightarrow\infty}  \mathbb{E}[ \mathrm{e}^{ix\sqrt{t}(Z_1+Z_2)} |\mathcal{F}_t] 
\lim_{t\rightarrow\infty} \mathbb{E}[ \mathrm{e}^{ix\sqrt{t}Z_3} ].$$
The weak convergence of $\mathbb{E}[ \mathrm{e}^{ix\sqrt{t}(Z_1+Z_2)} |\mathcal{F}_t] $ and $\mathbb{E}[ \mathrm{e}^{ix\sqrt{t}Z_3} ]$ are guaranteed by the weak convergence of $\sqrt{t}(Z_1+Z_2)|\mathcal{F}_t$ and $\sqrt{t}Z_3$. Hence, we obtain
$\sqrt{t}(Z_1+Z_2+Z_3) \Rightarrow \mathcal{N}(0,2 \sum_{s \in \mathcal{S}}\bar{n}_s^{-1} \partial_{\theta_s} \delta_{ik}^\top \Sigma_{D,s}\partial_{\theta_s} \delta_{ik}   +   \bar{m}_i^{-1} \sigma_i^2 +   \bar{m}_k^{-1}\sigma^2_k)$ as desired. 

\hfill $\blacksquare$

\par

\subsection{Proof of Theorem \ref{thm: optimality conditions}}
\label{appsec: thm optimality conditions}
We first introduce the following lemma that summarizes the property of the rate function $G_i, i\neq b$.
\begin{lemma} \label{lem: rate function}
Assume the average batch size of given input data is strictly positive, i.e.,  $\bar{n}_s>0, s\in \mathcal{S}_g$.  Denote by $G_i(\bar{m}_b,\bar{m}_i) = \frac{\delta^2_{bi}(\mathbf{\theta}^c) }{\Tilde{\sigma}_{bi}^2}$ the rate function for sub-optimal design $i$. Suppose Assumption \ref{assump: parametric input} - \ref{assump:input effect} hold. Then  $G_i$ is increasing and concave in $\bar{m}_b,\bar{m}_i$ for $\bar{\mathbf{m}} \ge 0$.
\end{lemma}

\textbf{Proof of Lemma \ref{lem: rate function}:}

    The monotone property is obvious. The convexity follows from Lemma \ref{lem: concave function}.
\hfill $\blacksquare$

\noindent\textbf{Proof of Theorem \ref{thm: optimality conditions}:}
We first show that the strong duality holds. By Lemma \ref{lem: concave function}, we know that \eqref{eq: simulation allocation} is a concave optimization. To prove the strong duality holds, we only need to show \eqref{eq: simulation allocation} satisfies the slater's condition (e.g., see chapter 5 in \cite{boyd2004convex}). Notably,  $\{\Bar{m}_i = \frac{1}{K} \ \forall i\in\mathcal{K}, z = -1\}$ is a strict feasible solution. Hence, the Slater's condition holds and we can prove Theorem \ref{thm: optimality conditions} using KKT condition.

 For positive solution $\{\bar{m}\}$, apply the KKT conditions and we obtain 
% \begin{equation*}
% L := \textbf{APCS} -\lambda (\sum_{s \in [S]} c_s \bar{n}_s + \sum_{i\in\mathcal{K}}\bar{m}_i - B),
% \end{equation*}
% We obtain
\begin{align}
    & 1-\sum_{i \neq b} \lambda_i = 0 \label{eq: KKT mu_i}\\
    & d_i \gamma_0 -  \frac{\lambda_i}{2} \frac{\delta^2_{bi}(\mathbf{\theta}^c)}{\Tilde{\sigma}_{bi}^3}\frac{1}{\bar{m}_i^2}   \sigma_i^2(\mathbf{\theta}^c) = 0  \quad &\forall i \neq  b \label{eq: KKT m_i}\\
    &d_b \gamma_0 -\sum_{i\neq b}  \frac{\lambda_i}{2} \frac{\delta^2_{bi}(\mathbf{\theta}^c)}{\Tilde{\sigma}_{bi}^3}\frac{1}{\bar{m}_b^2}   \sigma_b^2(\mathbf{\theta}^c) = 0 \label{eq: KKT m_b} \\
    & \lambda_i( \frac{\delta^2_{bi}(\mathbf{\theta}^c) }{\Tilde{\sigma}_{bi}^2} - z ) = 0 \quad  &\forall i \neq b \label{eq: KKT local}
    % & \sum_{s \in [S]} c_s \bar{n}_s  =  T_I.\label{eq: KKT input budget}\\
    % &\sum_{i\in\mathcal{K}} \bar{m}_i = T  _S \label{eq: KKT simulation budget}
\end{align}
For necessity, from \eqref{eq: KKT mu_i} we have there exists $i \neq b$ such that $\lambda_i > 0$. For this $i$, from \eqref{eq: KKT m_i} we obtain $\gamma_0 > 0$. Hence for all $i \neq b$, $\lambda_i = \frac{2d_i\gamma_0  \Tilde{\sigma}_{bi}^3 \bar{m}_i^2}{ \sigma^2_i(\mathbf{\theta^c})\delta_{bi}^2(\mathbf{\theta}^c) }> 0$. Then, from \eqref{eq: KKT local}, we have 
$
    \frac{\delta_{bi}^2(\mathbf{\theta^c})}{\Tilde{\sigma}_{bi}} = \frac{\delta_{bj}^2(\mathbf{\theta^c})}{\Tilde{\sigma}_{bj}}  \quad \forall i\neq j\neq b,
$
which proves \eqref{thmeq: rate balance}. Substituting $\lambda_i= \frac{2d_i\gamma  \Tilde{\sigma}_{bi}^3 \bar{m}_i^2}{ \sigma^2(\mathbf{\theta^c})\delta_{bi}^2(\mathbf{\theta}^c) }$ in \eqref{eq: KKT m_b}, we obtain
$
    \bar{m}_b^2 = \frac{\sigma^2_{b}(\mathbf{\theta^c})}{d_b} \sum_{i\neq b} \frac{d_i\bar{m}_i^2}{\sigma^2_i(\mathbf{\theta^c})},
$
which proves \eqref{thmeq: global balance}.

For sufficiency,  it suffices to show KKT conditions are satisfied if the two optimality conditions \eqref{thmeq: rate balance}, \eqref{thmeq: global balance} are satisfied. Let $i_0 \neq b$ be some fixed sub-optimal design. Let
$z =  \frac{\delta^2_{bi_0}(\mathbf{\theta}^c) }{\Tilde{\sigma}_{bi_0}^2}$, $
 \lambda_i = \left. \frac{\Tilde{\sigma}_{bi}^3 \bar{m}_i^2 d_i}{\sigma_i^2(\mathbf{\theta^c})\delta_{bi}^2(\mathbf{\theta}^c) } \middle/  \sum_{j\neq b} \frac{ \Tilde{\sigma}_{bj}^3 \bar{m}_j^2 d_j }{\sigma_j^2(\mathbf{\theta^c})\delta_{bj}^2(\mathbf{\theta}^c) }\right. $ ,  $
 \gamma_0 =  \left. 1 \middle/  \sum_{j\neq b} \frac{2 \Tilde{\sigma}_{bj}^3 \bar{m}_j^2 d_j}{ \sigma_j^2(\mathbf{\theta^c})\delta_{bj}^2(\mathbf{\theta}^c) }\right..$
Then one can verify all the KKT conditions are satisfied. 
\hfill $\blacksquare$

\subsection{Proof of Lemma \ref{lem: estimate consistency}}
\label{appsec: lem estimate consistency}
\textbf{Proof:}
     We first consider the estimator for the expected performance. 
     %It is sufficient to prove for $\eta = 0$. To see this, let $\hat{\mu}_{i,t}^\eta$ denote the estimated expected performance at stage $t$ with drop rate $\eta$. Then $ \hat{\mu}_{i,t}^\eta = \frac{\hat{\mu}_{i,t}^0 - \eta \hat{\mu}_{i,t_\eta}^0}{1-\eta}$. We simply denote by $\hat{\mu}_{i,t}$ the performance estimator with drop state $\eta= 0$. 
     The idea is to partition the estimation error into two parts, accounting for SU and IU, separately. Specifically, to prove (a), rewrite 
     \begin{equation} \label{eq: consistency mu}
         \begin{aligned}
        \widehat{\mu}_{i,t} -\mu_i(\Bar{\mathbf{\theta}}) =  &\frac{1}{M_{i,t}} \sum_{\tau=1}^t \sum_{r=1}^{m_{i,\tau}} [X^r_{i}(\widehat{\mathbf{\theta}}_\tau) - \mu_i(\widehat{\mathbf{\theta}}_\tau) ]\\& + \frac{1}{M_{i,t}} \sum_{\tau=1}^t m_{i,\tau} [\mu_i(\widehat{\mathbf{\theta}}_\tau ) -\mu_i(\Bar{\mathbf{\theta}})].
    \end{aligned}
     \end{equation}
    
We fixed the design $i$ and drop this subscription for simplicity. Furthermore, denote by $X_\ell$ the $\ell$th simulation output for design $i$, $t_\ell$ be the stage that $X_\ell$ is generated, that is, $\{t| M_{i,t-1} < \ell \le M_{i,t} \}$. Then $X_\ell = X^{\ell - M_{t_\ell -1}}(\hat{\theta}_{t_\ell})$. Let $Z_\ell = X_\ell - \mu(\hat{\theta}_{t_\ell})$. Then we have $Z_\ell$ is a Martingale difference sequence (MDS). To see this, let $H_\ell$ be the sigma algebra generated by $\{X_r,\hat{\theta}_\tau, r =1,\ldots,\ell, \tau =1,\ldots, t_\ell\}$. 
Then $Z_r \in H_\ell$ for $r\le \ell$ and $\mathbb{E} [Z_{\ell+1} |H_\ell] = 0$. Hence, we have 
\begin{align*}
    &\mathbb{E}\left[Z_{\ell+1} | Z_1,\ldots,Z_\ell\right] \\
    = & \mathbb{E}\left[ \mathbb{E}[Z_{\ell+1} | H_\ell]| Z_1,\ldots,Z_\ell\right]\\
    =&0.
\end{align*}
Furthermore, since $|X_\ell| \le \bar{x}$, $|Z_\ell| \le 2\bar{x}$. 
By the Strong Law of Large Number (SLLN) for Martingale difference sequence (e.g., see Theorem 1 in \cite{csorgHo1968strong}), we have almost surely,
\begin{equation*}
    \frac{1}{\ell}\sum_{r=1}^\ell Z_r \rightarrow 0,
\end{equation*}
which proves the convergence of the first term in \eqref{eq: consistency mu}.

For the second term, since $\hat{\theta}_{s,t} \rightarrow \Bar{\theta}_s$ almost surely and $\mu$ is continuous in $\theta$, we know $\mu(\theta_\tl) \rightarrow \mu(\Bar{\theta})$ as $\ell \rightarrow \infty$. Then, almost surely, for an arbitrary $\varepsilon > 0$, there exists $T_\varepsilon$ such that for $t \ge T_\varepsilon$, $|\mu(\theta_t) - \mu(\Bar{\theta})| \le \varepsilon$. Then, almost surely,
\begin{align*}
    &\left|\frac{1}{M_{t}} \sum_{\tau=1}^t m_{\tau} [\mu(\widehat{\mathbf{\theta}}_\tau ) -\mu(\Bar{\mathbf{\theta}})]\right| \\
    \le& \left|\frac{1}{M_{t}} \sum_{\tau=T_\varepsilon+1}^{t} m_{\tau} [\mu(\widehat{\mathbf{\theta}}_\tau ) -\mu(\Bar{\mathbf{\theta}})]\right| + \left|\frac{1}{M_{t}} \sum_{\tau=1}^{T_\varepsilon} m_{\tau} [\mu(\widehat{\mathbf{\theta}}_\tau ) -\mu(\Bar{\mathbf{\theta}})]\right|\\
    \le &  \frac{M_t - M_{T_\varepsilon}}{M_t} \varepsilon + \left|\frac{1}{M_{t}} \sum_{\tau=1}^{T_\varepsilon} m_{\tau} [\mu(\widehat{\mathbf{\theta}}_\tau ) -\mu(\Bar{\mathbf{\theta}})]\right| \\
    \rightarrow &\varepsilon
\end{align*} 
as $t\rightarrow \infty$. This implies $\left|\frac{1}{M_{t}} \sum_{\tau=1}^t m_{\tau} [\mu(\widehat{\mathbf{\theta}}_\tau ) -\mu(\Bar{\mathbf{\theta}})]\right| \rightarrow 0$ as $t\rightarrow0$ almost surely, which further implies  $\left|\frac{1}{M_\tl} \sum_{\tau=1}^\tl m_{\tau} [\mu(\widehat{\mathbf{\theta}}_\tau ) -\mu(\Bar{\mathbf{\theta}})]\right| \rightarrow 0$ as $\ell \rightarrow \infty$.  This proves (a).

For (b), write 
\begin{align*}
    \hat{\sigma}^2_\ell - \sigma^2(\Bar{\theta}) = & \frac{1}{M_\tl -1 } \sum_{r=1}^{M_\tl} (X_\ell-{\mu}(\hat{\theta}_\tl))^2 \\
    =&\frac{1}{M_\tl -1 } \left\{ \underbrace{\left[ \sum_{r=1}^{M_\tl}(X_r - \mu(\hat{\theta}_{t_r}))^2 - \sigma^2(\hat{\theta}_{t_r}) \right]}_{I_1} + \underbrace{2\sum_{r=1}^{M_\tl} (X_r - \mu(\hat{\theta}_{t_r}))\mu(\hat{\theta}_{t_r})}_{I_2} \right.\\
    &\left.+\underbrace{\sum_{r=1}^{M_\tl} \mu(\hat{\theta}_{t_r})^2 - M_\tl(\hat{\mu}_\ell)^2}_{I_3} + \underbrace{\sum_{r=1}^{M_\tl} [ \sigma^2(\hat{\theta}_{t_r})-\sigma^2(\Bar{\theta})]}_{I_4}\right\}
\end{align*}
Both $I_1$ and $I_2$ is bounded MDSs. Hence, $\frac{1}{M_\tl -1} I_1 \rightarrow 0$ and $\frac{1}{M_\tl -1} I_2 \rightarrow 0$ almost surely as $\ell \rightarrow \infty$. For $I_3$, since $\mu(\hat{\theta}_\tl) \rightarrow \mu(\Bar{\theta})$, $\frac{1}{M_\tl} \sum_{r=1}^\ell \mu(\hat{\theta}_{t_r}) \rightarrow \mu(\Bar{\theta})$. We also have proved $\hat{\mu}_\ell \rightarrow  \mu(\Bar{\theta})$ almost surely. Hence $\frac{1}{M_\tl-1} I_3 \rightarrow \mu(\Bar{\theta}) - \mu(\Bar{\theta}) = 0$. For the last term $I_4$. Since $\hat{\theta}_\ell \rightarrow \Bar{\theta}$ and $\sigma(\theta)$ is continuous in $\theta$. With the same argument for proof of $\hat{\mu}_t$, we obtain $\frac{1}{M_\tl-1}I_4 \rightarrow 0$. Hence, we proved (b).

For (c), let $Y = \frac{\nabla_\theta q_{\theta}(\xi)}{q_{\theta}(\xi)} X(\xi)$ where $\xi \sim q_{\theta}$. Then $\nabla_{\theta} \mu(\theta) = \mathbb{E}[Y]$. Denote by $X_\tau^r = X(\xi_\tau^r) $. We can use the same technique as for $\hat{\mu}_t$. To be specific, write 
\begin{equation} \label{eq: consistency nabla mu}
    \begin{aligned}
     \widehat{\nabla \mu}_{t} -\nabla\mu(\Bar{\mathbf{\theta}}) =  &\frac{1}{M_{t}} \sum_{\tau=1}^t \sum_{r=1}^{m_{\tau}} \left[\frac{\nabla q_{\hat{\theta}_\tau}(\xi_\tau^r)}{q_{\hat{\theta}_\tau}(\xi_\tau^r)}X(\xi_\tau^r) - \nabla \mu(\widehat{\mathbf{\theta}}_\tau) \right]\\& + \frac{1}{M_{t}} \sum_{\tau=1}^t m_{\tau} [\nabla\mu(\widehat{\mathbf{\theta}}_\tau ) -\nabla\mu_i(\Bar{\mathbf{\theta}})].
\end{aligned}
\end{equation}

The first term is an MDS, which can be shown in the same way as the first term in \eqref{eq: consistency mu}. To use the SLLN for MDSs (Theorem 1 in \cite{csorgHo1968strong}, it is sufficient to show 
$ \mathbb{E}\left[ \left(\frac{\nabla q_{\hat{\theta}_\tau}(\xi_\tau^r)}{q_{\hat{\theta}_\tau}(\xi_\tau^r)}X^r_\tau  - \nabla \mu(\widehat{\mathbf{\theta}}_\tau)\right)^2 \right] \le 2\mathbb{E}\left[ \left(\frac{\nabla q_{\hat{\theta}_\tau}(\xi_\tau^r)}{q_{\hat{\theta}_\tau}(\xi_\tau^r)}X^r_\tau \right)^2\right] + 2\mathbb{E}\left[ \left(\nabla \mu(\widehat{\mathbf{\theta}}_\tau)\right)^2 \right]$ is bounded by a constant for all $\tau$ and $r$.
Since $|X^r_\tau| \le \bar{x}$ and $\mathbb{E}\left[\left(\frac{\nabla_\theta q_\theta(\xi)}{q_\theta(\xi)}\right)^2 \right] \le \bar{Q}$ by Assumption \ref{assump: consistency sigma nable}, we know $ \mathbb{E}\left[ \left(\frac{\nabla q_{\hat{\theta}_\tau}(\xi_\tau^r)}{q_{\hat{\theta}_\tau}(\xi_\tau^r)}X^r_\tau \right)^2\right] \le \bar{x}^2 \bar{Q}$. $ \mathbb{E}\left[ \left(\nabla \mu(\widehat{\mathbf{\theta}}_\tau)\right)^2 \right]$ is also bounded by a constant under assumption \ref{assump:CLT}.(iii). Hence, applying the SLLN for MDS, we obtain $\frac{1}{M_{t}} \sum_{\tau=1}^t \sum_{r=1}^{m_{\tau}} \left[\frac{\nabla q_{\hat{\theta}_\tau}(\xi_\tau^r)}{q_{\hat{\theta}_\tau}(\xi_\tau^r)}X^r_\tau - \nabla \mu(\widehat{\mathbf{\theta}}_\tau) \right] \rightarrow 0$ almost surely. For the second term in \eqref{eq: consistency nabla mu}, again since $\nabla \mu$ is continuous in $\theta$ and that $\hat{\theta}_\tau \rightarrow \Bar{\theta}$, with the same argument as for the second term in \eqref{eq: consistency mu}, we know almost surely, $ \frac{1}{M_{t}} \sum_{\tau=1}^t m_{\tau} [\nabla\mu(\widehat{\mathbf{\theta}}_\tau ) -\nabla\mu_i(\Bar{\mathbf{\theta}})] \rightarrow 0$. Hence, we proved (c).
\hfill $\blacksquare$\par

\subsection{Proof of Theorem \ref{thm: consistency}}
\label{appsec: thm consistency}

Let $\Omega$ be the sample space containing all the random factors which include all the input data collection and simulation outputs. Let $\omega \in \Omega$ be a sample path. In the following proof $\omega$ is always fixed. By Lemma \ref{lem: estimate consistency}, we know almost surely, $\widehat{\theta}_{s,t},\widehat{\mu}_{i,t},\widehat{\sigma}^2_{i,t},\widehat{\nabla \mu}_{i,t}$ converge. This is because if $N_{s,t}, M_{i,t} \rightarrow \infty$, then all estimates converge to their true value. Otherwise, some parameter estimates only have finite samples, then their estimate remains the same after some time stage. For instance, if $N_{s,t} = N_{s,\tau}$ for all $s$ and $t\ge \tau$, then $\widehat{\mathbf{\theta}}_t$ converges to $\widehat{\mathbf{\theta}}_\tau$ and $\widehat{\mu}_{i,t}$ converges to $\mu_i(\widehat{\mathbf{\theta}}_\tau)$ if $M_{i,t} \rightarrow \infty$. As a result, all estimators $\widehat{\theta}_t$, $\widehat{\Sigma}_{D,s,t}$,  $\widehat{\mu}_{i,t},\widehat{\sigma}_{i,t}$, $\widehat{g}_t(i,s)$ converge (under fixed $\omega$) to the limit denoted by $\theta^\infty$, $\Sigma^\infty_{D,s}$, $ \mu^\infty_i$, $\sigma^\infty_i$ and $g^\infty(i,s)$, respectively for all $i,s$ and $\widehat{b}_t = \arg\max_{i} \widehat{\mu}_{i,t}$ also converges to some $b^\infty$. 

We first prove $N_{s,t} \rightarrow \infty, \forall s\in\mathcal{S}$. 
We introduce the budget allocation iteration counter $\ell_j$. That is, the $\ell_j$th time that budget $U_j$ is allocated to collect some input data. Accordingly, let $N_s^{\ell_j}$ denote the number of input data collected for input distribution $s\in\mathcal{S}_j$.  Furthermore, we use $t^j_{\ell_j}$ to denote the stage where $\ell_j$th iteration of allocation of $U_j$ happens. To be specific, $(t^j_{\ell_j} - 1) \times U_j \le \sum_i \sum_{s\in\mathcal{S}_j} c_s (N_s^{\ell_j} - n_0) < t^j_{\ell_j}\times U_j $.

Since there exists $i\neq b$, $\partial_{\theta_s} \delta_{bi}(\theta) \neq 0$ for almost every $\theta$ and that both input distribution and output distribution are continuous distributions, for this $i$, we obtain ${g}^\infty(i,s) > 0$. Then, by the same argument of Lemma \ref{lem: convex input rate optimization}, with all unknown parameters replaced with their estimators, we know the computed optimal solution $\widehat{n}^t >0$, where the superscript $t$ refers to the stage counter. Since all estimators converge, and that the objective function of \eqref{eq: PAE optimize} is continuous in the aforementioned parameters, we have $\widehat{n}^t \rightarrow n^\infty$ for some 
$n^\infty >0$. Suppose for some $\bar{s} \in \mathcal{S}_j$, $N_{\bar{s},t}$ remains a constant $\underline{N}_{\bar{s}} < \infty $ for all large $t$. Take $\ell_j$ sufficiently large.
Since $(t^j_{\ell_j}-1)\times U_j<\sum_{s\in\mathcal{S}_j} c_s (N^{\ell_j}_{s}-n_0) \le t^j_{\ell_j}\times U_j$ and $\sum_{s\in\mathcal{S}_j} c_s \widehat{n}^{t^j_{\ell_j}}_s = U_j $, there exists $s' \in\mathcal{S}_j$, $c_{s'} n^{t^j_{\ell_j}}_{s'} < \frac{N^{\ell_j}_{s'}}{{t^j_{\ell_j}}-1}$. Since $N_{s'}^{\ell_j}$ increases only if $s'$ is collected at $\ell_j$, we can choose $\ell_j$ such that $s'$ is collected at $\ell_j$. Then we obtain
$$t^j_{\ell_j} \times \widehat{n}^{t^j_{\ell_j}}_{s'} - N^{\ell_j}_{s'} \le \frac{N^{\ell_j}_{s'}}{{t^j_{\ell_j}}-1} + \frac{t}{t-1}\sum_{s\in\mathcal{S}_j}c_s n_0 < C,$$
where $C$ is some constant. We also have 
$$t^j_{\ell_j} \times \widehat{n}^{t^j_{\ell_j}}_{\bar{s}} - N^{\ell_j}_{\bar{s}} \ge t^j_{\ell_j} \times \frac{1}{2} n^\infty_{\bar{s}} - \underline{N}_{\bar{s}} \rightarrow \infty,$$
where the first inequality holds since $\widehat{n}^t \rightarrow n^\infty$. This implies $s'$ cannot be chosen to collect data at iteration $\ell_j$, a contradiction. Hence, $N_{s,t}\rightarrow \infty,~ \forall s\in\mathcal{S}$.

We next prove the following lemma, where we assume the assumptions of Theorem \ref{thm: consistency} hold. Similar as for definition of input allocation iteration $\ell_j$,  We introduce the simulation budget allocation iteration counter $\ell$. That is, the $\ell$th time that budget $M$ is allocated to run the simulation. Accordingly, let $M_i^{\ell}$ to denote the number of simulations run for design $i$ before the $\ell$th iteration.  Furthermore, and we use $t_\ell$ to denote the stage where $\ell$th iteration of simulation budget allocation happens, i.e., $(t_{\ell} - 1) \times M  \le \sum_i d_i (M_i^{\ell}-m_0) < t_{\ell}\times M $.

\begin{lemma} \label{lem: lower hatb}
     $\lim\sup_{\ell \rightarrow \infty } \frac{M_{\hat{b}}^\ell}{\tl} >0 $ and there exists $i \neq \hat{b}$, $\lim\sup_{\ell \rightarrow \infty } \frac{M_i^\ell}{\tl} >0 $.
\end{lemma}
We first prove the first part. Suppose not, then $\lim_{\ell \rightarrow \infty } \frac{M_{\hat{b}}^\ell}{\tl} =0$. Since $\sum_{i\in\mathcal{K}} d_i M^\ell_i \ge (t_\ell-1)\times M$,
we know there exists $i_0 \neq \hat{b}$ and a constant $\underline{m} > 0$, such that $ \lim \sup _{\ell \rightarrow \infty } \frac{M_{i_0}^\ell}{\tl} \ge \underline{m} > 0$.  Since $\widehat{\sigma}^2_{i,t}$ converges, then there exists $\bar{\sigma} >\underline{\sigma} >0$ and $\bar{t}$, such that for $t>\bar{t}$, $\underline{\sigma}^2 <\hsigma^2_{i,t} < \bar{\sigma}^2 $. Also denote by $\underline{d} = \min_{i}d_i$ and $\bar{d} = \max_i d_i$. Let $\varepsilon $ be small such that $\varepsilon < \underline{m}\frac{(1-\varepsilon)}{(K-1)} \frac{\underline{\sigma}}{\bar{\sigma}}\sqrt{\frac{\underline{d}}{\bar{d}}}$; $t>\bar{t}$ such that $M_{\hatb,t} < \varepsilon   B t$ and there exists an iteration $\ell$ such that $\tl > \bar{t}$ and  $i_0$ is simulated at $\ell$th iteration. Then we have at iteration $\ell$,
\begin{align*}
    (M^\ell_{\hatb})^2 - \frac{\hsigma^2_{\hatb,\tl}}{d_\hatb} \sum_{i\neq \hatb} \frac{d_i (M^\ell_{i})^2}{\hsigma^2_{i,\tl}}  \le &  (M^\ell_{\hatb})^2  - \frac{\underline{\sigma}^2\underline{d}}{\bar{\sigma}^2\bar{d}}(M^\ell_{i_0})^2 \\
    < &\varepsilon^2   B^2 (\tl)^2 - \frac{\underline{d}}{\bar{d}}\frac{\underline{\sigma}^2}{\bar{\sigma}^2}\frac{(1-\varepsilon)^2}{(K-1)^2} \underline{m}^2  B^2(\tl)^2\\
    = &    B^2(\tl)^2[\varepsilon^2 - \underline{m}^2\frac{\underline{d}}{\bar{d}}\frac{(1-\varepsilon)^2}{(K-1)^2} \frac{\underline{\sigma}^2}{\bar{\sigma}^2} ] <0,
\end{align*}
a contradiction to $i_0$ is simulated at iteration $\ell$. 

For the second part, also prove by contradiction. Then we know for all $i\neq \hat{b}$, $\lim_{\ell\rightarrow\infty} \frac{M^\ell_i}{t} = 0$. 
We know there exists $\underline{m}_b>0$, such that $\limsup_{\ell\rightarrow\infty}  \frac{M_{\hatb}^\ell}{\tl} \ge \underline{m}_b > 0$.  Let $\varepsilon $ be small such that $\underline{m}_b^2 > (K-1)\varepsilon^2\frac{\bar{d}}{\underline{d}}\frac{\bar{\sigma}^2}{\underline{\sigma}^2}$; $t>\bar{t}$ such that $M_{i,t} < \varepsilon   B t$ for $i\neq \hatb$ and there exists a iteration $\ell$ such that $\tl > \bar{t}$ and $\hatb$ is simulated at $\ell$th iteration. Then
\begin{align*}
    (M^\ell_{\hatb})^2 - \frac{\hsigma^2_{\hatb,\tl}}{d_\hatb}\sum_{i\neq \hatb} \frac{d_i (M^\ell_{i})^2}{\hsigma^2_{i,\tl}}  > &M_{\hatb,\tl}^2 - \frac{\bar{d}}{\underline{d}}\frac{\bar{\sigma}^2}{\underline{\sigma}^2}\sum_{i\neq b}M_{i,t}^2 \\
    > & \bar{m}_\hatb^2   B^2 \tl^2 - (K-1)\varepsilon ^2\frac{\bar{d}}{\underline{d}}\frac{\underline{\sigma}^2}{\bar{\sigma}^2}  B^2t^2\\
    = &    B^2t^2[\underline{m}_\hatb^2 - (K-1)\varepsilon^2\frac{\bar{d}}{\underline{d}}\frac{\underline{\sigma}^2}{\bar{\sigma}^2} ]>0,
\end{align*}
a contradiction to $\hatb$ is simulated at $\ell$.
\hfill $\blacksquare$\par

With Lemma \ref{lem: lower hatb}, we know there exists $i_0 \neq \hatb$, such that $M_{i_0}^\ell \rightarrow \infty$.

Since we already prove $N_{s,t}   \rightarrow \infty, \forall s\in\mathcal{S}$, we have for all $s\in\mathcal{S}$,
$$ \frac{\hat{g}_t(i_0,s)}{N_{s,t}} \rightarrow 0.$$
Since we also have $\frac{\hsigma^2_{i_0,t}}{M_{i_0,t}} \rightarrow 0$, $\frac{\hsigma^2_{\hatb,t}}{M_{\hatb,t}} \rightarrow 0$, and that $\lim_{t\rightarrow} \hat{\delta}_{\hatb i_0, t} > 0$ by Lemma \ref{lem: estimate consistency} and Assumption \ref{assump: consistency sigma nable}. We know 

$$ \hat{G}_t(i_0) := \frac{(\widehat{\mu}_{\widehat{b},t}-\widehat{\mu}_{i_0,t})^2 }{ 2 \sum_{s\in\mathcal{S}}\frac{ \widehat{g}_t(i_0,s)}{N_{s,t}}  +  \frac{\widehat{\sigma}_{i_0,t}^2}{M_{i_0,t}}  +  \frac{\widehat{\sigma}^2_{\widehat{b},t}}{M_{\widehat{b},t}}} \rightarrow \infty.$$
This can happen if and only if 

$$ \hat{G}_t(i') \rightarrow \infty \ \forall i' \neq \hatb.$$
$\hat{G}_t(i') \rightarrow \infty$ further implies $M_{i',t} \rightarrow \infty$. This completes the proof of consistency.

\subsection{Proof of Theorem \ref{thm: optimality}}
\label{appsec: thm optimality}

First notice with almost the same proof for Theorem \ref{thm: consistency}, we can prove the consistency with modified rate Balance condition \eqref{thmeq: rate balance modified}. That is, we also have $N_{s,t} \rightarrow \infty$ and $M_{i,t} \rightarrow \infty$ almost surely as $t\rightarrow \infty$. Now we know for $t$ sufficiently large, $\hat{b} = b$. Hence, in the following proof, we replace $\hat{b}_t$ with $b$ when $t$ is chosen to be sufficiently large. Furthermore, as Lemma \ref{lem: estimate consistency} and Theorem \ref{thm: consistency} suggest, both $\hat{\mu}_{i,t}, \hsigma_{i,t}$ and $\hat{g}_t(s,i)$ will converge to its true value almost surely. Hence, almost surely, for $t$ large enough, we have there exist  $\bar{\sigma} > \underline{\sigma} > 0$, $\bar{g} > \underline{g} > 0$ such that $\bar{\sigma} \ge \hsigma_{i,t} \ge \underline{\sigma}$, $\bar{g} \ge \hat{g}_t(i,s) \ge \underline{g}$ for all $i,s$. We will use the notation in the following proof. Furthermore we also denote by $\bar{d} \ge d_i \ge \underline{d} > 0, \forall i$.  

We first prove the convergence of the input budgets allocation:

Since all estimators converge almost surely, we have $\widehat{n}^t \rightarrow n^*$ almost surely. Consider a fixed subgroup $\mathcal{S}_j$.  For an arbitrary small $\epsilon>0$, we have for $\ell_j$ sufficiently large, $|\widehat{n}^{t^j_{\ell_j}}_s - n^*_s| < \epsilon$, $\forall s\in \mathcal{S}_j$. Let 
$$A_j = \{s\in\mathcal{S}_j: t^j_{\ell_j} \times \widehat{n}^\tj_{s} - N^{\ell_j}_s < 0\}.$$
Fix a $s \in A_j$. Let $r$ be the last time before $\ell_j$ such that $s$ is collected at $r$. When $\ell_j$ is sufficiently large, $r$ is also sufficiently large because $s$ will be collected infinitely many times. We then have
$$ \widehat{n}^\tj_s - \frac{N^{\ell_j}_s}{\tj} = \underbrace{\widehat{n}^{t^j_r} - \frac{N^{r}_s}{t^j_r}}_{E_1} + \underbrace{\widehat{n}^\tj_s - \widehat{n}^{t^j_r}}_{E_2} + \underbrace{\frac{N^{r}_s}{t^j_r}- \frac{N^{\ell_j}_s}{\tj}}_{E_3}.$$
Since $$\sum_{s\in\mathcal{S}_j} c_s(N^r_s -n_0) \le t^j_r \times U_j = t^j_r\sum_{s\in\mathcal{S}_j} c_s \widehat{n}^{t_r^j}_s$$
and $s$ is collected at $r$,  we must have 
$$E_1 = \widehat{n}^{t^j_r} - \frac{N^{r}_s}{t^j_r}  - \frac{N^r_s}{t_r} \ge -\frac{n_0}{t_r} \ge -\epsilon,$$
for $r$ large enough. For $E_2$,
$$ |\widehat{n}^\tj_s - \widehat{n}^{t^j_r}| \le |\widehat{n}^\tj_s - n^*_s| + |n^* - \widehat{n}^{t^j_r}|<2\epsilon.$$
This implies $E_2 \ge -2\epsilon$. For $E_3$,
$$ \frac{N^{r}_s}{t^j_r}- \frac{N^{\ell_j}_s}{\tj} = \frac{N^{r}_s}{t^j_r}- \frac{N^{r}_s +1}{\tj} \ge \frac{N^{r}_s}{t^j_r}- \frac{N^{r}_s +1}{t^j_r + 1} = \frac{N^r_s - t^j_r}{t^j_r(t^j_r+1)} \ge -\frac{1}{t^j_r} \ge -\epsilon.$$
Hence, we obtain 
$$ \widehat{n}^\tj_s - \frac{N^{\ell_j}_s}{\tj} \ge -4\epsilon,~ \forall s \in A_j.$$

Furthermore, since 
$$ (\tj-1)\sum_{s\in\mathcal{S}_j} c_s \widehat{n}^{\tj}_s =(\tj-1) \times U_j < \sum_{s\in\mathcal{S}_j} c_s(N^{\ell_j}_s -n_0),$$
we obtain
\begin{align*}
    & \frac{\sum_{s\in\mathcal{S}_j }c_s n_0}{\tj-1}\\
    > &\sum_{s\in\mathcal{S}_j} c_s\left(\widehat{n}^\tj_s - \frac{N^{\ell_j}_s}{\tj-1}\right)\\
    =& \sum_{s\in\mathcal{S}_j} c_s\left(\widehat{n}^\tj_s - \frac{N^{\ell_j}_s}{\tj}\right) -\frac{N^\tj_s}{\tj(\tj-1)}\\
    =& \sum_{s\in A_j}c_s\left(\widehat{n}^\tj_s - \frac{N^{\ell_j}_s}{\tj}\right) + \sum_{s\in \mathcal{S}_j\backslash A_j}\left(\widehat{n}^\tj_s - \frac{N^{\ell_j}_s}{\tj}\right) -\frac{N^\tj_s}{\tj(\tj-1)} \\
    \ge & \sum_{s\in \mathcal{S}_j\backslash A_j}\left(\widehat{n}^\tj_s - \frac{N^{\ell_j}_s}{\tj}\right) -4|A_j|\epsilon -\frac{N^\tj_s}{\tj(\tj-1)}\\
    \ge & \sum_{s\in \mathcal{S}_j\backslash A_j}\left(\widehat{n}^\tj_s - \frac{N^{\ell_j}_s}{\tj}\right) -4|A_j|\epsilon -\epsilon\\
    \ge &\sum_{s\in \mathcal{S}_j\backslash A_j}\left(\widehat{n}^\tj_s - \frac{N^{\ell_j}_s}{\tj}\right) - (4|\mathcal{S}_j|+1)\epsilon,
\end{align*}
where the second last inequality holds when $\ell_j$ is sufficiently large. This then implies
$$\max_{s\in\mathcal{S}_j}\left(\widehat{n}^\tj_s - \frac{N^{\ell_j}_s}{\tj}\right)\le (4|\mathcal{S}_j|+1)\epsilon + \frac{\sum_{s\in\mathcal{S}_j }c_s n_0}{\tj-1} \le (4|\mathcal{S}_j|+2)\epsilon, $$
for $\ell_j$ large enough. Then, we obtain for any $s\in\mathcal{S}_j$,
$$ \left|\widehat{n}^\tj_s - \frac{N^{\ell_j}_s}{\tj}\right| \le \max\{4\epsilon,(4(|\mathcal{S}_j|+2)\epsilon\} =4(|\mathcal{S}_j|+2)\epsilon $$
Since this holds for all $j=1,\ldots,D$, we prove that $\forall s\in\mathcal{S}$,
$$\lim_{t\rightarrow \infty} \left(\widehat{n}^t_s - \frac{N_{s,t}}{t}\right) = 0 = n^*_s - \lim_{t\rightarrow\infty} \frac{N_{s,t}}{t}.$$ We finish the proof of the first part of Theorem \ref{thm: optimality}.

For the second part, we introduce the following several lemmas. 
\begin{lemma} \label{lem: i i'}
    $\lim\inf_{\ell\rightarrow\infty} \frac{M_i^\ell}{M^\ell_{i'}} > 0$, $\forall i\neq i' \neq b$.
\end{lemma}
\textbf{Proof:}
    Prove by contradiction. Suppose there exist $i$ and $i'$, such that $\lim\inf_{\ell\rightarrow\infty} \frac{M_i^\ell}{M^\ell_{i'}} = 0$. Note $ \frac{M_i^\ell}{M^\ell_{i'}}$ decreases if and only if $i'$ is simulated at $\ell$ and increases if and only if $i$ is simulated at $\ell$. Hence, for any positive constant $\varepsilon > 0$, we can find a sufficiently large $\ell$ such that $i'$ is sampled at $\ell$ and $ \frac{M_i^\ell}{M^\ell_{i'}} < \varepsilon$. Since $\hat{\mu}_{i,t}, \hat{\mu}_{i',t}, \hat{\mu}_{b,t}$ all converge to their true values almost surely. There exists $U>L>0$, such that for $\ell$ sufficiently large, $U > (\hat{\mu}_{b,\tl} - \hat{\mu}_{i,\tl})^2$, $L < (\hat{\mu}_{b,\tl} - \hat{\mu}_{i',\tl})^2$. Then,
    \begin{align}
        &\frac{(\hat{\mu}_{b,\tl}-\hat{\mu}_{i,\tl})^2}{\sum_s \frac{\hat{g}_\tl(i,s)}{N_{s,\tl}} + \frac{\hsigma_{i,\tl}^2}{M_i^\ell} + \frac{\hsigma_{b,\tl}^2}{M_{b}^\ell}} - \frac{(\hat{\mu}_{b,\tl}-\hat{\mu}_{i',\tl})^2}{\sum_s \frac{\hat{g}_\tl(i',s)}{N_{s,\tl}} + \frac{\hsigma_{i',\tl}^2}{M_{i'}^\ell} + \frac{\hsigma_{b,\tl}^2}{M_{b}^\ell}} \notag\\
        \le & \frac{U}{\sum_s \frac{\hat{g}_\tl(i,s)}{N_{s,\tl}} + \frac{\hsigma_{i,\tl}^2}{M_i^\ell} + \frac{\hsigma_{b,\tl}^2}{M_{b}^\ell}} - \frac{L}{\sum_s \frac{\hat{g}_\tl(i',s)}{N_{s,\tl}} + \frac{\hsigma_{i',\tl}^2}{M_{i'}^\ell} + \frac{\hsigma_{b,\tl}^2}{M_{b}^\ell}} \notag\\
        = & \frac{U \left(\sum_s \frac{\hat{g}_\tl(i',s)}{N_{s,\tl}} + \frac{\hsigma_{i',\tl}^2}{M_{i'}^\ell} + \frac{\hsigma_{b,\tl}^2}{M_{b}^\ell}\right) - L \left(\sum_s \frac{\hat{g}_\tl(i,s)}{N_{s,\tl}} + \frac{\hsigma_{i,\tl}^2}{M_i^\ell} + \frac{\hsigma_{b,\tl}^2}{M_{b}^\ell} \right)}{\left(\sum_s \frac{\hat{g}_\tl(i,s)}{N_{s,\tl}} + \frac{\hsigma_{i,\tl}^2}{M_i^\ell} + \frac{\hsigma_{b,\tl}^2}{M_{b}^\ell}\right)\left(\sum_s \frac{\hat{g}_\tl(i',s)}{N_{s,\tl}} + \frac{\hsigma_{i',\tl}^2}{M_{i'}^\ell} + \frac{\hsigma_{b,\tl}^2}{M_{b}^\ell}\right)}\notag\\
        \le & \frac{U \left(\sum_s \frac{\hat{g}_\tl(i',s)}{N_{s,\tl}} + \frac{\hsigma_{i',\tl}^2}{M_{i'}^\ell} + \frac{\hsigma_{b,\tl}^2}{M_{b}^\ell}\right) - L \frac{\hsigma_{i,\tl}^2}{M_i^\ell}}{\left(\sum_s \frac{\hat{g}_\tl(i,s)}{N_{s,\tl}} + \frac{\hsigma_{i,\tl}^2}{M_i^\ell} + \frac{\hsigma_{b,\tl}^2}{M_{b}^\ell}\right)\left(\sum_s \frac{\hat{g}_\tl(i',s)}{N_{s,\tl}} + \frac{\hsigma_{i',\tl}^2}{M_{i'}^\ell} + \frac{\hsigma_{b,\tl}^2}{M_{b}^\ell}\right)}\notag\\
        \le & \frac{U \left(\sum_s \frac{\hat{g}_\tl(i',s)}{N_{s,\tl}} + \frac{\hsigma_{i',\tl}^2}{M_{i'}^\ell} + \frac{\hsigma_{b,\tl}^2}{M_{b}^\ell}\right) - L \frac{\hsigma_{i,\tl}^2}{M_i^\ell}}{\left(\sum_s \frac{\hat{g}_\tl(i,s)}{N_{s,\tl}} + \frac{\hsigma_{i,\tl}^2}{M_i^\ell} + \frac{\hsigma_{b,\tl}^2}{M_{b}^\ell}\right)\left(\sum_s \frac{\hat{g}_\tl(i',s)}{N_{s,\tl}} + \frac{\hsigma_{i',\tl}^2}{M_{i'}^\ell} + \frac{\hsigma_{b,\tl}^2}{M_{b}^\ell}\right)} \notag \\
        \le & \frac{U \left(\sum_s \frac{\bar{g}}{N_{s,\tl}} + \frac{\bar{\sigma}}{M_{i'}^\ell} + \frac{\bar{\sigma}}{M_{b}^\ell}\right) - L \frac{\underline{\sigma}}{M_i^\ell}}{\left(\sum_s \frac{\hat{g}_\tl(i,s)}{N_{s,\tl}} + \frac{\hsigma_{i,\tl}^2}{M_i^\ell} + \frac{\hsigma_{b,\tl}^2}{M_{b}^\ell}\right)\left(\sum_s \frac{\hat{g}_\tl(i',s)}{N_{s,\tl}} + \frac{\hsigma_{i',\tl}^2}{M_{i'}^\ell} + \frac{\hsigma_{b,\tl}^2}{M_{b}^\ell}\right)} \label{eq: i i' 1}
    \end{align}
    Recall we have proved $\liminf_{t \rightarrow} \frac{N_{s,t}}{t} > 0, \forall s\in\mathcal{S}$. Hence, for a fixed $s$, there exists $C' > 0$, such that  for all $t$ large enough, $N_{s,t} \ge C' t \ge C' \frac{B}{d_{i'}} M_{i,t}$. Hence, for all $s\in\mathcal{S}$, there exists a constant $C$, such that for this sufficiently large $\ell$, we have $N_{s,\tl} \ge C M_{i'}^\ell$ for $s\in\mathcal{S}$. 

    Furthermore, we also have 
    $$ (M_{b}^\ell)^2 \ge \frac{\hsigma_{b,\tl}^2}{d_b} \sum_{k\neq b} \frac{d_k (M^\ell_{k})^2}{\hsigma_{k,\tl}^2} \ge \frac{\underline{\sigma} \underline{d}}{\bar{\sigma}\bar{d}} (M^\ell_{i'})^2.$$
    Hence, by choosing $C$ small than $ \sqrt{ \frac{\underline{\sigma} \underline{d}}{\bar{\sigma}\bar{d}} }$, we also have $ M_b^\ell \ge C M_{i'}^\ell $.
    
    Together, we have the numerator of \eqref{eq: i i' 1} can be upper bounded by
    \begin{align*}
        \frac{1}{ M_{i'}^\ell} \left( \frac{U|S|\bar{g}}{C} + \frac{\bar{\sigma}}{C}+\bar{\sigma} -L\underline{\sigma} \frac{M_{i'}^\ell}{M_i^\ell}\right) \le \frac{1}{ M_{i'}^\ell} \left( \frac{U|S|\bar{g}}{C} + \frac{\bar{\sigma}}{C}+\bar{\sigma} -L\underline{\sigma} \frac{1}{\varepsilon} \right) < 0
    \end{align*}
    for $\varepsilon$ sufficiently small. This indicates $i'$ cannot be sampled at $\ell$, a contradiction of how we choose $\ell$. Hence, we complete the proof.
\hfill $\blacksquare$\par
% %
\begin{lemma} \label{lem: b i}
    $\liminf_{\ell \rightarrow \infty} \frac{M_b^\ell}{M_i^\ell} > 0$, $\forall i\neq b$; $\liminf_{\ell \rightarrow \infty} \frac{M_{i}^\ell}{M_b^\ell } > 0$ , $\forall i\neq b$.
\end{lemma}
\textbf{Proof:}
    For the first part, prove by contradiction. Suppose not, then $\liminf_{\ell \rightarrow \infty} \frac{M_n^\ell}{M_i^\ell} = 0$. Again, for any positive constant $\varepsilon>0$, we can choose $\ell$ large enough such that $i$ is simulated at iteration $\ell$ and $ \frac{M_n^\ell}{M_i^\ell} < \varepsilon$. Then, we have
    \begin{align*}
        &(M_{b}^\ell)^2 - \frac{\hsigma_{b,\tl}^2}{d_b} \sum_{k\neq b} \frac{d_k (M^\ell_{k})^2}{\hsigma_{k,\tl}^2}  \\
        \le &   (M_{b}^\ell)^2 - \frac{\hsigma_{b,\tl}^2}{d_b} \frac{d_i (M^\ell_{i})^2}{\hsigma_{i,\tl}^2} \\ \le & (M^\ell_{i})^2 \left [\left(\frac{M_{b}^\ell}{M^\ell_{i}}\right)^2 - \frac{\underline{\sigma}^2}{d_b} \frac{d_i}{\bar{\sigma}^2} \right] \\ 
        \le & (M^\ell_{i})^2 \left [\varepsilon^2 - \frac{\underline{\sigma}^2}{d_b} \frac{d_i}{\bar{\sigma}^2} \right] <0,
    \end{align*}
    for $\varepsilon$ sufficiently small. This contradicts that $b$ will be sampled at $\ell$. Hence, we proved the first part. 
    
    For the second part, prove by contradiction. Suppose $\liminf_{\ell \rightarrow \infty} \frac{M_{i_0}^\ell}{M_b^\ell} = 0$. For any $\varepsilon>0$, we can find $\ell$ large enough such that $b$ is simulated at $\ell$ and $\frac{M_{i_0}^\ell}{M_b^\ell} < \varepsilon$. By Lemma \ref{lem: i i'}, we know there exists $C>0$, $ M_{i}^\ell \le CM_{i_0}^\ell, \forall i\neq i_0 \neq b$ for all $\ell$ sufficiently large. Then, we have
      \begin{align*}
        &(M_{b}^\ell)^2 - \frac{\hsigma_{b,\tl}^2}{d_b} \sum_{k\neq b} \frac{d_k (M^\ell_{k})^2}{\hsigma_{k,\tl}^2}  \\
        \ge &   (M_{b}^\ell)^2 - \frac{\bar{d}\bar{\sigma}^2}{\underline{d}\underline{\sigma}^2} \sum_{i \neq b} { (M^\ell_{i})^2}\\
        \ge & (M_{b}^\ell)^2 - \frac{\bar{d}\bar{\sigma}^2}{\underline{d}\underline{\sigma}^2} (KC^2+1) { (M^\ell_{i_0})^2} \\
        = & (M_{b}^\ell)^2 \left[ 1 - \frac{\bar{d}\bar{\sigma}^2}{\underline{d}\underline{\sigma}^2} (KC^2+1)  \left(\frac{M^\ell_{i_0}}{M_{b}^\ell}\right)^2 \right] \\
        \ge & (M_{b}^\ell)^2 \left[ 1 - \frac{\bar{d}\bar{\sigma}^2}{\underline{d}\underline{\sigma}^2} (KC^2+1)  \varepsilon^2 \right] > 0,
    \end{align*}
    for $\varepsilon$ sufficiently small. Hence, $b$ cannot be simulated at iteration $\ell$, which proves the second part.
    \hfill $\blacksquare$\par

With Lemma \ref{lem: i i'}-\ref{lem: b i}, we directly obtain the following result.
\begin{lemma} \label{lem: O(t)}
    $\liminf_{\ell \rightarrow\infty} \frac{M_{i}^\ell}{\ell} > 0, \forall 1\le i \le K$ or $\liminf_{t\rightarrow\infty} \frac{M_{i,t}}{t} > 0, \forall 1\le i \le K$.
\end{lemma}

The next Lemma \ref{lem:length} is a simple but useful result that we will use frequently in the following proof.
\begin{lemma} \label{lem:length}
Let $i$ be a fixed design. Suppose $i$ is sampled at iteration $r$. Let $\ell_r = \inf\{\ell>0: \mathbf{1}_{i}^{(r+\ell)} = 1\}$, where $\mathbf{1}_{i}^{(\ell)}=1$ represents design $i$ is simulated at iteration $r$. Hence $r+\ell_r$ is the next iteration $i$ will be sampled after $r$. Then we have $r<r+\ell_r = O(r) = O(t_r)$ \text{ almost surely }. Similarly, let $s$ be any fixed index for input distribution. Suppose $s$ is selected for data collection at iteration $r$ and $r+\ell_r$ the next iteration it is selected. Then, we also have $r+\ell_r = O(r) = O(t_r)$. 
\end{lemma}
\textbf{Proof:}
    The Lemma follows directly from Lemma \ref{lem: O(t)}.  For example, suppose there exists design $i_0$, such that for any $\varepsilon>0$, there exists $r$ large enough, such that $r < \varepsilon \ell_r$.
       Then, $\frac{M_{i}^{r+\ell_r}}{r+\ell_r} \le \frac{r+m_0+1}{r+\ell_r} \le 2 \varepsilon$. Here we take $r > m_0+1$. This contradicts that $\liminf_{\ell\rightarrow\infty} \frac{M_i^\ell}{\ell} > 0$ by Lemma \ref{lem: O(t)}.
\hfill $\blacksquare$\par

Now we are ready to prove the global balance optimality condition in Theorem \ref{thm: optimality}.\\
\textbf{Proof:}
By Lemma \ref{lem: estimate consistency}, we know all $\hsigma_{i,t},\hat{\mu}_{i,t},\hat{g}_t(i,s)$ converge to the true value as $t$ goes to infinity almost surely, $\forall i\in [K], s\in\mathcal{S}$. Also,we have proved $\lim_{t\rightarrow} \frac{N_{s,t}}{t} = n^*_s, ~\forall s\in\mathcal{S}$. Let $ \barm^\ell_i = \frac{M_i^\ell}{\tl}$ for $i\in [K]$. Then, for an arbitrary $\varepsilon>0$, almost surely there exists $L$ large enough, such that for iteration $\ell > L$, we have  $|\hsigma_{i,\tl}^2 - \sigma_i^2| \le \varepsilon $, $|\hat{\mu}_{i,\tl} - \mu_i| \le \varepsilon$ and $|\hat{g}_\tl (i,s) - g(i,s)| \le \varepsilon$.

 \textbf{(global Balance condition):}
 Let $r > L$ be a large iteration such that $b$ is sampled at iteration $r$. Then we must have 
$$ (\bar{m}_b^r)^2 - \frac{\hsigma_{b,t_r}^2}{d_b} \sum_{i\neq b} \frac{d_i (\bar{m}_i^r)^2}{\hsigma_{i,t_r}^2} \le 0.$$
Since $f(x,y) = \frac{x}{y}$ is Lipschitz continuous in a neighborhood of $(\sigma_b^2,\sigma_i^2)$ for each $i\neq b$ (the $\varepsilon$ is chosen to be small enough such that $\hsigma_{b,t_r}^2$ and $\hsigma_{i,t_r}^2 $ belong to the neighborhood) and that $\bar{m}_{i,\ell} \le \frac{B}{d_i} + m_0$, we have there exists a constant $C$ independent of $r$ and $\varepsilon$, such that 
$$ (\bar{m}_b^r)^2 - \frac{\sigma_{b}^2}{d_b} \sum_{i\neq b} \frac{d_i (\bar{m}_i^r)^2}{\sigma_{i}^2} \le (\bar{m}_b^r)^2 - \frac{\hsigma_{b,t_r}^2}{d_b} \sum_{i\neq b} \frac{d_i (\bar{m}_i^r)^2}{\hsigma_{i,t_r}^2} + C\varepsilon \le C\varepsilon.$$
This implies, $b$ cannot be selected to simulate if 
$GB(\bar{m}^\ell):= (\bar{m}_b^\ell)^2 - \frac{\sigma_{b}^2}{d_b} \sum_{i\neq b} \frac{d_i (\bar{m}_i^\ell)^2}{\sigma_{i}^2} \ge C\varepsilon$ for any $\ell > L$. Notice when $GB(\bar{m}^\ell) \ge 0$, it can increase only if $b$ is simulated. And that simulating $b$ once only increases $GB(\bar{m}^\ell)$ by $O(1/\ell)$ for any $\ell > L$. This implies 
$$ GB(\bar{m}^\ell) \le C\varepsilon + O(1/\ell), \forall \ell > L. $$
Take $\ell \rightarrow \infty$, we know
$$\limsup_{\ell \rightarrow\infty} GB(\bar{m}^\ell) \le C\varepsilon.$$
Furthermore, by the arbitrary choice of $\varepsilon$, we obtain
$$\limsup_{\ell \rightarrow\infty} GB(\bar{m}^\ell) \le 0.$$

To prove $\liminf_{\ell \rightarrow\infty} GB(\bar{m}^\ell) \ge 0$, we can follow a similar idea. Let $r > L$ be an iteration where some sub-optimal design $k \neq b$ is simulated. Then we have 
$$ (\bar{m}_b^r)^2 - \frac{\hsigma_{b,t_r}^2}{d_b} \sum_{i\neq b} \frac{d_i (\bar{m}_i^r)^2}{\hsigma_{i,t_r}^2} \ge 0.$$
And  
$$ (\bar{m}_b^r)^2 - \frac{\sigma_{b}^2}{d_b} \sum_{i\neq b} \frac{d_i (\bar{m}_i^r)^2}{\sigma_{i}^2} \ge (\bar{m}_b^r)^2 - \frac{\hsigma_{b,t_r}^2}{d_b} \sum_{i\neq b} \frac{d_i (\bar{m}_i^r)^2}{\hsigma_{i,t_r}^2} - C\varepsilon \ge - C\varepsilon.$$
This implies any sub-optimal design cannot be selected to simulate if 
$GB(\bar{m}^\ell):= (\bar{m}_b^\ell)^2 - \frac{\sigma_{b}^2}{d_b} \sum_{i\neq b} \frac{d_i (\bar{m}_i^\ell)^2}{\sigma_{i}^2} \le - C\varepsilon$ for any $\ell > L$. Notice when $GB(\bar{m}^\ell)\le0$, it can decrease only if some $ i\neq b $ is simulated. And that simulating $i$ once only decreases $GB(\bar{m}^\ell)$ by $O(1/\ell)$ for any $\ell > L$. This implies 
$$ GB(\bar{m}^\ell) \ge C\varepsilon - O(1/\ell), \forall \ell > L. $$
Take $\ell \rightarrow \infty$, we know
$$\liminf_{\ell \rightarrow\infty} GB(\bar{m}^\ell) \ge- C\varepsilon.$$
Again, by the arbitrary choice of $\varepsilon$, we obtain
$$\liminf_{\ell \rightarrow\infty} GB(\bar{m}^\ell) \ge 0.$$
This completes the proof of \eqref{eq: global balance converge}.

\hfill $\blacksquare$\par
To prove the (modified) local Balance condition in Theorem \ref{thm: optimality}, we still need several lemmas.

\begin{lemma} \label{lem: num o(t)}
\begin{enumerate}
    \item Let $r$ denote some iteration where a sub-optimal design is simulated. Let $r+\ell_r$ be the next iteration where a sub-optimal design is first simulated after iteration $r$. Then, $\ell_r = o(r)$ almost surely.
    \item Let $r$ denote some iteration where the optimal design is simulated. Let $r+\ell_r$ be the next iteration where the optimal design is first simulated after iteration $r$. Then, $\ell_r = o(r)$ almost surely.
\end{enumerate}
    
\end{lemma}
\textbf{Proof:}
    Suppose 1. does not hold. Then there exists a constant $C>0$, such that for any $R>0$, there exists $r > R$ such that $r+\ell_r \ge (1+C) r$.
    During $r$ and $r+\ell_r$,$b$ is always chosen to be simulated. Hence,  $M_b^{r+\ell_r} - M_b^r \ge C r$.  Let $r'=r+\ell_r-1$ be the last iteration before $r+\ell_r$ that $b$ is simulated.  Then,  $\barm_b^{r} = \barm_b^{r'} \frac{M_b^{r}}{M_b^{r'}} \frac{r'}{r} \le \frac{\barm_b^{r'}}{1+C} \frac{r'}{r} $ and $\barm_i^{r'} = \barm_i^r \frac{r}{r'}+ \frac{1}{r'}$. 
    Furthermore by \eqref{eq: global balance converge}, we know for any $\varepsilon > 0$, we can set $R$ large enough such that 
    $$ |(\bar{m}_b^r)^2 - \frac{\sigma_{b}^2}{d_b} \sum_{i\neq b} \frac{d_i (\bar{m}_i^r)^2}{\sigma_{i}^2}| \le \varepsilon$$
    and 
    \begin{equation} \label{eq: lem num o(t) 1}
        |(\bar{m}_b^{r'})^2 - \frac{\sigma_{b}^2}{d_b} \sum_{i\neq b} \frac{d_i (\bar{m}_i^{r'})^2}{\sigma_{i}^2}| \le \varepsilon
    \end{equation}
    However,
    \begin{align*}
        &(\bar{m}_b^{r'})^2 - \frac{\sigma_{b}^2}{d_b} \sum_{i\neq b} \frac{d_i (\bar{m}_i^{r'})^2}{\sigma_{i}^2} \\
        \ge &(\frac{r}{r'})^2 \left[(1+C)^2 (\bar{m}_b^r)^2 - \frac{\sigma_{b}^2}{d_b} \sum_{i\neq b} \frac{d_i (\bar{m}_i^r)^2}{\sigma_{i}^2} - O(\frac{1}{r})\right] \\
        \ge & (\frac{r}{r'})^2 [ ((C^2)+2C) (\barm_b^r)^2 - 2\varepsilon],
    \end{align*}
where $r$ is set to be large enough such that the term $O(\frac{1}{r})$ is bounded by $\varepsilon$. Let $C_1 = (C^2)+2C)$ and by Lemma \ref{lem:length}, there exists $C_2>0$ such that $\frac{r}{r'} \ge C_2$ for large $r$. Also by Lemma \ref{lem: O(t)} there exists $\underline{m}$ such that $\barm_b^\ell \ge \underline{m}$ for all large $\ell$.  Then
$$ (\bar{m}_b^{r'})^2 - \frac{\hsigma_{b,t_r}^2}{d_b} \sum_{i\neq b} \frac{d_i (\bar{m}_i^{r'})^2}{\hsigma_{i,t_r}^2} \ge C^2_2(C^2_1 \underline{m}^2 - 2\varepsilon).$$
By choosing $2\varepsilon < \frac{C^2_2 C^2_1 \underline{m}^2}{1+C^2_2}$, we obtain
 $$|(\bar{m}_b^{r'})^2 - \frac{\sigma_{b}^2}{d_b} \sum_{i\neq b} \frac{d_i (\bar{m}_i^{r'})^2}{\sigma_{i}^2}| > \varepsilon,$$
 a contradiction to \eqref{eq: lem num o(t) 1}.
 2. and 3. can be proved in a similar manner.
 \hfill $\blacksquare$\par

 \begin{lemma} \label{lem: local converge 1}
     For a fixed sub-optimal design $i_0$, let $r$ denote an iteration where $i_0$ is simulated. Let $r'$ be the first iteration after $r$ such that $i_0$ is simulated. If for some $C>0$, there exists infinitely many $r$ and $r'$, such that $\frac{r'}{r} \ge 1+C$, then for such sufficiently large $r$, there exists $C_1 > 0$ (depends on $C$ but not on $r$), an iteration $r<u<r'$ and another sub-optimal design $i_1$, such that $m_{i_0}^u \ge (1+C_1) m_{i_0}^r$.
 \end{lemma}
 \textbf{Proof:}
     Notice for any iteration $\ell$, the following relations hold:
     $$  M\le  \sum_{i\in [K]} d_i \barm_i^\ell \le M + \frac{M+\sum_{i\in\mathcal{K}}d_im_0}{\tl}. $$
     Hence we have 
     $$  \sum_{i\in [K]} d_i (\barm_i^{r'} - \barm_i^r) \ge -\frac{M+\sum_{i\in\mathcal{K}}d_im_0}{t_r}. $$
     Then, since $\barm_{i_0}^{r'} = \barm_{i_0}^r \frac{r}{r'} \le \barm_{i_0}^r \frac{1}{1+C} $, $\barm_{i_0}^{r} - \barm_{i_0}^{r'} \ge C \barm_{i_0}^{r'} \ge C \underline{m} $ , where $\underline{m} \le \barm_i^\ell$ for all $i\in[K]$ and $\ell$ sufficiently large by Lemma \ref{lem: O(t)}. 
     Then, we obtain
     $$  \sum_{i\neq i_0} d_i (\barm_i^{r'} - \barm_i^r) \ge C\underline{m}-\frac{M+\sum_{i\in\mathcal{K}}d_im_0}{t_r} \ge \frac{C\underline{m}}{2}$$
     for $r$ sufficiently large. Hence, by pigeonhole principle, there exists $i \neq i_0, \barm_i^{r'} -\barm_i^r \ge \frac{C\underline{m}}{d_iK} $.
     
     \textbf{(i).} If there exists such $i \neq b \neq i_0$, then the lemma holds as we $\barm_i^\ell$ is always upper bounded by some $\nu>0$ and we have $\frac{\barm_i^{r'}}{\barm_i^r} \ge (1+ \frac{C\underline{m}}{d_i K \nu}) =:1+C_1$. 
     
     \textbf{(ii).} Otherwise, if such $i = b$, let $C_2 = \frac{C\underline{m}}{d_b K}$. By \eqref{eq: global balance converge}, for an arbitrary $\varepsilon > 0$, 
     $$ |(\bar{m}_b^r)^2 - \frac{\sigma_{b}^2}{d_b} \sum_{i\neq b} \frac{d_i (\bar{m}_i^r)^2}{\sigma_{i}^2}| \le \varepsilon,  \quad |(\bar{m}_b^{r'})^2 - \frac{\sigma_{b}^2}{d_b} \sum_{i\neq b} \frac{d_i (\bar{m}_i^{r'})^2}{\sigma_{i}^2}| \le \varepsilon.$$
     This implies 
     $$ (\bar{m}_b^{r'})^2 - \frac{\sigma_{b}^2}{d_b} \sum_{i\neq b} \frac{d_i (\bar{m}_i^{r'})^2}{\sigma_{i}^2} - \left [ (\bar{m}_b^r)^2 - \frac{\sigma_{b}^2}{d_b} \sum_{i\neq b} \frac{d_i (\bar{m}_i^r)^2}{\sigma_{i}^2}\right] \le 2\varepsilon.$$
     Since $\barm_{i_0}^{r'} = \barm_{i_0}^r \frac{r}{r'} + \frac{1}{r'} \le \barm_{i_0} + \frac{1}{r'}, $ $\frac{\sigma_{b}^2}{d_b} \frac{d_{i_0} (\bar{m}_{i_0}^{r'})^2}{\sigma_{i_0}^2} - \frac{\sigma_{b}^2}{d_b} \frac{d_{i_0} (\bar{m}_{i_0}^{r})^2}{\sigma_{i_0}^2}  \le \varepsilon$ for all sufficiently large $r$. We further obtain 
     $$  \frac{\sigma_{b}^2}{d_b} \sum_{i\neq b \neq i_0} \frac{d_i }{\sigma_{i}^2} \left((\bar{m}_i^{r'})^2 - (\bar{m}_i^{r})^2\right) \ge (\bar{m}_b^{r'})^2  -(\bar{m}_b^r)^2 - 3\varepsilon $$
     Since $ (\bar{m}_b^{r'})^2  -(\bar{m}_b^r)^2 \ge 2C_2\underline{m} $, by choosing $3\varepsilon < C_2\underline{m}$, we have $$  \frac{\sigma_{b}^2}{d_b} \sum_{i\neq b \neq i_0} \frac{d_i }{\sigma_{i}^2} \left((\bar{m}_i^{r'})^2 - (\bar{m}_i^{r})^2\right) \ge C_2\underline{m}.$$
     Again by the pigeonhole principle, there exists $i_1 \neq i_0 \neq b$, such that 
     $  (\bar{m}_{i_1}^{r'})^2 - (\bar{m}_{i_1}^{r})^2 \ge \frac{d_b C_2\underline{m}\sigma_{i_0}^2}{\sigma_b^2 (K-2)d_{i_1}}$,
     let $\nu \ge \mu_i^\ell, \forall i\in [K]$, then 
     $\bar{m}_{i_1}^{r'} - \bar{m}_{i_1}^{r} \ge \frac{d_b C_2\underline{m}\sigma_{i_0}^2}{2\nu \sigma_b^2 (K-2)d_{i_1}} = : C'_1$. Let $u<r'$ be the last time $i_1$ is simulated before $r'$, then $\frac{\bar{m}_{i_1}^{u}}{\bar{m}_{i_1}^{r}}  \ge 1 +\frac{C'_1}{\nu} =:1+C_1$. The lemma also holds true. 
 \hfill $\blacksquare$\par

\begin{lemma} \label{lem: local converge 2}
    For a fixed sub-optimal design $i_0$. Let $r$ denote an iteration where $i_0$ is simulated. Let $r'+1$ be the first iteration after $r$ such that $i_0$ is simulated. Then $r'-r = o(r)$ almost surely.
\end{lemma}
\textbf{Proof:}
    Prove by contradiction. Then there exists $C>0$, such that there are infinitely many $r$ and $r'$, such that $r'-r \ge Cr$. By Lemma \ref{lem: local converge 1}, there exists $C_1> 0 $, $i_1 \neq i_0 \neq b$ and an iteration $u$ such that $i_1$ is simulated at $u$ and $\frac{\barm_{i_1}^u}{\barm_{i_1}^r} \ge 1+C_1 $.
    Since $i_0$ is simulated at $r$,
    $$ \frac{(\hat{\mu}_{b,t_r}-\hat{\mu}_{i_0,t_r})^2 }{ 2 \sum_{s\in\mathcal{S}}\frac{ \hat{g}_{t_r}(i_0,s)}{\bar{n}_{s,t_r}}  +  \frac{\hsigma_{i_0,t_r}^2}{\bar{m}^r_{i_0}} }  \le \frac{(\hat{\mu}_{b,t_r}-\hat{\mu}_{i_1,t_r})^2 }{ 2 \sum_{s\in\mathcal{S}}\frac{ \hat{g}_{t_r}(i_1,s)}{\bar{n}_{s,t_r}}  +  \frac{\hsigma_{i_1,{t_r}}^2}{\bar{m}^r_{i_1} }}. $$
    Or equivalently
    \begin{align*}
        (\hat{\mu}_{b,t_r}-\hat{\mu}_{i_0,t_r})^2 \left(2 \sum_{s\in\mathcal{S}}\frac{ \hat{g}_{t_r}(i_1,s)}{\bar{n}_{s,t_r}}  +  \frac{\hsigma_{i_1,{t_r}}^2}{\bar{m}^r_{i_1} }\right) - (\hat{\mu}_{b,t_r}-\hat{\mu}_{i_1,t_r})^2 \left(  2 \sum_{s\in\mathcal{S}}\frac{ \hat{g}_{t_r}(i_0,s)}{\bar{n}_{s,t_r}}  +  \frac{\hsigma_{i_0,t_r}^2}{\bar{m}^r_{i_0}}\right) \le 0
    \end{align*}
    By the consistency of all estimators and Assumption \ref{assump: asymptotic optimality}, for an arbitrary $\varepsilon>0$, almost surely there exists $L$ large enough, such that for iteration $\ell > L$, we have  $|\hsigma_{i,\tl}^2 - \sigma_i^2| \le \varepsilon $, $|\hat{\mu}_{i,\tl} - \mu_i| \le \varepsilon$ and $|\hat{g}_\tl (i,s) - g(i,s)| \le \varepsilon$, $|\bn_{s,\tl} - n^*_s| \le \varepsilon$. We choose $r > L$, then there exists $C_2 > 0$ independent of $r$ and $\varepsilon$, such that 
      \begin{align*}
        &(\hat{\mu}_{b,t_u}-\hat{\mu}_{i_0,t_u})^2 \left(2 \sum_{s\in\mathcal{S}}\frac{ \hat{g}_{t_u}(i_1,s)}{\bar{n}_{s,t_u}}  +  \frac{\hsigma_{i_1,{t_u}}^2}{\bar{m}^u_{i_1} }\right) - (\hat{\mu}_{b,t_u}-\hat{\mu}_{i_1,t_u})^2 \left(  2 \sum_{s\in\mathcal{S}}\frac{ \hat{g}_{t_u}(i_0,s)}{\bar{n}_{s,t_u}}  +  \frac{\hsigma_{i_0,t_u}^2}{\bar{m}^u_{i_0}}\right) \\
        \le &(\hat{\mu}_{b,t_r}-\hat{\mu}_{i_0,t_r})^2 \left(2 \sum_{s\in\mathcal{S}}\frac{ \hat{g}_{t_r}(i_1,s)}{\bar{n}_{s,t_r}}  +  \frac{\hsigma_{i_1,{t_r}}^2}{\bar{m}^u_{i_1} }\right) - (\hat{\mu}_{b,t_r}-\hat{\mu}_{i_1,t_r})^2 \left(  2 \sum_{s\in\mathcal{S}}\frac{ \hat{g}_{t_r}(i_0,s)}{\bar{n}_{s,t_r}}  +  \frac{\hsigma_{i_0,t_r}^2}{\bar{m}^u_{i_0}}\right) + C_2\varepsilon \\
        \le &(\hat{\mu}_{b,t_r}-\hat{\mu}_{i_0,t_r})^2 \left(2 \sum_{s\in\mathcal{S}}\frac{ \hat{g}_{t_r}(i_1,s)}{\bar{n}_{s,t_r}}  +  \frac{\hsigma_{i_1,{t_r}}^2}{\bar{m}^r_{i_1} } \frac{\bar{m}^r_{i_1}}{\bar{m}^u_{i_1}}\right) - (\hat{\mu}_{b,t_r}-\hat{\mu}_{i_1,t_r})^2 \left(  2 \sum_{s\in\mathcal{S}}\frac{ \hat{g}_{t_r}(i_0,s)}{\bar{n}_{s,t_r}}  +  \frac{\hsigma_{i_0,t_r}^2}{\bar{m}^r_{i_0}}\right) + O(\frac{1}{r}) + C_2\varepsilon\\
        \le &(\hat{\mu}_{b,t_r}-\hat{\mu}_{i_0,t_r})^2 \left(2 \sum_{s\in\mathcal{S}}\frac{ \hat{g}_{t_r}(i_1,s)}{\bar{n}_{s,t_r}}  +  \frac{\hsigma_{i_1,{t_r}}^2}{\bar{m}^r_{i_1} } \frac{1}{1+C_1}\right) - (\hat{\mu}_{b,t_r}-\hat{\mu}_{i_1,t_r})^2 \left(  2 \sum_{s\in\mathcal{S}}\frac{ \hat{g}_{t_r}(i_0,s)}{\bar{n}_{s,t_r}}  +  \frac{\hsigma_{i_0,t_r}^2}{\bar{m}^r_{i_0}}\right) + 2C_2\varepsilon\\
        =&(\hat{\mu}_{b,t_r}-\hat{\mu}_{i_0,t_r})^2 \left(2 \sum_{s\in\mathcal{S}}\frac{ \hat{g}_{t_r}(i_1,s)}{\bar{n}_{s,t_r}}  +  \frac{\hsigma_{i_1,{t_r}}^2}{\bar{m}^r_{i_1} } \right) - (\hat{\mu}_{b,t_r}-\hat{\mu}_{i_1,t_r})^2 \left(  2 \sum_{s\in\mathcal{S}}\frac{ \hat{g}_{t_r}(i_0,s)}{\bar{n}_{s,t_r}}  +  \frac{\hsigma_{i_0,t_r}^2}{\bar{m}^r_{i_0}}\right) \\
        &+ 2C_2\varepsilon - \frac{C_1}{1+C_1}(\hat{\mu}_{b,t_r}-\hat{\mu}_{i_0,t_r})^2\frac{\hsigma_{i_1,{t_r}}^2}{\bar{m}^r_{i_1}} \\
        \le &2C_2\varepsilon - \frac{C_1}{1+C_1}(\hat{\mu}_{b,t_r}-\hat{\mu}_{i_0,t_r})^2\frac{\hsigma_{i_1,{t_r}}^2}{\bar{m}^r_{i_1}}.
    \end{align*}
    Let $C_3$ be a constant such that $0<C_3 \le (\hat{\mu}_{b,t_r}-\hat{\mu}_{i_0,t_r})^2\frac{\hsigma_{i_1,{t_r}}^2}{\bar{m}^r_{i_1}}$. $C_3$ is independent of $r$ since $\hat{\mu}_{i}^\ell,\hsigma_i^\ell$ converge and that $\barm_i^\ell \ge \underline{m} > 0$.
    Hence, by choosing $\varepsilon < \frac{C_1 C_3}{2C_2(1+C_1)}$, we haved 
    $$(\hat{\mu}_{b,t_u}-\hat{\mu}_{i_0,t_u})^2 \left(2 \sum_{s\in\mathcal{S}}\frac{ \hat{g}_{t_u}(i_1,s)}{\bar{n}_{s,t_u}}  +  \frac{\hsigma_{i_1,{t_u}}^2}{\bar{m}^u_{i_1} }\right) - (\hat{\mu}_{b,t_u}-\hat{\mu}_{i_1,t_u})^2 \left(  2 \sum_{s\in\mathcal{S}}\frac{ \hat{g}_{t_u}(i_0,s)}{\bar{n}_{s,t_u}}  +  \frac{\hsigma_{i_0,t_u}^2}{\bar{m}^u_{i_0}}\right) < 0,$$
    this implies $i_1$ cannot be simulated at iteration $u$, a contradiction to the definition of $u$. 
\hfill $\blacksquare$\par

Now we can prove the local Balance condition in Theorem \ref{thm: optimality}.\\
 \textbf{Proof:}
     \textbf{(local Balance):} 
     For an arbitrary $\varepsilon>0$, almost surely there exists $L$ large enough, such that for iteration $\ell > L$, we have  $|\hsigma_{i,\tl}^2 - \sigma_i^2| \le \varepsilon $, $|\hat{\mu}_{i,\tl} - \mu_i| \le \varepsilon$ and $|\hat{g}_\tl (i,s) - g(i,s)| \le \varepsilon$. Also, by Lemma \ref{lem: local converge 2}, we can set $L$ large enough such that 
     for any $r>L$ be some iteration such that design $i$ is sampled at $r$ and $r'>r$ the first iteration after $r$ such that $i$ is simulated again, $r'-r \le \varepsilon r$. Arbitrary pick iteration $\ell > L$, and find $r< \ell < r'$ where $i$ is simulated at $r$ and $r'$, $r'-r \le \varepsilon r$.
     For $i'\neq i \neq b$, since $\ell -r \le \varepsilon r$, we have 
$$\barm_{i'}^\ell -\barm_{i'}^r  \le \frac{M_{i'}^r+\varepsilon r}{r+\varepsilon r} - \frac{M_{i'}^r}{r} \le \frac{\varepsilon r}{r} = \varepsilon, \text{ and } \barm_{i'}^\ell -\barm_{i'}^r \ge\frac{M_{i'}^r}{r+\varepsilon r} - \frac{M_{i'}^r}{r} = -\frac{\varepsilon r M_{i'}^r}{r^2(1+\varepsilon)} \ge -\varepsilon$$
Since $i$ is simulated at $r$,
$$ \frac{(\hat{\mu}_{b,t_r}-\hat{\mu}_{i,t_r})^2 }{ 2 \sum_{s\in\mathcal{S}}\frac{ \hat{g}_{t_r}(i,s)}{\bar{n}_{s,t_r}}  +  \frac{\hsigma_{i,t_r}^2}{\bar{m}^r_{i}} } - \frac{(\hat{\mu}_{b,t_r}-\hat{\mu}_{i',t_r})^2 }{ 2 \sum_{s\in\mathcal{S}}\frac{ \hat{g}_{t_r}(i',s)}{\bar{n}_{s,t_r}}  +  \frac{\hsigma_{i',{t_r}}^2}{\bar{m}^r_{i'} }} \le 0 $$
By Lemma \ref{lem: O(t)}, we know almost surely, there exists $\nu>0$, $\bar{m}_i^\ell \ge \nu >0$ for all $i\in [K]$ and sufficiently large $\ell$.
Let $f(x,y,z,w,m) = \frac{x}{2\sum_{s\in\mathcal{S}} \frac{y_s}{z_s } + \frac{w}{m}}$. Then  $f$ is Lipschitz in a neighborhood of $(x,y,z,w) = ((\mu_b-\mu_i)^2, (g(i,s))_{s\in\mathcal{S}}, (n^*_s)_{s\in\mathcal{S}},\sigma_i^2)$ uniformly for $ m \ge \nu$. That is, there exists a constant $C$, 
$$|f((\mu_b-\mu_i)^2, (g(i,s))_{s\in\mathcal{S}}, (n^*_s)_{s\in\mathcal{S}},\sigma_i^2, \bar{m}_i^\ell) - f((\hat{\mu}_{b,\tl}-\hat{\mu}_{i,\tl})^2, (\hat{g}_{\tl}(i,s))_{s\in\mathcal{S}}, (\bn_{s,\tl})_{s\in\mathcal{S}},\hsigma_{i,\tl}^2,\barm_i^\ell)| \le C \varepsilon $$
for all sufficiently small $\varepsilon, i\in [K], \ell > L$. Then, we obtain
\begin{align*}
&\frac{({\mu}_{b}-{\mu}_{i})^2 }{ 2 \sum_{s\in\mathcal{S}}\frac{ {g}(i,s)}{n^*_s}  +  \frac{\sigma_{i}^2}{\bar{m}^r_{i}} } - \frac{({\mu}_{b}-{\mu}_{i'})^2 }{ 2 \sum_{s\in\mathcal{S}}\frac{ {g}(i',s)}{n^*_s}  +  \frac{\sigma^2_{i'}}{\bar{m}^r_{i'} }}  \\
\le &\frac{(\hat{\mu}_{b,t_r}-\hat{\mu}_{i,t_r})^2 }{ 2 \sum_{s\in\mathcal{S}}\frac{ \hat{g}_{t_r}(i,s)}{\bar{n }_{s,t_r}}  +  \frac{\hsigma_{i,t_r}^2}{\bar{m}^r_{i}} } - \frac{(\hat{\mu}_{b,t_r}-\hat{\mu}_{i',t_r})^2 }{ 2 \sum_{s\in\mathcal{S}}\frac{ \hat{g}_{t_r}(i',s)}{\bar{n}_{s,t_r}}  +  \frac{\sigma_{i',{t_r}}^2}{\bar{m}^r_{i'} }}  + C\varepsilon \\
\le & C\varepsilon
\end{align*}
Since $|\barm_i^\ell - \barm_i^r| = O(\frac{1}{r}) \le \varepsilon$ for large $r$ and $|\barm_{i'}^\ell - \barm_{i'}^r| \le \varepsilon$, we have there exists $C' > 0$ independent of $r$ and $\varepsilon$, 
\begin{align*}
&\frac{({\mu}_{b}-{\mu}_{i})^2 }{ 2 \sum_{s\in\mathcal{S}}\frac{ {g}(i,s)}{n^*_s}  +  \frac{\sigma_{i}^2}{\bar{m}^\ell_{i}} } - \frac{({\mu}_{b}-{\mu}_{i'})^2 }{ 2 \sum_{s\in\mathcal{S}}\frac{ {g}(i',s)}{n^*_s}  +  \frac{\sigma^2_{i'}}{\bar{m}^\ell_{i'} }}  \\
\le &\frac{({\mu}_{b}-{\mu}_{i})^2 }{ 2 \sum_{s\in\mathcal{S}}\frac{ {g}(i,s)}{n^*_s}  +  \frac{\sigma_{i}^2}{\bar{m}^r_{i}} } - \frac{({\mu}_{b}-{\mu}_{i'})^2 }{ 2 \sum_{s\in\mathcal{S}}\frac{ {g}(i',s)}{n^*_s}  +  \frac{\sigma^2_{i'}}{\bar{m}^r _{i'} }} + C'\varepsilon  \\
\le & (C+C')\varepsilon.
\end{align*}
Due to the arbitrary choice of $\varepsilon$, we obtain 
$$ \limsup_{\ell \rightarrow \infty } \frac{({\mu}_{b}-{\mu}_{i})^2 }{ 2 \sum_{s\in\mathcal{S}}\frac{ {g}(i,s)}{n^*_s}  +  \frac{\sigma_{i}^2}{\bar{m}^\ell_{i}} } - \frac{({\mu}_{b}-{\mu}_{i'})^2 }{ 2 \sum_{s\in\mathcal{S}}\frac{ {g}(i',s)}{n^*_s}  +  \frac{\sigma^2_{i'}}{\bar{m}^\ell_{i'} }} \le 0.$$
Switch $i$ and $i'$, symmetrically we can also prove $$ \limsup_{\ell \rightarrow \infty } \frac{({\mu}_{b}-{\mu}_{i'})^2 }{ 2 \sum_{s\in\mathcal{S}}\frac{ {g}(i',s)}{n^*_s}  +  \frac{\sigma_{i'}^2}{\bar{m}^\ell_{i'}} } - \frac{({\mu}_{b}-{\mu}_{i})^2 }{ 2 \sum_{s\in\mathcal{S}}\frac{ {g}(i,s)}{n^*_s}  +  \frac{\sigma^2_{i}}{\bar{m}^\ell_{i} }} \le 0.$$
Combining together we complete the proof.
 \hfill $\blacksquare$\par

\end{document}